\documentclass{article}
\usepackage{lmodern}
\usepackage{xcolor}
\usepackage{lipsum}

\usepackage{arxiv}

\usepackage[utf8]{inputenc} 
\usepackage[T1]{fontenc}    
\usepackage{hyperref}       
\usepackage{url}            
\usepackage{booktabs}       
\usepackage{amsfonts}       
\usepackage{nicefrac}       
\usepackage{microtype}      
\usepackage{lipsum}
\usepackage{graphicx}
\usepackage{subcaption}
\usepackage{float}
\usepackage{multirow}
\usepackage{multicol}
\usepackage{tabularx}
\usepackage{amsmath}
\usepackage{pdfpages}
\usepackage{siunitx}
\graphicspath{ {./Fig/} }
\usepackage{longtable}

\usepackage{color}
\definecolor{corr}{rgb}{0,0,0}
\newcommand{\modif}[1]{\textcolor{corr}{#1}}

\title{PEg TRAnsfer Workflow recognition challenge report: Do multi-modal data improve recognition?}

\author{
 Arnaud Huaulmé \\
  Univ Rennes\\
  INSERM, LTSI - UMR 1099\\
  F35000, Rennes, France\\
  \texttt{arnaud.huaulme@univ-rennes1.fr} 
   \And
  Kanako Harada\\
  Department of Mechanical Engineering\\
  the University of Tokyo\\
  Tokyo 113-8656, Japan
  \And
  Quang-Minh Nguyen \\
  Univ Rennes\\
  INSERM, LTSI - UMR 1099\\
  F35000, Rennes, France
  \And
  Bogyu Park \\
  VisionAI hutom \\
  Seoul, Republic of Korea
  \And
  Seungbum Hong \\
  VisionAI hutom \\
  Seoul, Republic of Korea
  \And
  Min-Kook Choi \\
  VisionAI hutom \\
  Seoul, Republic of Korea
  \And
  Michael Peven \\
  Johns Hopkins University\\
  Baltimore, USA
  \And
  Yunshuang Li\\
  Zhejiang University\\
  Hangzhou, China
  \And
  Yonghao Long\\
  Department of Computer Science and Engineering,\\ The Chinese University of Hong Kong,\\ Hong Kong
  \And
  Qi Dou\\
  Department of Computer Science and Engineering,\\ The Chinese University of Hong Kong,\\ Hong Kong
  \And
  Satyadwyoom Kumar\\
  Netaji Subhas University of Technology\\
  Delhi, India
  \And
  Seenivasan Lalithkumar\\
  National University of Singapore\\
  Singapore, Singapore
  \And
  Ren Hongliang\\
  National University of Singapore\\
  Singapore, Singapore\\
  The Chinese University of Hong Kong\\
  Hong Kong, Hong Kong
  \And
  Hiroki Matsuzaki\\
  National Cancer Center Japan East Hospital\\
  Tokyo 104-0045, Japan
  \And
  Yuto Ishikawa\\
  National Cancer Center Japan East Hospital\\
  Tokyo 104-0045, Japan
  \And
  Yuriko Harai\\
  National Cancer Center Japan East Hospital\\
  Tokyo 104-0045, Japan
  \And
  Satoshi Kondo\\
  Muroran Institute of Technology\\
  Hokkaido, Japan
  \And
  Mamoru Mitsuishi\\
  Department of Mechanical Engineering\\
  the University of Tokyo\\
  Tokyo 113-8656, Japan
  \And
  Pierre Jannin\\
  Univ Rennes\\
  INSERM, LTSI - UMR 1099\\
  F35000, Rennes, France\\
  \texttt{pierre.jannin@univ-rennes1.fr} 
}

\begin{document}
\maketitle
\begin{abstract}
\textbf{Background and Objective}: In order to be context-aware, computer-assisted surgical systems require accurate, real-time automatic surgical workflow recognition. In the past several years, surgical video has been the most commonly-used modality for surgical workflow recognition. But with the democratization of robot-assisted surgery, new modalities, such as kinematics, are now accessible. Some previous methods use these new modalities as input for their models, but their added value has rarely been studied. This paper presents the design and results of the “PEg TRAnsfer Workflow recognition'' (PETRAW) challenge with the objective of developing surgical workflow recognition methods based on one or more modalities and studying their added value.

\textbf{Methods}: The PETRAW challenge included a data set of 150 peg transfer sequences performed on a virtual simulator. This data set included videos, kinematic data, semantic segmentation data, and annotations, which described the workflow at three levels of granularity: phase, step, and activity. Five tasks were proposed to the participants: three were related to the recognition at all granularities simultaneously using a single modality, and two addressed the recognition using multiple modalities. The mean application-dependent balanced accuracy (AD-Accuracy) was used as an evaluation metric to take into account class balance and is more clinically relevant than a frame-by-frame score.

\textbf{Results}: Seven teams participated in at least one task with four participating in every task. The best results were obtained by combining video and kinematic data (AD-Accuracy of between 93\% and 90\% for the four teams that participated in all tasks). 

\textbf{Conclusion}: The improvement of surgical workflow recognition methods using multiple modalities compared with unimodal methods was significant for all teams. However, the longer execution time required for video/kinematic-based methods(compared to only kinematic-based methods) must be considered. Indeed, one must ask if it is wise to increase computing time by 2,000 to 20,000\% only to increase accuracy by 3\%. The PETRAW data set is publicly available at \url{www.synapse.org/PETRAW} to encourage further research in surgical workflow recognition.

\end{abstract}


\section{Introduction}

To fully integrate computer-assisted surgery systems in the operating room, a complete and explicit understanding of the surgical procedure is needed. A surgical process model (SPM) is a ``simplified pattern of a surgical process that reflects a predefined subset of interest of the surgical process in a formal or semi-formal representation'' \cite{Jannin2001}, thus allowing for the surgical procedure to be rigorously modeled and described. The SPM methodology consists of decomposing a surgical procedure into five increasingly-coarse levels of granularity: dexeme, surgeme, activity, step, and phase \cite{Lalys2013,Despinoy2015}. A dexeme, the lowest granularity level, is a numeric representation of the motion. A surgeme represents a surgical motion with an explicit semantic interpretation of the immediate motion (e.g., pulling). An activity describes the motion's overall action (action verbs; e.g., cut) performed on a specific target (e.g., the pouch of Douglas) by a specific surgical instrument (e.g., a scalpel). A step is the succession of these activities which together achieve a specific surgical objective (e.g., resection of the pouch of Douglas). Finally, a phase is the succession of steps that constitute a main period of the intervention (e.g., resection). SPM's are used for learning and expertise assessment \cite{Huaulme2018,Forestier2018}, robot assistance \cite{Ko2007}, operating room optimization and management \cite{Sandberg2005,Bhatia2007}, decision-making support \cite{Quellec2015}, and quality supervision \cite{Huaulme2020}.

The primary limitation of the state-of-the-art in SPM's \cite{Despinoy2015,Huaulme2018, Forestier2018,Sandberg2005,Quellec2015,Huaulme2020} is their need to be manually interpreted by human observers, which is observer-dependent, time-consuming, and subject to error \cite{Huaulme2019}. Thus, the proposed solutions can not be directly used to bring context-awareness into computer-assisted surgery applications in the operating room. To overcome this limitation, automatic workflow recognition methods have been developed for multiple granularity levels, including phase \cite{Bhatia2007,Padoy2010,Twinanda2017}, step \cite{Bouarfa2011,James2007}, and activity \cite{Ko2007,Lalys2012}. With the emergence of deep learning, most of these recent automatic workflow recognition methods are based on convolutional neural networks, such as AlexNet \cite{krizhevsky2012imagenet} or ResNet \cite{He2016DeepRecognition}; on recurrent neural networks, such as LSTM \cite{Hochreiter1997LongMemory} or gated recurrent unit (GRU) \cite{cho2014properties}; and more recently on transformers \cite{vaswani2017attention}.

Along with what methodology to use, it is also an open question as to which data modalities should be used as input for this task. In robot-assisted surgery and virtual reality training environments, video and kinematic data are both readily available. Despite this, most state-of-the-art workflow recognition methods are based on a single modality, such as only video \cite{Sarikaya,Funke2019} or only kinematic data \cite{Despinoy2015,DiPietro2019}. Few studies have used workflow recognition method based on both video and kinematic data \cite{Huaulme2021MIcro-SurgicalReport,Long2020RelationalSurgery,Qin2021LearningEstimation}. However, with the exception of the study by Long \textit{et al.}\cite{Long2020RelationalSurgery}, they do not compare the results obtained based on the number and type of input modalities.

Semantic segmentation of surgical video is also essential for surgical understanding and is an active area of research. For example, in five editions of the EndoVis MICCAI Challenge (2015 to 2020), six of the 19 proposed sub-challenges were dedicated to this topic. However, to the best of our knowledge, semantic segmentation has rarely been used as a supplementary task paired with, or as additional input for, surgical workflow recognition.

Therefore, the “PEg TRAnsfer Workflow recognition by different modalities” (PETRAW) sub-challenge, which is part of EndoVis, provided a unique data set for automatic recognition of surgical workflows containing video, kinematic, and segmentation data on 150 peg transfer training sequences. Participants were asked to develop model(s) to recognize phases, steps, and activities using one or several of the available modalities.

\section{Methods: Challenge Design}
This section describes the challenge design, organization, objective, data set, and assessment methods.

\subsection{Challenge organization}\label{sec:challenge_organization}
The PETRAW challenge was a one-time event organized as part of EndoVis during the online 2021 international conference on Medical Image Computing and Computer-Assisted Intervention (MICCAI2021). Four people were involved in the organization: Arnaud Huaulmé and Pierre Jannin from the University of Rennes 1 (France), and Kanako Harada and Mamoru Misthuishi from Tokyo University (Japan). Complete information about the challenge was made available to participants using the Synapse platform: \url{www.synapse.org/PETRAW}.

Challenge participants were subject to the following rules:
\begin{itemize}
\item Participants had to submit a fully automatic method that could recognize phases, steps, and activities on the same model using one or several modalities; and
\item Only data provided by the organizers and publicly available data sets, including pre-trained networks, were authorized for use in training. The publicly available data sets must have been open or otherwise available to all participants at the time the PETRAW data set was released.
\end{itemize}
The results of all participating teams were announced publicly during the challenge day. Challenge organizers and people from the organizing institutions could also participate in the challenge but were excluded from the competitive rankings. Participating teams were encouraged (but not required) to provide their code as open access.

For a valid submission, the participating teams had to provide the following elements: a write-up, a Docker image allowing the organizers to compute the results, and a pre-recorded talk to limit technical issues during the challenge day (online event). Multiple Docker images could be submitted, but only the last submission was officially used to generate the evaluation results. No leaderboard or evaluation results were provided prior to the challenge day.

The challenge schedule was as follows: The training data set, including videos, kinematic data, and workflow annotations, was released on June 1, 2021; corresponding semantic segmentation data was released on June 9, 2021; submissions were accepted until September 12, 2021 (23:59 PST); and the evaluation results were announced on October 1, 2021, during the online MICCAI2021 event. Some teams obtained unexpectedly poor results (i.e., workflow recognition rates inferior to 50\%), which made further analysis of the results not relevant. Therefore, each team was allowed to provide a new submission before October 31, 2021. The teams that made a new submission are identified in Section \ref{subsec:Teams}. The challenge test data set and the organizers' evaluation scripts were released with this paper at \url{www.synapse.org/PETRAW}

\subsection{Challenge objective}
The objective of the PETRAW challenge was to study the contribution of each modality (either alone or in combination) to surgical workflow recognition. To achieve this goal, participants were asked to create a single classification model to determine the surgical task at three levels of granularity (phase, step, and action). Five different tasks were offered as part of the challenge: three concerned the development of unimodal models (i.e., video-based, kinematic-based, or semantic segmentation-based models); and two concerned multimodal-based models. The unimodal-based models were used as a baseline for comparison with the multimodal-based models. In order to keep to a reasonable number of tasks, not all multimodal configurations could be studied. For models based on semantic segmentation data (and to reflect the fact that clinically this modality can be only obtained through a trained segmentation model), participants were asked to use the output of such model as input for PETRAW.

\subsection{Challenge data set}
The challenge data set was composed of 150 sequences of peg transfer training sessions. The objective of the peg transfer session was to transfer six blocks from the left peg to the right and then back. Each block needed to be extracted from the peg using a grasper (operated by one hand), transferred to the other grasper (in the other hand), and finally inserted onto the peg on the opposite side of the board.

All sequences were acquired by a non-medical expert at the LTSI Laboratory, University of Rennes 1, France. The data set was divided into training data (n=90 sequences) and test data (n=60 sequences). Each sequence included kinematic data, video, semantic segmentation of the video for each frame, and workflow annotations at each level of granularity. Only the training data set was provided to participants. 

\subsubsection{Data acquisition}
\begin{figure}[H]
    \centering
    \includegraphics{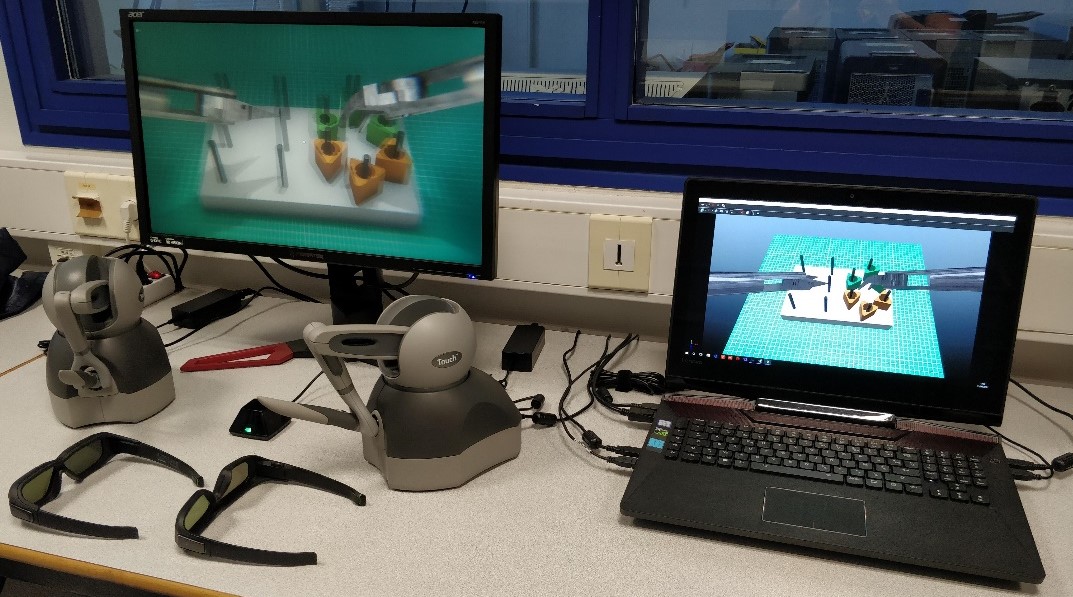}
    \caption{The virtual reality simulator used for data acquisition.}
    \label{fig:simulator}
\end{figure}

The challenge data was acquired on a virtual reality simulator (Figure~\ref{fig:simulator}) developed at the Department of Mechanical Engineering, University of Tokyo, Japan \cite{HerediaPerez2018HapticSimulation}, consisting of a laptop (i7-700HQ, 16Go RAM, GTX 1070), a 3D rendering setup (3D screen: 24 inches, 144Hz; and 3D glasses), and two haptic user interfaces (3D system Touch\textsuperscript{TM}).

For data acquisition, a single operator performed a series of five consecutive peg transfer tasks followed by a break of at least 5 hours to limit fatigue. This was repeated 30 times to yield a total of 150 peg transfer task sequences. The COVID-19 crisis (acquisition made in 2020-2021) did not allow us to recruit multiple participants. To limit the effect of immediate learning or fatigue in a single session, three sequences from each series were randomly chosen for training, and the remaining two for testing.
 
The kinematic data and videos were synchronously acquired at 30 Hz during each peg transfer task. Each video had a resolution of 1920x1080 pixels and semantic segmentation was performed for each frame off-line following the task. Kinematic data included the position, rotation quaternion, forceps aperture angle, linear velocity (obtained from simulation, not derived from position), and angular velocity (obtained from simulation, not derived from orientation) of the left and right instruments (i.e., graspers). The position and linear velocity were measured in centimeters and centimeters per second, respectively. The angle and angular velocity were measured in degrees and degrees per second, respectively.

\begin{figure}[H]
    \centering
    \includegraphics[width=.5\linewidth]{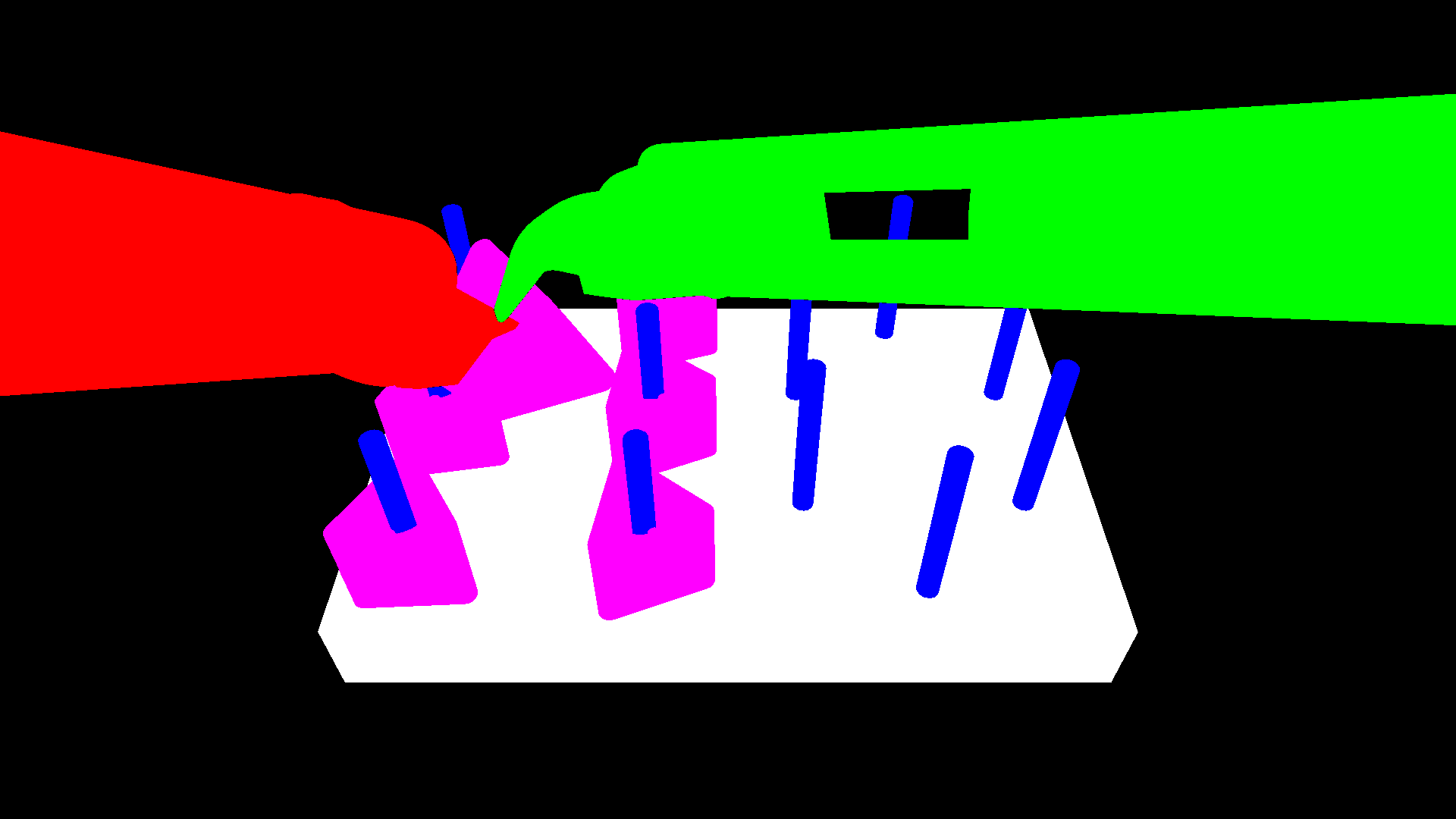}
    \caption{Representative segmentation mask with the six classes: background (black), base (white), left instrument (red), right instrument (green), pegs (blue) and blocks (magenta).}
    \label{fig:segmentation_frame}
\end{figure}

The semantic segmentation included six classes (shown in Figure~\ref{fig:segmentation_frame}): background (black, hexadecimal code:\#000000), base (white, \#FFFFFF), left instrument (red, \#FF0000), right instrument (green, \#00FF00), pegs (blue, \#0000FF), and blocks (magenta, \#FF00FF).

The workflow annotations were automatically computed using the scene information and the ASURA method \cite{Huaulme2019}. The challenge organizers had previously demonstrated in \cite{Huaulme2019} that ASURA is more accurate and robust than manual annotation on peg transfer tasks. Two phases, twelve steps, six action verbs, two targets, and one surgical instrument were identified to describe the workflow (Table~\ref{tab:vocabulary}). Each phase corresponded to the transfer of all of the blocks in one direction (e.g. ``L2R'' for left to right). Each step (six per phase) corresponded to the transfer of a single block (e.g.``Block1 L2R'' for the transfer of the first block from the left to the right). For the activities, two targets were differentiated: ``block'' and ``other block''. ``Block'' corresponds to the one that is currently being transferred. ``Other block'' is an additional target used to differentiate when the user accidentally interacts with any block other than the one to be transferred.

\begin{table}[!ht]
\begin{center}
\caption{Peg-transfer vocabulary. }
\label{tab:vocabulary}
\begin{tabular}{|c|c|c|c|c|c|}
\cline{1-2}\cline{4-6}
   \multirow{2}{*}{Phases}&\multirow{2}{*}{Steps}&& \multicolumn{3}{c|}{Activities}\\\cline{4-6}
   &&&Verb&Target&Tool\\\cline{1-2}\cline{4-6}
   \multirow{6}{.20\linewidth}{Transfer Left To Right (L2R)}  & Block 1 L2R&&Catch&Block&Grasper \\
                                           & Block 2 L2R& &Drop& Other block&\\
                                           & Block 3 L2R& &Extract&&\\
                                           & Block 4 L2R& &Hold&&\\
                                           & Block 5 L2R& &Insert&&\\
                                           & Block 6 L2R& &Touch&&\\\cline{1-2}\cline{4-6}
   \multirow{6}{.20\linewidth}{Transfer Right To Left (R2L)} & Block 1 R2L \\
                                           & Block 2 R2L \\
                                           & Block 3 R2L \\
                                           & Block 4 R2L \\
                                           & Block 5 R2L \\
                                           & Block 6 R2L \\\cline{1-2}
                                           
\end{tabular}
\end{center}
\end{table}

One limitation of the method presented by \cite{Huaulme2019} was the inability to accurately differentiate between the action verbs “catch” and “touch”, as each tool tip was considered as a unique virtual object. 
The virtual reality simulator was updated to include four separating regions rather than one, allowing these actions to be readily differentiated. Accordingly, the workflow annotations were manually examined and corrected to ensure annotation quality.

\subsubsection{Data pre-processing}
The original workflow annotations were formatted in terms of start and finish time, expressed in milliseconds. These annotations were sampled to provide a discrete sequence at 30Hz, synchronized with the kinematic, video, and segmentation data to allow for frame-by-frame annotation. 
Due to their lack of variability, the two targets and the tool were not included in the workflow annotation. Furthermore, when no phase, step, or activity occurred, the term ``idle'' was used. For each timestamp, the following information was provided: timestamp\_number, phase\_value, step\_value, verb\_Left\_Hand, verb\_Right\_Hand.

\subsubsection{Ground truth uncertainties}
\label{seq:GT_uncertainties}
\begin{figure}
    \centering
    \includegraphics[width=.5\linewidth]{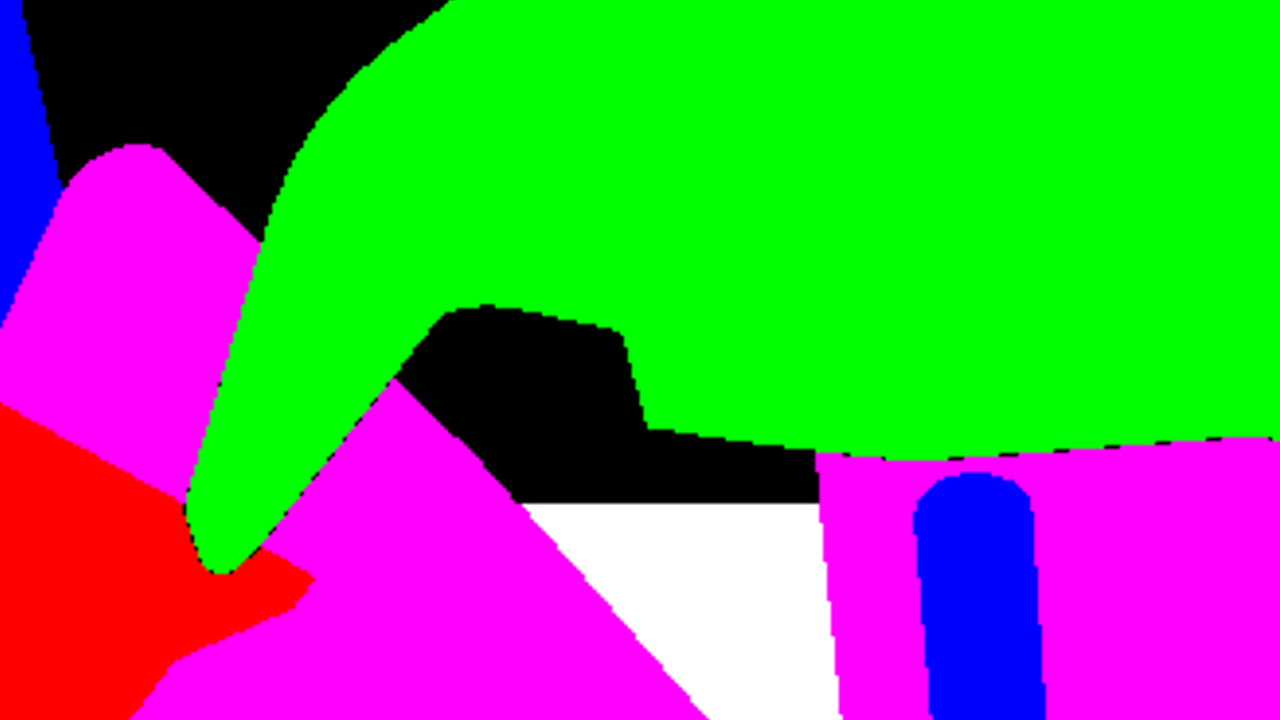}
    \caption{Zoom of 219x123 pixels from Figure~\ref{fig:segmentation_frame} to highlight segmentation errors. Right instrument/block (green/magenta) and left/right instruments (red/green) errors are shown where pixels are labeled as background (black). On this zoom, only 51 pixels were miss-segmented (around 0.2\%).}
    \label{fig:segmentation_zoom}
\end{figure}

The semantic segmentations were the primary source of uncertainty in the ground truth. Due to the transformation of 3D meshes into 2D images, some pixels were attributed to the wrong class, especially at boundaries between the right instrument/peg, left instrument/peg, left instrument/block, and left/right instruments (Figure~\ref{fig:segmentation_zoom}). We estimated this uncertainty by counting the number of mis-segmented pixels on 10 images that included many boundary regions, such as those between surgical instruments, pegs, and blocks. On each image, the number of mis-segmented pixels represents less than 0.25\% of the total image. To take into account the fact that this manual assessment was not representative of the whole data set, we estimated that this mis-segmentation represents less than 0.5\% of pixels.

Workflow annotations were another source of uncertainty. Although the ASURA method is consistent (i.e., it generates the same result in two identical situations) and a manual check was performed to limit inaccuracies, some components could not be recognized with complete certainty. Two particular instances were identified. First, in sequence 130 of the training data set, the block in step ``Block 1 R2L'' was inserted in a non-standard way. Specifically, the block was released by the operator, and while falling became inserted in the peg. Therefore, the insert action was absent. The other instance concerned sequence 79 of the test data set. This time, the operator caught a block before the previous one had been fully inserted, leading to an overlap between the steps ``Block 5 R2L'' and ``Block 6 R2L''. The second was chosen as the sole annotation to maintain the true beginning of the step.

\subsubsection{Data set characteristics}

\begin{figure}[]
    \centering
     \begin{subfigure}[t]{0.45\textwidth}
         \centering
         \includegraphics[width=\textwidth]{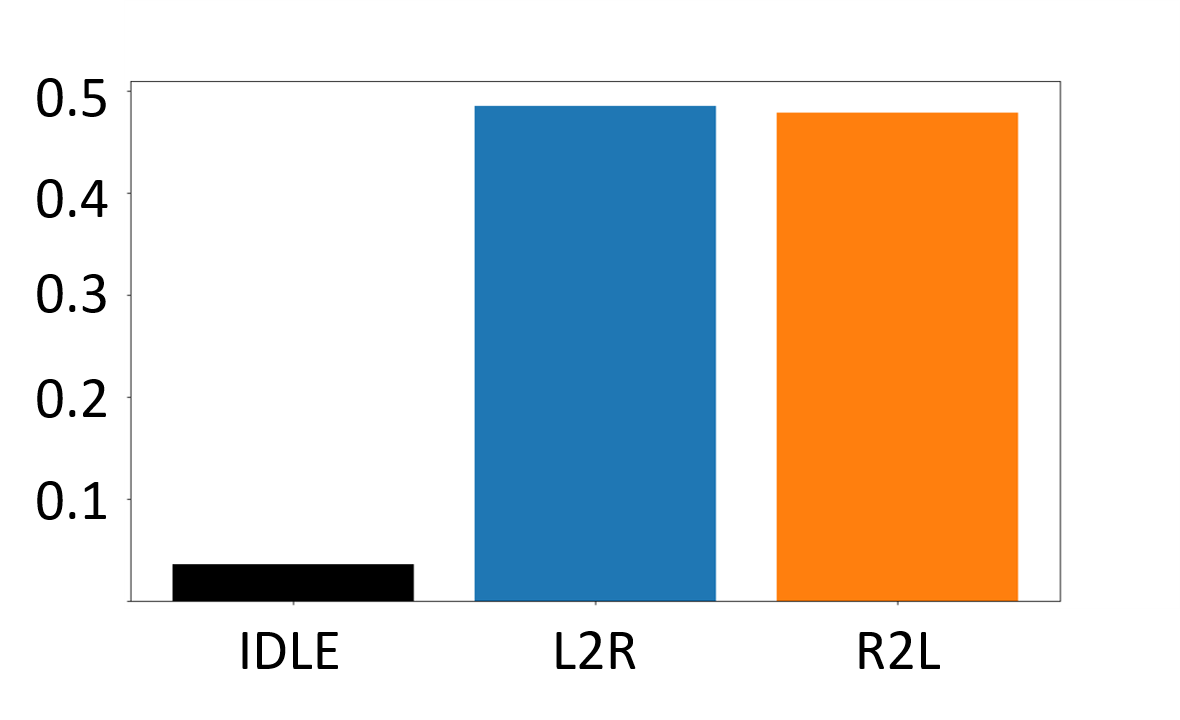}
         \vspace{-10mm}
         \caption{Training phases}
         \label{fig:Training_Phase}
     \end{subfigure}
     \hfill
     \begin{subfigure}[t]{0.45\textwidth}
         \centering
         \includegraphics[width=\textwidth]{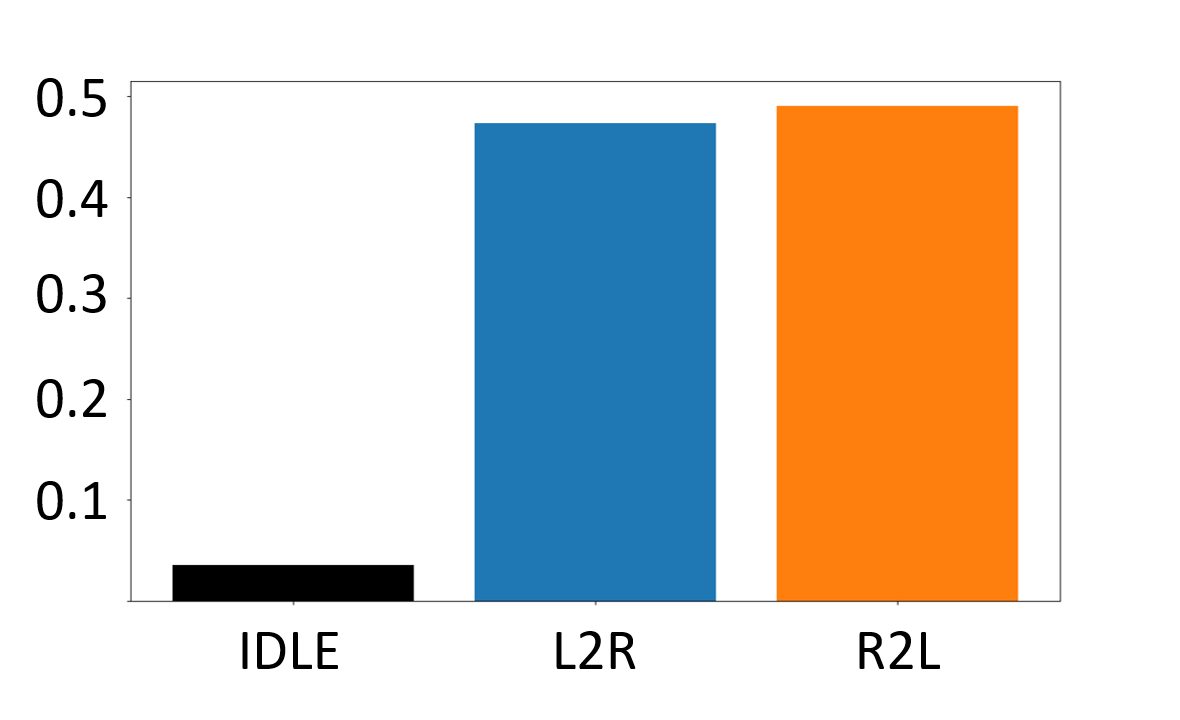}
         \vspace{-10mm}
         \caption{Test phases.}
         \label{fig:Test_Phase}
     \end{subfigure}
     \\
     \begin{subfigure}[t]{0.45\textwidth}
         \centering
         \includegraphics[width=\textwidth]{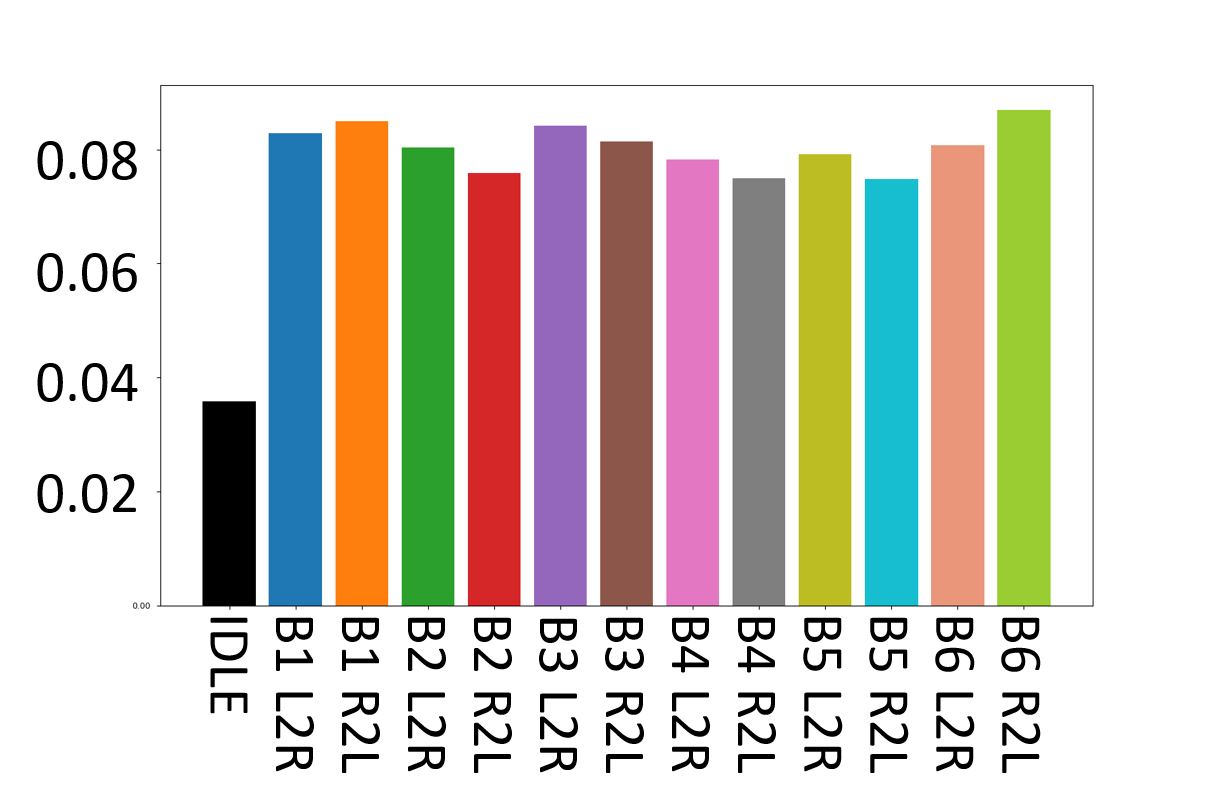}
         \vspace{-10mm}
         \caption{Training phases.}
         \label{fig:Training_Step}
     \end{subfigure}
     \hfill
     \begin{subfigure}[t]{0.45\textwidth}
         \centering
         \includegraphics[width=\textwidth]{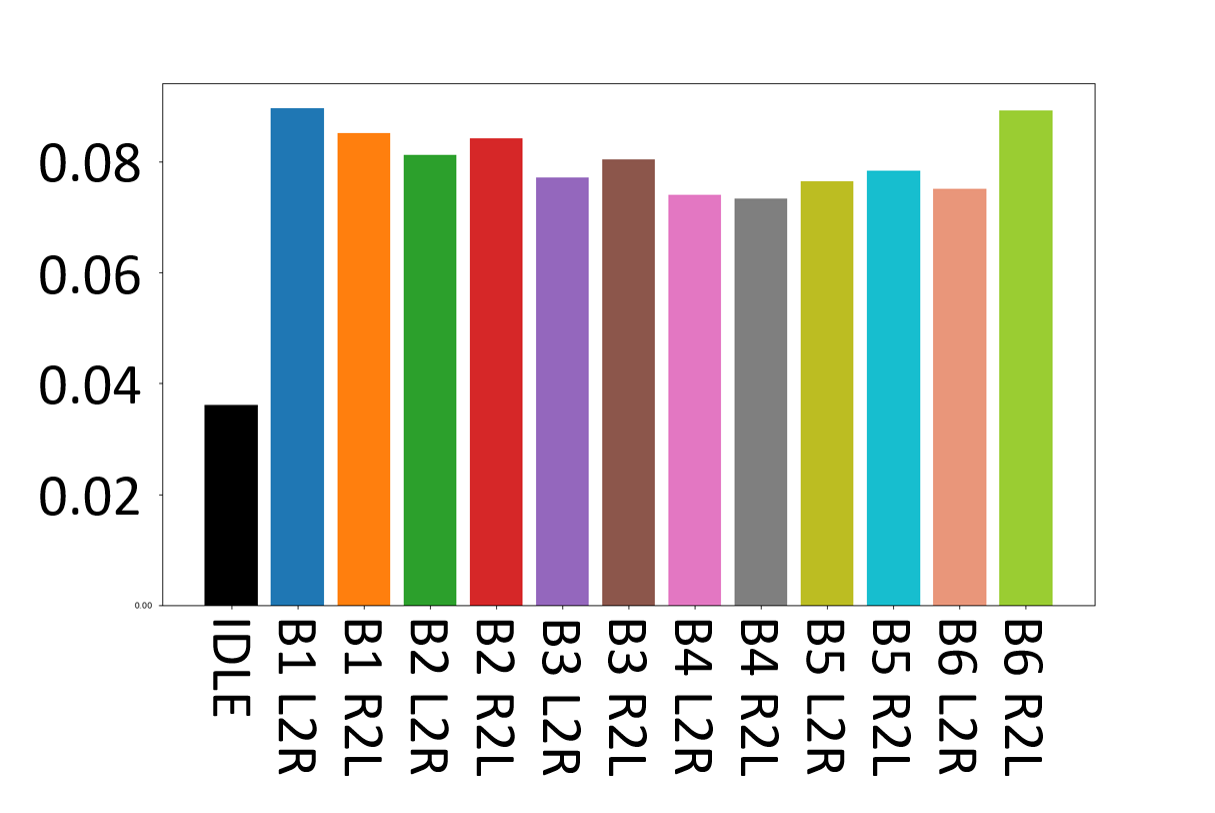}
         \vspace{-10mm}
         \caption{Test steps.}
         \label{fig:Test_Step}
     \end{subfigure}
     \\
      \begin{subfigure}[t]{0.45\textwidth}
         \centering
         \includegraphics[width=\textwidth]{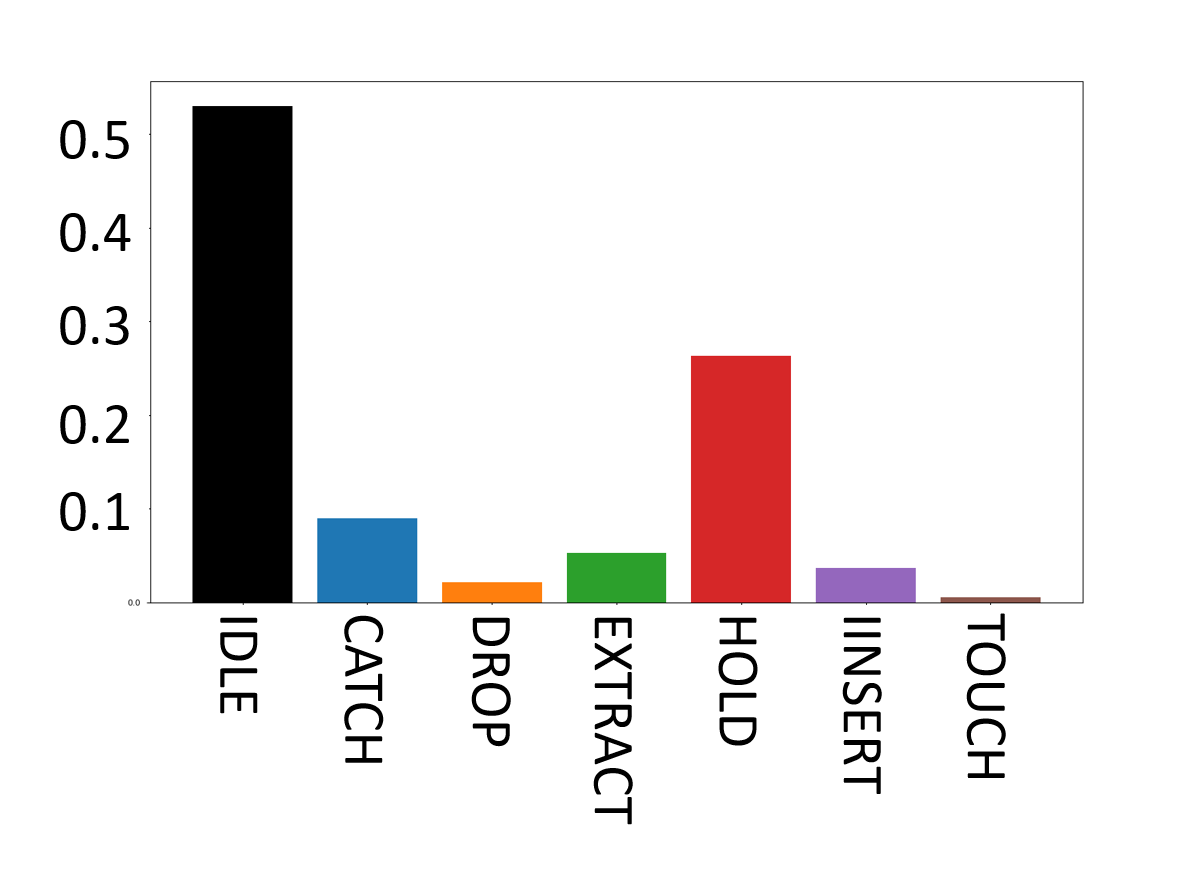}
         \vspace{-10mm}
         \caption{Training verb left hand.}
         \label{fig:Training_Verb_Left}
     \end{subfigure}
     \hfill
     \begin{subfigure}[t]{0.45\textwidth}
         \centering
         \includegraphics[width=\textwidth]{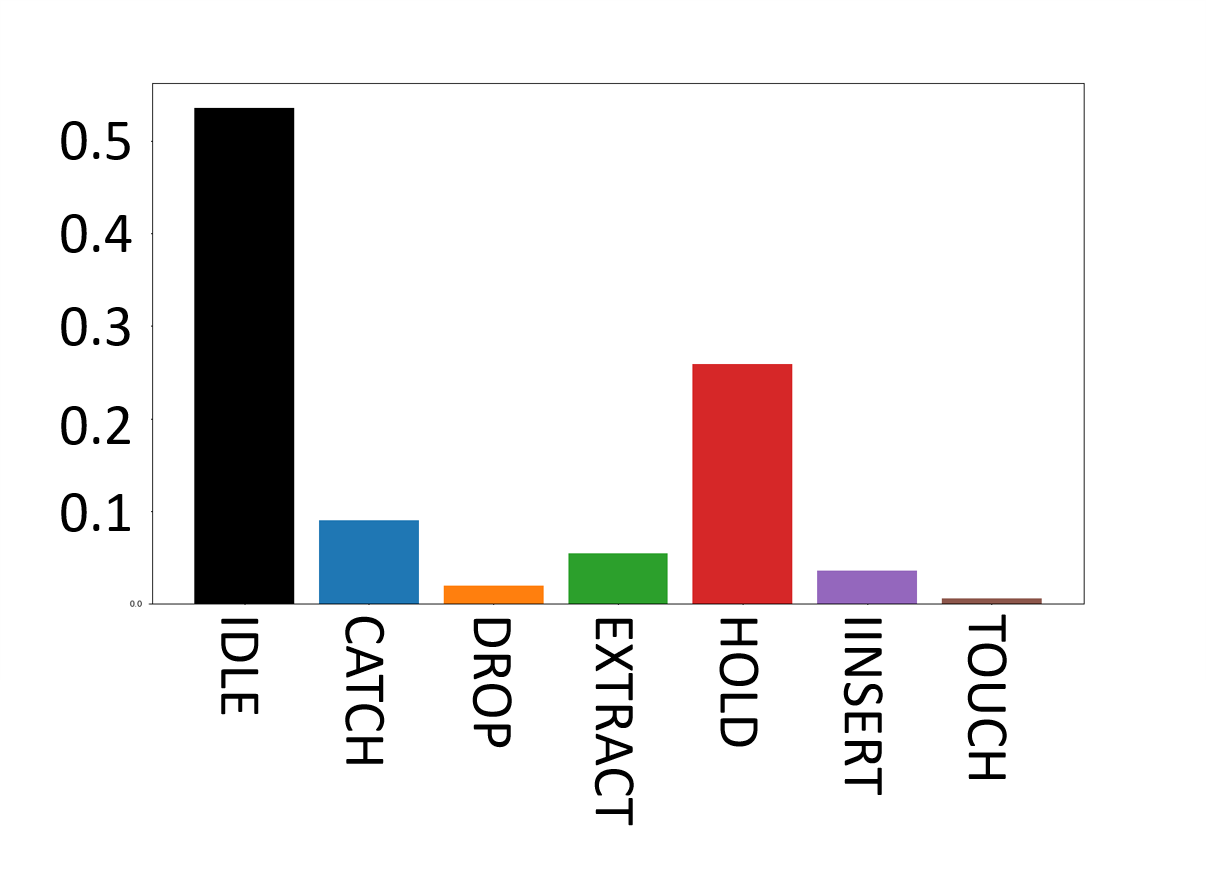}
         \vspace{-10mm}
         \caption{Test verb left hand.}
         \label{fig:Test_Verb_Left}
     \end{subfigure}
     \\
      \begin{subfigure}[t]{0.45\textwidth}
         \centering
         \includegraphics[width=\textwidth]{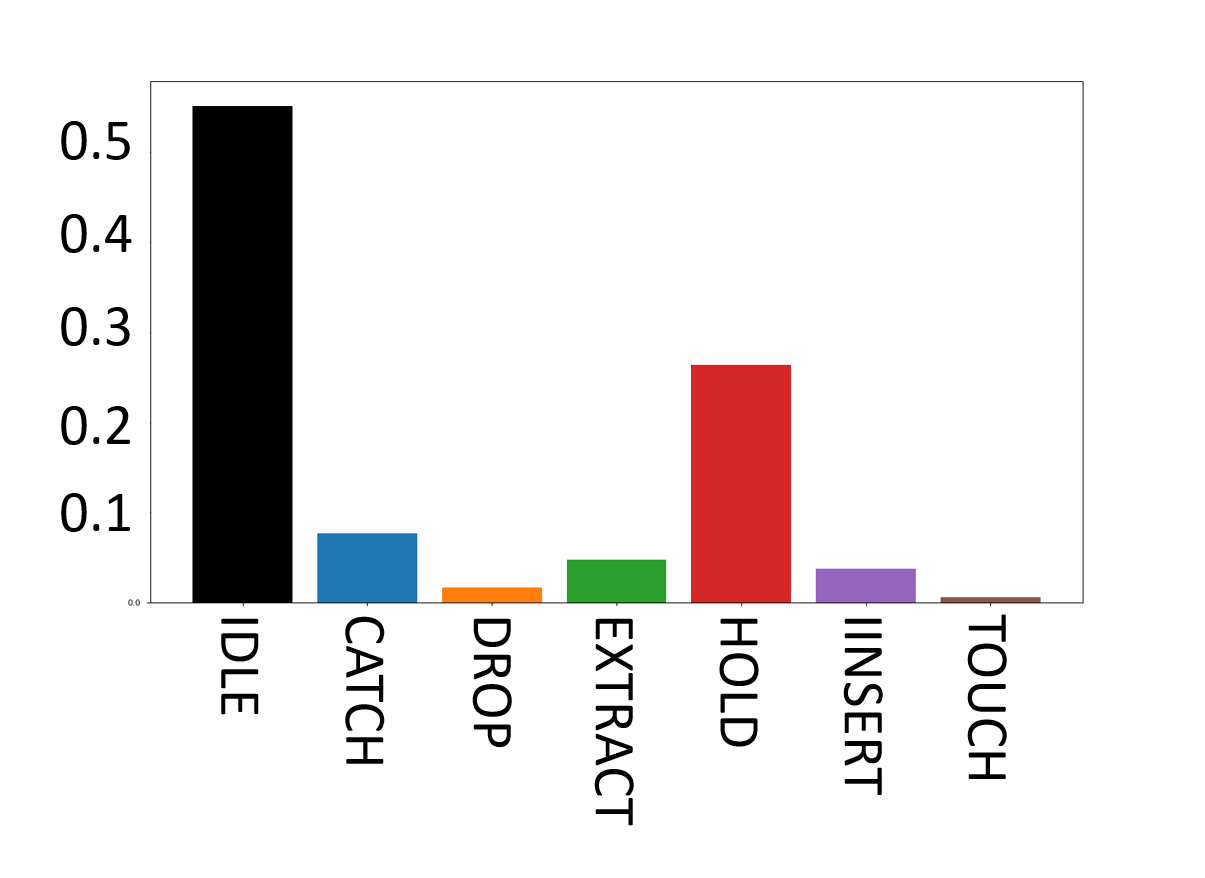}
         \vspace{-10mm}
         \caption{Training verb right hand.}
         \label{fig:Training_Verb_Right}
     \end{subfigure}
     \hfill
     \begin{subfigure}[t]{0.45\textwidth}
         \centering
         \includegraphics[width=\textwidth]{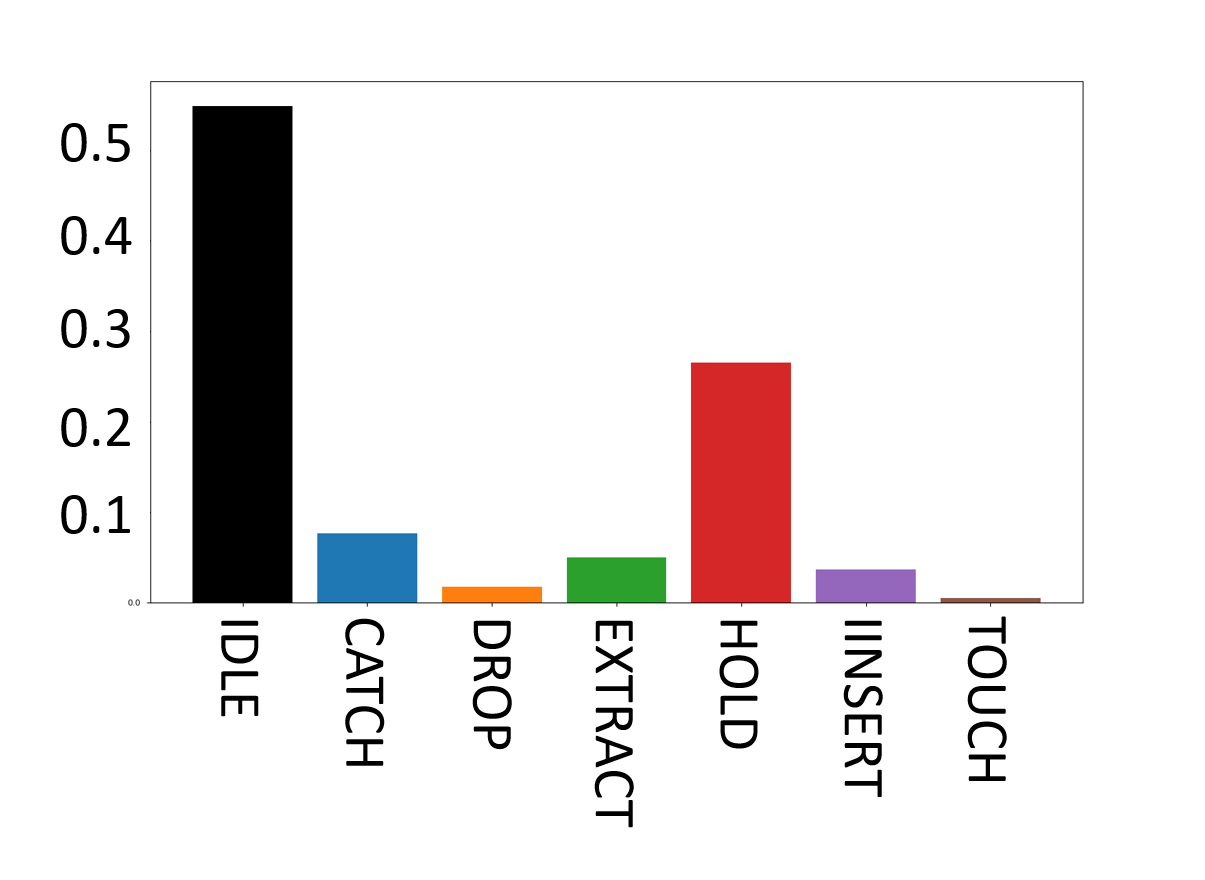}
         \vspace{-10mm}
         \caption{Test verb right hand.}
         \label{fig:Test_Verb_Right}
     \end{subfigure}
     \\
        \caption{Distribution of each term at each granularity level in the training and test data sets.The y-axis represents the percentage of frames. In (\subref{fig:Training_Phase}) and (\subref{fig:Test_Phase}), ``L2R'' means transfer left to right and ``R2L'' means transfer right to left. In (\subref{fig:Training_Step}) and (\subref{fig:Test_Step}), ``B1 L2R'' means block 1 left to right, ``B2 L2R'' means block 2 left to right.}
        \label{fig:GranularityDistribution}
\end{figure}
The training and test data sets presented similar characteristics. The mean and standard deviation duration was $140.2\pm18.9$ seconds for the training data set and $141.7\pm18.0$ seconds for the test data set. Figure \ref{fig:GranularityDistribution} presents the distribution of every vocabulary component for each granularity level in the training data set (Figures~\ref{fig:Training_Phase}, \ref{fig:Training_Step}, \ref{fig:Training_Verb_Left}, \ref{fig:Training_Verb_Right}) and the test data set (Figures~\ref{fig:Test_Phase}, \ref{fig:Test_Step}, \ref{fig:Test_Verb_Left}, \ref{fig:Test_Verb_Right}). Even for underrepresented components, the distribution was very similar in both data sets. For instance, the verb ``touch'' (left hand) represented 0.59\% and 0.60\% of the samples in the training and test data sets, respectively, and ``touch'' (right hand) represented 0.62\% and 0.48\%, respectively. The distribution of each vocabulary component between each data set is only statistically different (Mann-Whitney test) for two steps: ``Block 1 L2R'' and ``Block 6 L2R'', with p=0.045 and p=0.036 respectively.

Another important characteristic of the data sets was the high class unbalance of at least one vocabulary term for each granularity level. For the phases, the term ``idle'' represented less than 4\% of all data, whereas the other phase terms accounted for more than 47\% (L2R and R2L). For the steps, the term ``idle'' represented less than 4\%, whereas the non-idle steps accounted for approximately more than 7.5\% of each data set(Figures ~\ref{fig:Training_Phase}-\ref{fig:Test_Step}). This unbalance was more pronounced at the action level, where the least represented verb (i.e., ``touch'') represented approximately 0.6\% of the data set, whereas the verb ``idle'' accounted for more than 53\%. The detailed distribution values for each granularity level in both data sets are provided in supplementary material.

\subsection{Assessment method}

\subsubsection{Metrics}
To assess the participants' workflow recognition models and to take into account the high class unbalance, balanced versions of accuracy, precision, recall, and F1 were used. 

In practice, however, some small variations in surgical task recognition are not clinically meaningful and do not constitute a true error. Motivated by this, Dergachyova \textit{et al.} \cite{Dergachyova2016AutomaticWorkflow} proposed a re-estimation of these classic frame-by-frame scores, called application-dependent scores, to take into account an acceptable delay $d$. When a predicted transition occurs within a transition window ($2d$) centered on the ground truth transition, all frames between the two transitions are considered correct if it is the same transition type ( e.g. transition for verb ``catch'' or verb ``extract''). Therefore, the balanced application-dependent accuracy (AD-Accuracy) was used and the acceptable delay was fixed at 250 ms.

To assess the participants' segmentation models, the mean Intersection-Over-Union (IoU) over all classes was also used, also known as the Mean Jaccard Index over all classes. The IoU is the area of overlap between the predicted segmentation ($Pred$) and the ground truth ($GT$), divided by the area of union between the $Pred$ and the $GT$. In our cases, there was a multi-class segmentation problem, therefore the mean IoU value of the image was calculated by taking the IoU of each class and averaging it over the classes:

\begin{align}
    MeanIoU_{frame} &= \frac{1}{6} \sum_{class} IoU_{class}\nonumber\\
                    &= \frac{1}{6} \sum_{class} \frac{|GT \cap Pred|_{class}}{|GT \cup Pred|_{class}}\\
                    &=\frac{1}{6} \sum_{class} \frac{TP_{class}}{TP_{class}+FP_{class}+FN_{class}},\nonumber
\end{align}
where $TP$ (True Positives) is the number of pixels inside the $GT$ area that are correctly predicted, $FP$ (False Positives) is the number of pixels outside the $GT$ area but predicted as belonging to the class, and $FN$ (False Negatives) is the number of pixels inside the $GT$ area that are incorrectly predicted. 

\subsubsection{Ranking method}\label{subsect:ranking}
The ranking of the participating methods used only the surgical task recognition metrics. Metrics computed for evaluating the segmentation models were provided for information purposes only.

A metric-based aggregation method using the AD-Accuracy values across all test sequences was used for the ranking. Metric-based aggregation was used according to the recommendations made in \cite{Maier-Hein2018}, which show it to be one of the most robust. As all tasks consisted of recognizing the phase, step, and the actions of the left and right hands (i.e., the left and right verbs), the ranking score for the algorithm $a_i$ was computed as follows:
\begin{equation}
    s(a_i) = \frac{s_{phase}(a_i)+s_{step}(a_i)+s_{verb\_left}(a_i)+s_{verb\_right}(a_i)}{4}
\end{equation}
with,
\begin{equation}
    s_{phase}(a_i) = \frac{\sum_{t=0}^{T}phase\_balance\_accuracy\_case\_t}{T},
    \label{equ:s_phase}
\end{equation}
where $T$ is the number of sequences to test. Similar equations were used for the other terms ($s_{step}(a_i)$, $s_{verb\_left}(a_i)$ and $s_{verb\_right}(a_i)$) with a numerator specific to each, i.e., $\sum_{t=0}^{T}step\_balance\_accuracy\_case\_t$ for $s_{step}(a_i)$, etc.

If a participant method did not produce a prediction for one or several granularity levels, the accuracy given for each missing granularity level was that expected for uniformly random predictions. For example, if a model did not predict the phase, $s_{phase}$ would be set to 1/3 corresponding to the phase having 3 potential values. In practice, this was not encountered and each evaluated model produced results for each level of granularity.

Ranking stability was assessed by testing different ranking methods: meanThenRank, medianThenRank, rankThenMean, rankThenMedian, and testBased. MeanThenRank was chosen for the ranking. MedianThenRank differs from the previous method because it used the median instead of the mean in equation \ref{equ:s_phase}. For rankThenMean and rankThenMedian, first, the results of each sequence were ranked among participants, and then the final results were the mean or median of all ranks. The testBased method is based on bootstrapping. The ranking was considered stable if a team was ranked in the same position with the majority of ranking methods. If the ranking was not stable according to the chosen methods, a tie between teams was pronounced. The ranking computation and analysis were performed with the ChallengeR package provided by \cite{Wiesenfarth2021MethodsResults}.

\subsubsection{Online recognition compatibility}
To be online compatible, the proposed methods must satisfy two conditions:
\begin{itemize}
\item to produce predictions faster than the duration between the two samples (i.e., faster than 30 Hz); and
\item to be causal (i.e., not use data from a future time point to make predictions).
\end{itemize}
The computation time was not studied because it could not be assessed fairly for all teams. Indeed, the teams provided a unique Docker image for all tasks, and some teams did not write the output file to standard output as it was received, which did not allow for their durations to be precisely measured. 

To verify that the methods were causal, the online availability of the frames was mimicked. One additional sequence of 10 seconds, corresponding to the transfer of the first block from the left to the right, was recorded. This sequence was used to generate 300 sub-sequences, each one a frame longer than the previous. Thus, the first sequence only contained the information of the first frame, the second one contained the information of the two first frames, etc. The models were run on the 300 sub-sequences and the last prediction of each sub-sequence to create a definitely causal prediction sequence. A method was considered causal if and only if this definitely causal prediction sequence was identical to the prediction sequence given by the full 300 frames. This causality-testing method is fully automated and also takes into account the complete pipeline used to perform the prediction, such as pre- and post-processing steps, which could lead to a non-causal method even if the network only uses causal components. For reasons of computation time and environmental responsibility, this test was not performed on a whole sequence or the whole test data set. By testing the entire data set, we could be more confident in the causality of the proposed methods, but this would quickly display diminishing returns.

\subsection{Additional analyses}
\modif{To further analyze the impact of using multimodal instead of unimodal models, we performed two additional analyses that were not initially included in the challenge design: the statistical significance to use multimodal models instead of unimodal models, and the execution time. These additional analyses only concerned the teams that participated in the multimodal tasks (4 and 5) with a combination of the same or similar models used for the unimodal tasks.}

\subsubsection{Comparison between unimodal and multimodal models}
\modif{To assess the impact of each modality and its combinations on automatic workflow recognition, we performed a statistical analysis with the Wilcoxon test. The difference was significant if the p-value was inferior to 0.05.}

\subsubsection{Execution time}
\modif{Performance is not the only important factor when developing automatic recognition models. Indeed, environmental aspects must also be taken into account \cite{Jannin2021TowardsCare}. To answer this question, we examined the execution time to compute the results of the 60 test sequences. These durations were interpolations that assumed the predictions in each task were computed independently and not the real execution time. Indeed, one team (Hutom, see section~\ref{subsubsec:Hutom}) used the predictions from tasks 1 to 3 as input for those of tasks 4 and 5, so the interpolation for the multimodal tasks took into account the execution time for the unimodal ones.}

\section{Results: Reporting of the Challenge Outcomes}
\subsection{Challenge submission}
By September 12, 2021, 29 participants had registered for the PETRAW challenge: 17 were members of one of the six competing teams. The organizers also submitted results as a non-competing team to provide a baseline. As explained in Section \ref{sec:challenge_organization}, some teams obtained unexpected results and three teams resubmitted results for at least one task.

\subsection{Information on the participating teams and their methods}
\label{subsec:Teams}
This section describes each team, the methods they used, and the tasks in which they participated. Competing teams are presented in alphabetical order and not in terms of their ranking.

\subsubsection{Hutom}
\label{subsubsec:Hutom}
The Hutom team (Bogyu Park, Seungbum Hong, and Minkook Choi from VisionAI hutom) participated in all proposed tasks. They resubmitted a Docker image for all tasks except the kinematic-based recognition task.

Before training, they performed a simple pre-processing step. To preserve temporal information, they split data into clips of 8 frames. They normalized kinematic data by standardizing the raw input without data augmentation. They resized video data to $256\times256$ pixels, followed by random cropping ($224\times224$ pixels) and normalization. The cropping was limited to preserve the spatial information in each frame of the clip. They resized segmentation data to $512\times512$ pixels.

They used a similar baseline architecture for tasks based on the same modality. They computed segmentation data from the video recording using a DeepLabV3+ architecture \cite{Chen2018Encoder-decoderSegmentation}. They used a 3D ResNet network \cite{Hara2018CanImageNet} for workflow recognition based on the video modality. For the segmentation modality, they used a SlowFast50 network \cite{Feichtenhofer2019SlowFastRecognition} for segmentation-based recognition and a 3D ResNet network for video/kinematic/segmentation-based workflow recognition. They inputted kinematic data on a bi-directional long short-term memory (Bi-LSTM) network \cite{Schuster1997BidirectionalNetworks}. For multimodal recognition tasks, they used a convolutional feature fusion layer to efficiently perform the fusion of the feature output of each modality. They obtained embedding features with individual modal inputs from each model trained accordingly. Then, they compared the embedding features of each modality with those of other modalities to learn the different representations of each modality. They used the stop gradient-based SimSiam method \cite{Chen2020ExploringLearning} to compare representations between embedding features. Concomitantly, they stacked embedding features by modality into one block as a chunk and fused them into one embedding through a convolution operation. The approach assumed that feature elements for each modality in the same column have similar temporal information in similar positions. For all networks, they used the Adam optimizer and an initial learning rate of 1e\textsuperscript{-3}, with a combination of Equalization loss v2 \cite{Tan2020EqualizationDetection} and Normsoftmax Loss \cite{Zhai2018ClassificationLearning} as long-tail recognition for addressing data imbalance.

\subsubsection{JHU-CIRL}
The JHU-CIRL team (Michael Peven and Gregory D. Hager; Johns Hopkins University) participated in the kinematic-based workflow recognition task.

They performed an under-sampling of the kinematic data to reduce the time dimension size in order to prevent vanishing gradient issues during training. For the test, they used the same under-sampling. The JHU-CIRL team did not perform any other pre-processing because they considered that besides the positional data, the addition of velocity data was sufficient for the recognition.

They used a unidirectional LSTM network \cite{Dipietro2016RecognizingNetworks} to recognize the four workflow components. They trained the model using traditional cross-entropy loss and the Adam optimizer. They paid special attention to the selection of the following hyperparameters: sampling rate, learning rate, LSTM hidden dimension size, and the number of layers in the LSTM. They ran 5-fold cross-validation to obtain results from each of these hyperparameters. Then, they selected the best set of hyperparameters for the final training: 15Hz sampling rate, 1e\textsuperscript{-3} learning rate, 256 LSTM Hidden dimension, and 2 LSTM layers.

\subsubsection{MedAIR}
The MedAIR team (Yunshuang Li, Yonghao Long, and Qi Dou, Zhejiang University and the Chinese University of Hong Kong) participated in three tasks: video-based, kinematic-based, and video/kinematic-based workflow recognition. They resubmitted a Docker image for the video-based workflow recognition task.

The MedAIR team resized videos to $224\times224$ pixels and then augmented the data using a random horizontal flip and a random rotation of 5°. For kinematic data, they used a linear layer to obtain 2048 dimensions from the 28 dimensions to enrich the information.

For unimodal-based workflow recognition (video-based and kinematic-based tasks), the MedAIR team used a Trans-SVNet model \cite{Gao2021Trans-SVNet:Transformer}. First, they trained two different convolutional neural networks (CNN) to extract spatial features, one for steps and another for left and right verbs. Then, they trained three multi-stage temporal convolutional networks (TCN) to obtain temporal features for steps and verbs. Finally, they used three transformer layers to combine spatial and temporal features to obtain the final output for the three labels. Phases were not directly predicted by the networks, but identified based on the predicted step. They used a stochastic gradient descent (SGD) optimizer with a cross-entropy loss and a learning rate of 5e\textsuperscript{-4}.

For multimodal-based workflow recognition (video/kinematic-based task), they used a multi-modal relational graph network (MRG-Net) \cite{Long2020RelationalSurgery}. Like for unimodal-based workflow recognition, they used two CNNs to extract features from each frame in the video for steps and verbs. Then, they obtained the step labels using the original MRG-Net structure, which was the result of the fully connected layer with the output of three nodes in the graph. For the verb labels, the MedAIR team used fully connected layers to produce outputs $k^l_t$ and $k^r_t$, the final label prediction for left and right verb labels. They identified phases based on the predicted step. They used an Adam optimizer with cross-entropy loss and learning rate of 1e\textsuperscript{-4}.

\subsubsection{MMLAB}
The MMLAB team was composed of Satyadwyoom Kumar, Lalithkumar Seenivasan, and Hongliang Ren from the Netaji Subhas University of Technology, National University of Singapore, and the Chinese University of Hong Kong. They participated in the video/kinematic-based recognition task.

MMLAB team proposed a multi-task learning model to perform the recognition. First, each video frame was resized to $224\times224$ pixels. A ResNet 50 \cite{He2016DeepRecognition} pre-trained on ImageNet was used to extract visual features for each video frame. These features were passed with the frame-specific kinematic data through four label-specific networks (one per component). Each label-specific network was composed of two LSTMs \cite{Hochreiter1997LongMemory}, one for each modality, to capture the temporal features. The sequential length was set to 5, allowing the model to infer based on the current and past 4 temporal information sets. The resulting temporal features were then passed through a single linear layer for recognition. Each label-specific network was trained independently with cross-entropy loss, Adam optimizer, and a learning rate of 1e\textsuperscript{-3} for phase and step recognition, and 1e\textsuperscript{-2} for hand verbs.

\subsubsection{NCC NEXT}
The NCC NEXT team (Hiroki Matsuzaki, Yuto Ishikawa, Kazuyuki Hayashi, Yuriko Harai, and Nobuyoshi Takeshita, National Cancer Center Japan East Hospital) participated in all proposed tasks. They resubmitted a Docker image for all tasks except the kinematic-based recognition task.

They resized the initial video frames to a resolution of $512\times256$ pixels for video-based workflow recognition and of $480\times270$ pixels for segmentation-based workflow recognition. This was followed by normalization. They did not perform any preprocessing of kinematic data.

For video-based workflow recognition they used Xception networks \cite{Chollet2016Xception:Convolutions} pre-trained on ImageNet, one per component. They used the Radam optimizer \cite{Liu2019OnBeyond} with different learning rates with a batch size of 4, 1e\textsuperscript{-3} for phases and steps, and 1e\textsuperscript{-4} with a cosine decay scheduler for hand verbs. They also used cross-entropy loss.

For kinematic-based workflow recognition, the NCC NEXT used the light gradient boosting machine (LightGBM) framework \cite{Ke2017LightGBM:Tree}. Like for the previous task, they did the training and tuning of hyperparameters (i.e., learning rate, minimum data in leaf, number of iterations, and number of leaves) separately for each component (Table \ref{tab:NCC_hyperparameter}). They chose gradient boosting as a predictor optimizer and the mean absolute error (MAE) as loss of function.

\begin{table}[H]
\centering
\begin{tabular}{|l|c|c|c|c|}
\hline
Parameters & Phase & Step & Verb\_Left & Verb\_Right \\ \hline
Learning rate & 0.1 & 0.05 & 0.05 & 0.05 \\ \hline
min\_data\_in leaf & 9 & 9 & 3 & 9 \\ \hline
num\_iteration & 200 & 100 & 100 & 50 \\ \hline
num\_leaves & 11 & 31 & 11 & 11 \\ \hline
\end{tabular}
\caption{Hyperparameters for the kinematic based model developed by the NCC NEXT team}
\label{tab:NCC_hyperparameter}
\end{table}

The segmentation was performed by a Deeplabv3+ architecture \cite{Chen2018Encoder-decoderSegmentation} with an Xception backbone pre-trained on the Pascal visual object classes (PascalVOC) data set \cite{Everingham2009TheChallenge}. With the predicted segmentation, they trained a multi-output classification model, based on the EfficientNetB7 architecture \cite{Tan2019EfficientNet:Networks}, with Radam optimizer, cross-entropy loss function, a learning rate of 0.0001 with a cosine decay scheduler, and a batch size of 16.

For the multimodal workflow recognition tasks, the NCC NEXT team selected the method used in the three previous tasks that displayed the highest accuracy for each component. Specifically, for video/kinematic-based workflow recognition task, they used the video-based architecture for phase and step recognition and the kinematic-based architecture for hand verb recognition. For the video/kinematic/segmentation-based model, they used the video-based architecture for phase recognition, the segmentation-based architecture for step recognition, and the kinematic-based architecture for hand verb recognition.

\subsubsection{SK}
The SK team (Satoshi Kondo, Muroran Institute of Technology) participated in all proposed tasks.

For preprocessing, the SK team resized the images to $640\times353$ pixels and then used random shifting (maximum shift size of 10\% of the image size), scaling (0.9 to 1.1 times), rotation (-5 to 5 degrees), color jitter (-0.9 to 1.1 times for brightness, contrast, saturation, and hue), and Gaussian blurring (maximum sigma value = 1.0) for data augmentation. Finally, the images were normalized and the kinematic data were normalized in each dimension.

For the video-based workflow recognition task, the SK team used an 18-layer ResNet network \cite{He2016DeepRecognition}, pre-trained on ImageNet. The SK team omitted the final fully-connected layer of ResNet and fed its input 512-dimensional feature vector into two fully-connected layers to obtain a prediction of the step and hand verbs. Between these fully-connected layers, they inserted one ReLU and Dropout layers. The team used an Adam optimizer, with learning rate changes with cosine annealing with an initial value of 7.2e\textsuperscript{-4}, and a batch size of 96. The team optimized the initial learning rates for each task with the Optuna library \cite{Akiba2019Optuna:Framework}. The team chose cross-entropy loss as the loss function, with weights for each class depending on the class frequency for hand verbs. Phases were not directly predicted from the image, but identified based on the predicted step.

The SK team used a stacked LSTM \cite{Hochreiter1997LongMemory} with two layers and 28 hidden layers for the kinematic-based workflow recognition task. The LSTM output was fed into three fully connected layers as done for the previous task. The same optimizer and loss function were used. The initial learning rate was 1.5e\textsuperscript{-3} with a batch size of 6 and the number of data in a sequence was 30.

Image segmentation was done using the U-Net architecture \cite{Ronneberger2015U-Net:Segmentation} with ResNet18 as encoder with the summation of cross-entropy loss and dice loss. The SK team exploited the same model used for the video-based workflow recognition task and for the segmentation-based task. Both models were trained separately with an Adam optimizer and an initial learning rate of 2.4e\textsuperscript{-5} with a batch size of 32 for segmentation, and a learning rate of 1e\textsuperscript{-4} with a batch size of 6 for recognition.

For the video/kinematic-based task and video/kinematic/segmentation-based task, the SK team ensembled the previously trained dedicated modality networks to obtain a new prediction. As the SK team used the network parameters trained for the previous task, they did not train any network for these tasks.

\subsubsection{MediCIS: non-competing team}
The MediCIS team was a non-competing team due to the presence of challenge organizers (Quang-Minh Nguyen and Arnaud Huaulmé, University of Rennes 1). The team participated in all proposed tasks.

For the preprocessing step, they resized the frames to $256\times512$ pixels. Additionally, to train the segmentation model, they down-sampled the data to 6 Hz. They z-normalized the kinematic data.

For the video-based workflow recognition task, the MediCIS team used a hierarchical RestNet50 network \cite{He2016DeepRecognition} pre-trained on ImageNet to extract spatial features. Then, they used a Multi-Stage Temporal Convolutional Network called MS-TCN++ \cite{Li2020MS-TCN++:Segmentation}, with two stages, trained from scratch.

For the kinematic-based workflow recognition task, they directly used data as features for a two-stage MS-TCN++.

They selected as their segmentation model a U-Net \cite{Simonyan2014VeryRecognition} network trained from scratch with the Adam optimizer, cross-entropy loss, learning rate of 1e\textsuperscript{-4}, and batch size of 10. Like for the video-based task, workflow recognition was done by hierarchical ResNet50 followed by a two-stage MS-TCN++.

For the video/kinematic-based and video/kinematic/segmentation-based tasks, the MediCIS team extracted unimodal spatial features using a hierarchical ResNet50 network for video and segmentation data, followed by concatenation. Then, they trained a two-stage MS-TCN++.

They trained all workflow recognition models with the Adam optimizer, cross-entropy loss, learning rate of 1e\textsuperscript{-4}, and batch size of 2. For the hierarchical ResNet50 network, they emphasized the training for granularities that are harder to recognize using the following weights in the loss: 1 for phases, 2 for steps, and 5 for both action verbs. They set the number of dilated convolutional layers in MS-TCN++ to 10, except for the first layer where it was 11. The number of feature maps for each layer was 64.

\subsection{Workflow recognition results}

All results were computed on the organizers' hardware via the provided Docker images. This section only presents the results used for the ranking (balanced AD-Accuracy). Other results, such as application-dependent scores for each sequence and task, for each participating team, are available as supplementary material and at \url{www.synapse.org/PETRAW}.

\subsubsection{Task 1: Video-based workflow recognition}
Task 1 consisted of recognizing phases, steps, and hand verbs using video data only. Table \ref{tab:summary-task1} summarizes the algorithms used by the five teams that submitted models for this task.

\begin{table}[h]
    \centering
    \begin{tabular}{|l|c|c|c|c|c|}
    \hline
    Team & Hutom * & MedAIR * & NCC Next * & SK & MediCIS \\ \hline
    Preprocessing & X & X & X & X & X \\ \hline
    Augmentation & X & X &  & X &  \\ \hline
    Model & 3DResNet & Trans-SVNet & Xception & ResNet18 & \begin{tabular}[c]{@{}c@{}}ResNet50\\ \& MS-TCN++\end{tabular} \\ \hline
    Optimizer & Adam & SGD & Radam & Adam & Adam \\ \hline
    Loss & \begin{tabular}[c]{@{}c@{}}Equalization v2 \\ \& Normsoftmax\end{tabular} & cross-entropy & cross-entropy & cross-entropy &cross-entropy \\ \hline
    Learning Rate & 1e\textsuperscript{-3} & 5e\textsuperscript{-4} & \begin{tabular}[c]{@{}c@{}}1e\textsuperscript{-3}\\ \& 1e\textsuperscript{-4}\end{tabular} & 7.2e\textsuperscript{-4} & 1e\textsuperscript{-4} \\ \hline
    Causal & & & & & X \\\hline
    \end{tabular}
    \caption{Algorithms used for task 1. Teams that resubmitted models are highlighted with an asterisk. An “X” means that the method performed preprocessing, data augmentation, or is causal.}
    \label{tab:summary-task1}
\end{table}

\begin{figure}[H]
    \centering
    \includegraphics[width=\linewidth]{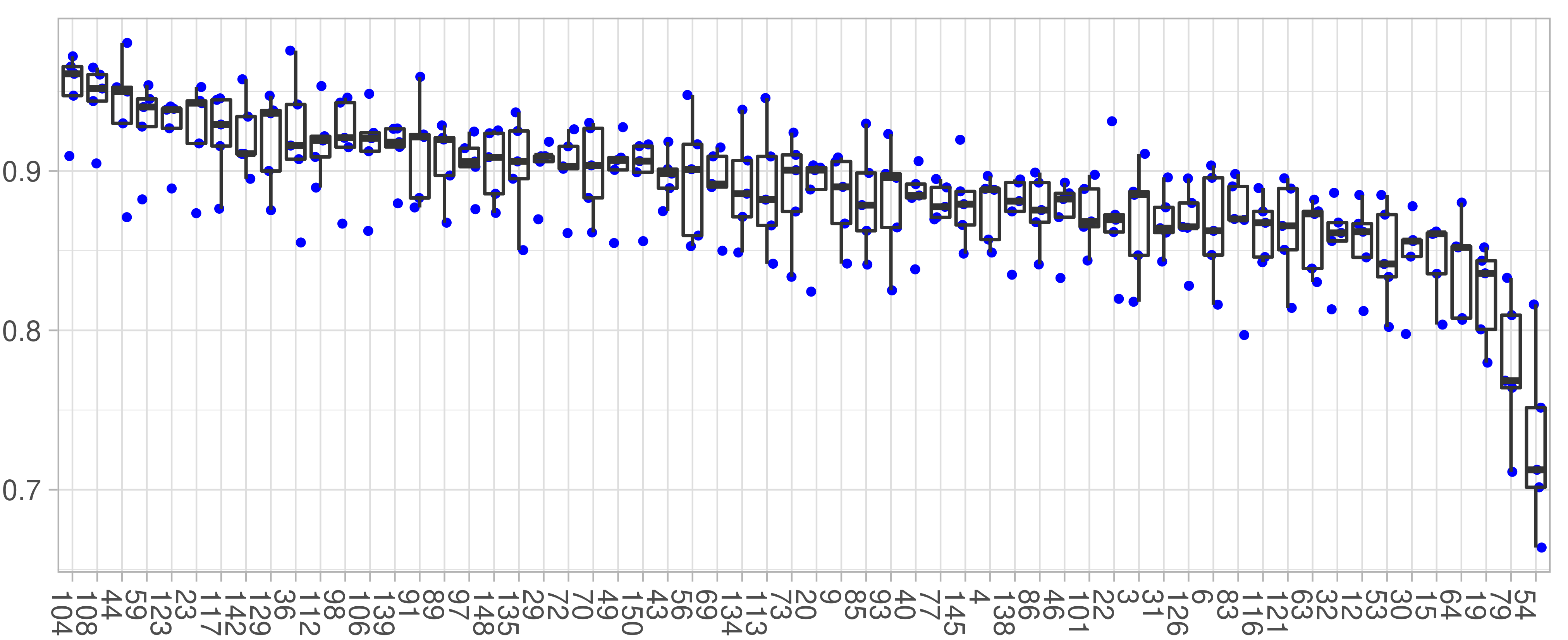}
    \caption{Task 1 recognition AD-Accuracy values (\%) for each sequence. Each dot represents the AD-Accuracy of one model. The x-axis represent the test sequence id. }
    \label{fig:T1_seq_results}
\end{figure}

Comparison of the mean AD accuracy values for each test sequence (all models) (Figure \ref{fig:T1_seq_results}) showed only a slight performance decrease (from 95.1\% to 82.2\%), but sequences 79 and 54 displayed the lowest performance (77.7\% and 72.9\%, respectively). Moreover, for all the test sequences, one model displayed lower AD-Accuracy values than the other models.

\begin{figure}[H]
    \centering
    \includegraphics[width=\linewidth]{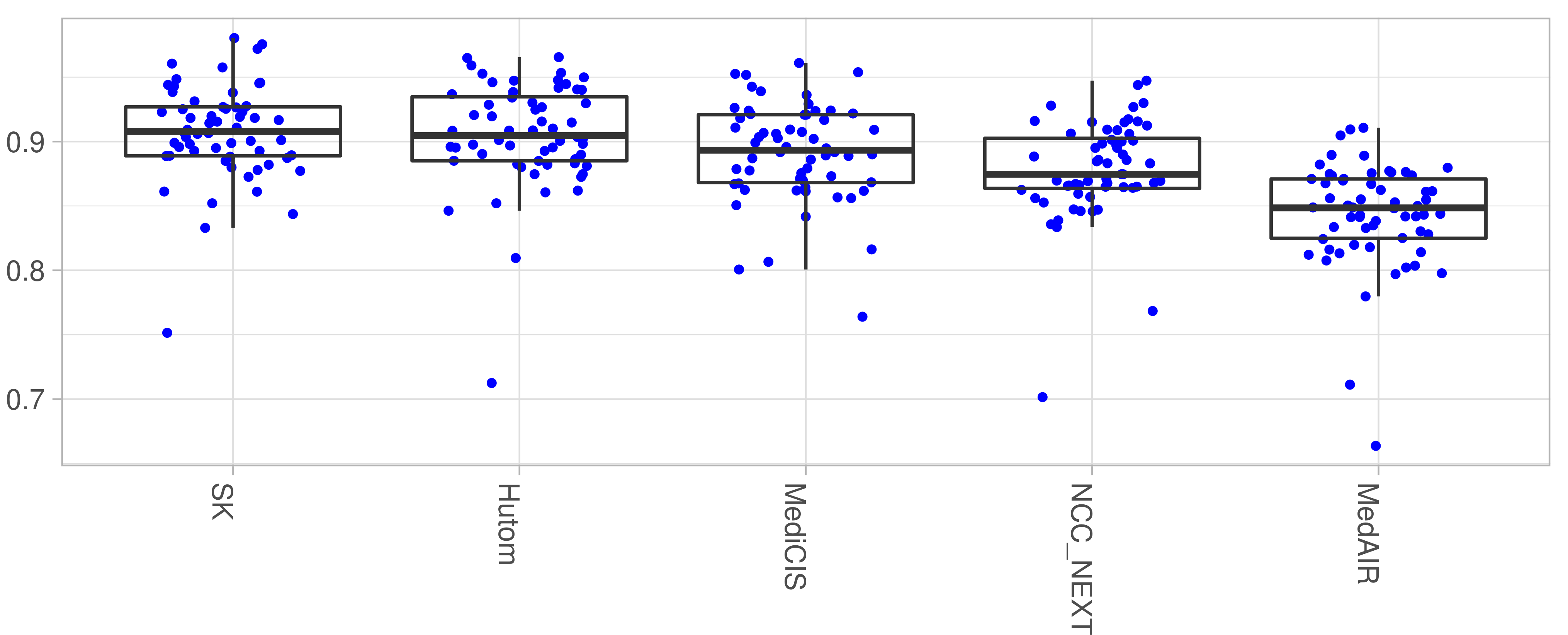}
    \caption{Mean task 1 recognition AD-Accuracy for each model. Each dot represents the AD-Accuracy for one sequence.}
    \label{fig:T1_Team_results}
\end{figure}

 Comparison of the mean AD-Accuracy value for each model (Figure \ref{fig:T1_Team_results} showed that team SK and team Hutom, obtained the highest values (>90\%), followed by team MediCIS and team NCC NEXT (>87\%). MedAIR obtained the lowest results ($\approx$84\%).

\begin{figure}[H]
    \centering
    \includegraphics[width=\linewidth]{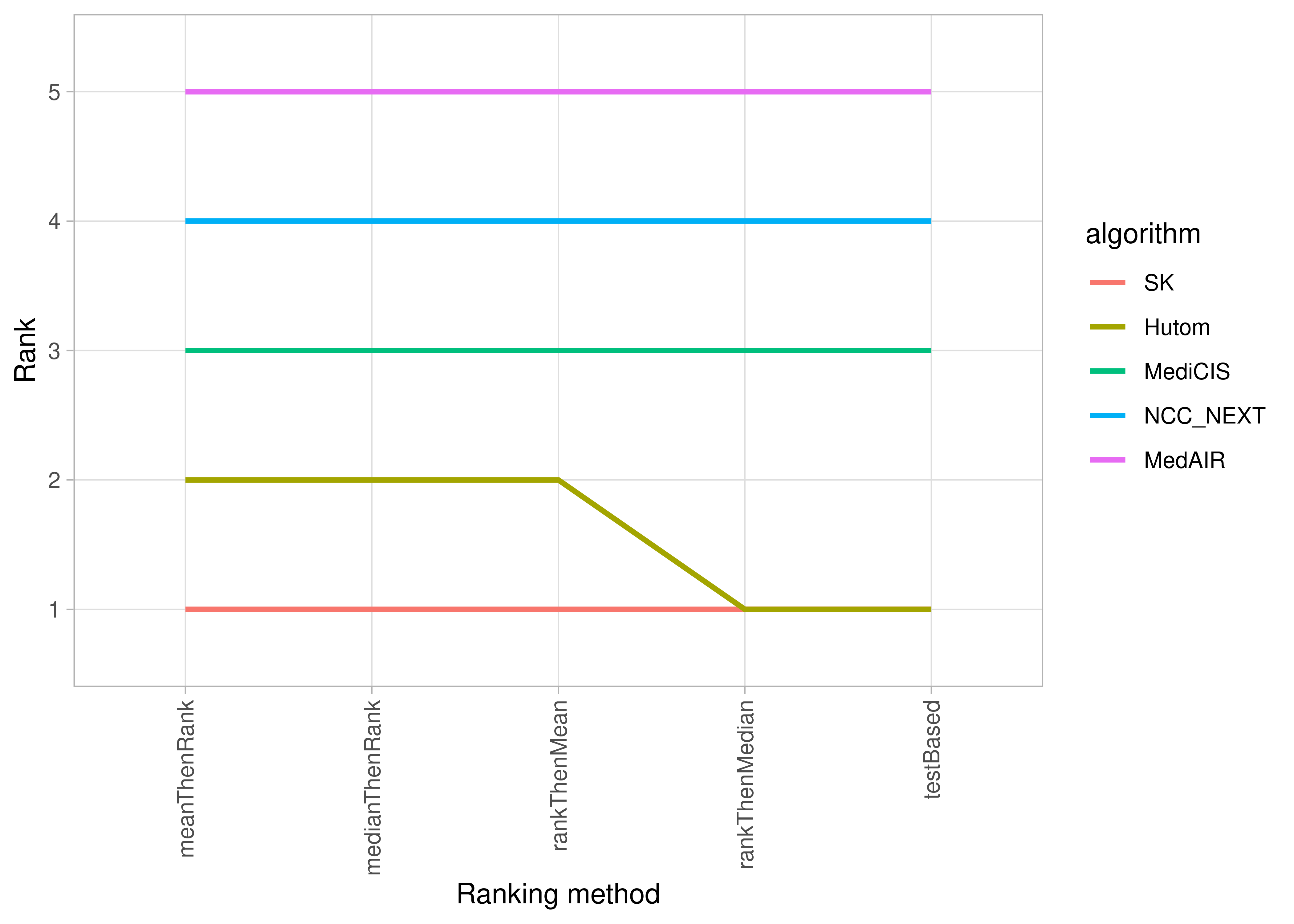}
    \caption{Task 1 recognition ranking stability using different ranking methods. Rank 1 indicates the best method.}
    \label{fig:T1_Team_rank}
\end{figure}

Team ranking was not influenced by the chosen method (Figure \ref{fig:T1_Team_rank}), except for the ranking of the SK and Hutom teams using the rankThenMedian and testBased methods.

\subsubsection{Task 2: Kinematic-based workflow recognition}
Task 2 consisted of recognizing phases, steps, and hand verbs using kinematic data only. Table \ref{tab:summary-task2} summarizes the methods used by the six participating teams for this task.

\begin{table}[h]
    \centering
    \begin{tabular}{|l|c|c|c|c|c|c|}
    \hline
    Team & Hutom & JHU-CIRL & MedAIR & NCC Next & SK & MediCIS \\ \hline
    Preprocessing & X & X & X & X & X & X \\ \hline
    Augmentation &  &  &  &  &  &  \\ \hline
    Model & Bi-LSTM & Uni-LSTM & \begin{tabular}[c]{@{}c@{}}Trans\\-SVNet \end{tabular}& LightGBM & \begin{tabular}[c]{@{}c@{}}Stacked\\-LSTM\end{tabular} & MS-TCN++ \\ \hline
    Optimizer & Adam & Adam & SGD & \begin{tabular}[c]{@{}c@{}}Gradient\\Boosting \end{tabular} & Adam & Adam \\ \hline
    Loss & \begin{tabular}[c]{@{}c@{}}Equalization v2 \\ \& Normsoftmax\end{tabular} & \begin{tabular}[c]{@{}c@{}}cross-\\entropy\end{tabular} & \begin{tabular}[c]{@{}c@{}}cross-\\entropy\end{tabular} & MAE & \begin{tabular}[c]{@{}c@{}}cross-\\entropy\end{tabular} & \begin{tabular}[c]{@{}c@{}}cross-\\entropy\end{tabular} \\ \hline
    Learning Rate & 1e\textsuperscript{-3} & 1e\textsuperscript{-3} &5e\textsuperscript{-4} & \begin{tabular}[c]{@{}c@{}}1e\textsuperscript{-1}\\ \& 5e\textsuperscript{-2}\end{tabular} & 1.5e\textsuperscript{-3} & 1e\textsuperscript{-4} \\ \hline
    Causal &  & X & & X & X & \\\hline
    \end{tabular}
    \caption{Summary of the models used for task 2. An “X” means that the method performed preprocessing, data augmentation, or is causal.}
    \label{tab:summary-task2}
\end{table}

\begin{figure}[H]
    \centering
    \includegraphics[width=\linewidth]{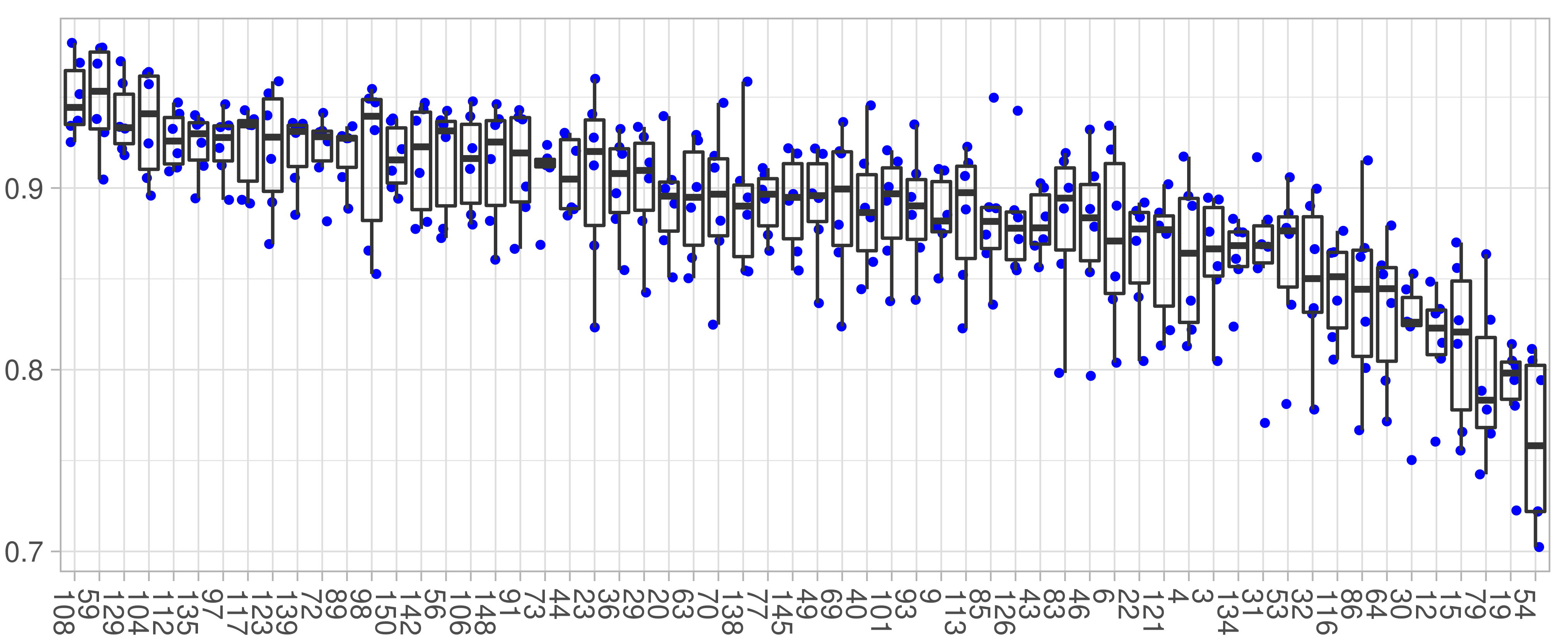}
    \caption{Task 2 recognition AD-Accuracy for each sequence. Each dot represents the AD-Accuracy of one model.}
    \label{fig:T2_seq_results}
\end{figure}

As with task 1, the performance per sequence slightly decreased (Figure \ref{fig:T2_seq_results}). The highest AD-Accuracy values were superior to 90\% for all teams. Three sequences (including sequences 79 and 54) had mean AD-Accuracy values inferior to 80\%. Unlike task 1, the majority of sequences did not have outliers.

\begin{figure}[H]
    \centering
    \includegraphics[width=\linewidth]{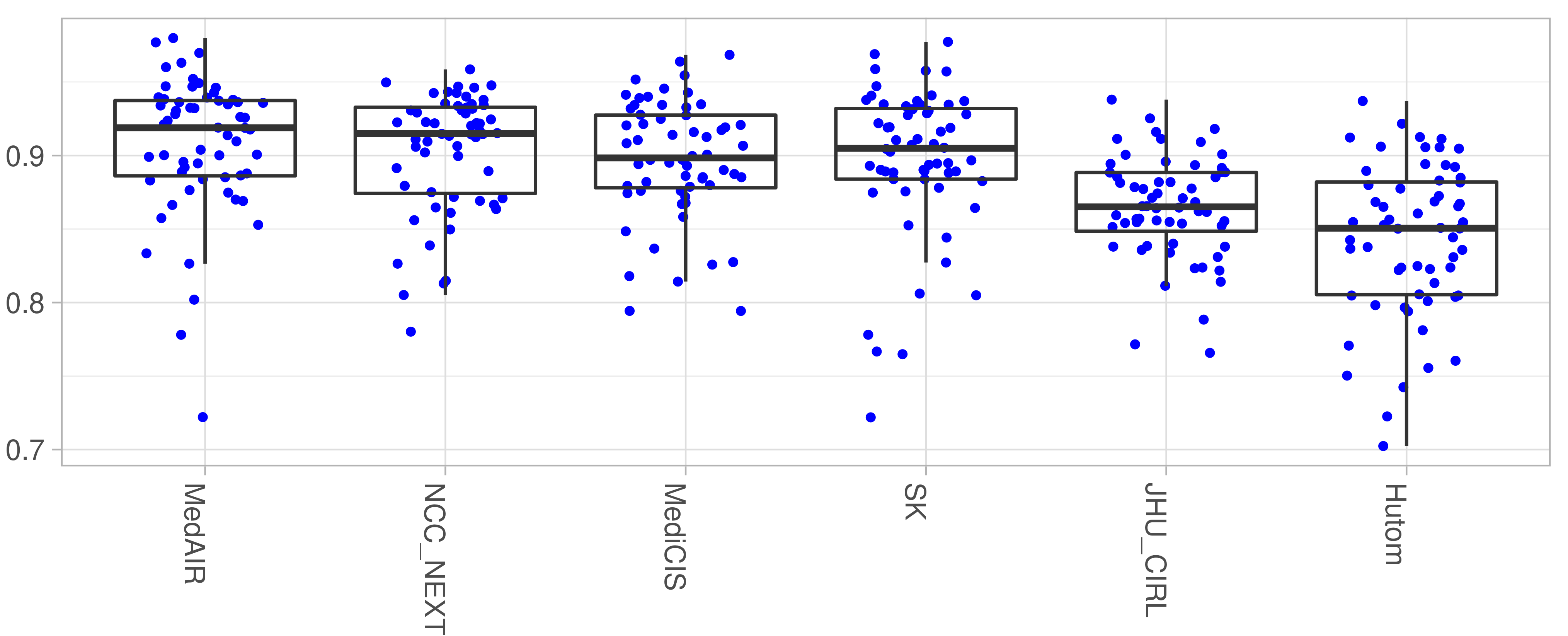}
    \caption{Mean task 2 recognition AD-Accuracy for each model. Each dot represents the AD-Accuracy for one sequence.}
    \label{fig:T2_Team_results}
\end{figure}

Results were very similar among teams (Figure \ref{fig:T2_Team_results}). Four had a mean AD-Accuracy value of between 89.7\% and 90.7\%, and the other two displayed mean AD-accuracy values of 86.4\% and 84.3\%, respectively.

\begin{figure}[H]
    \centering
    \includegraphics[width=\linewidth]{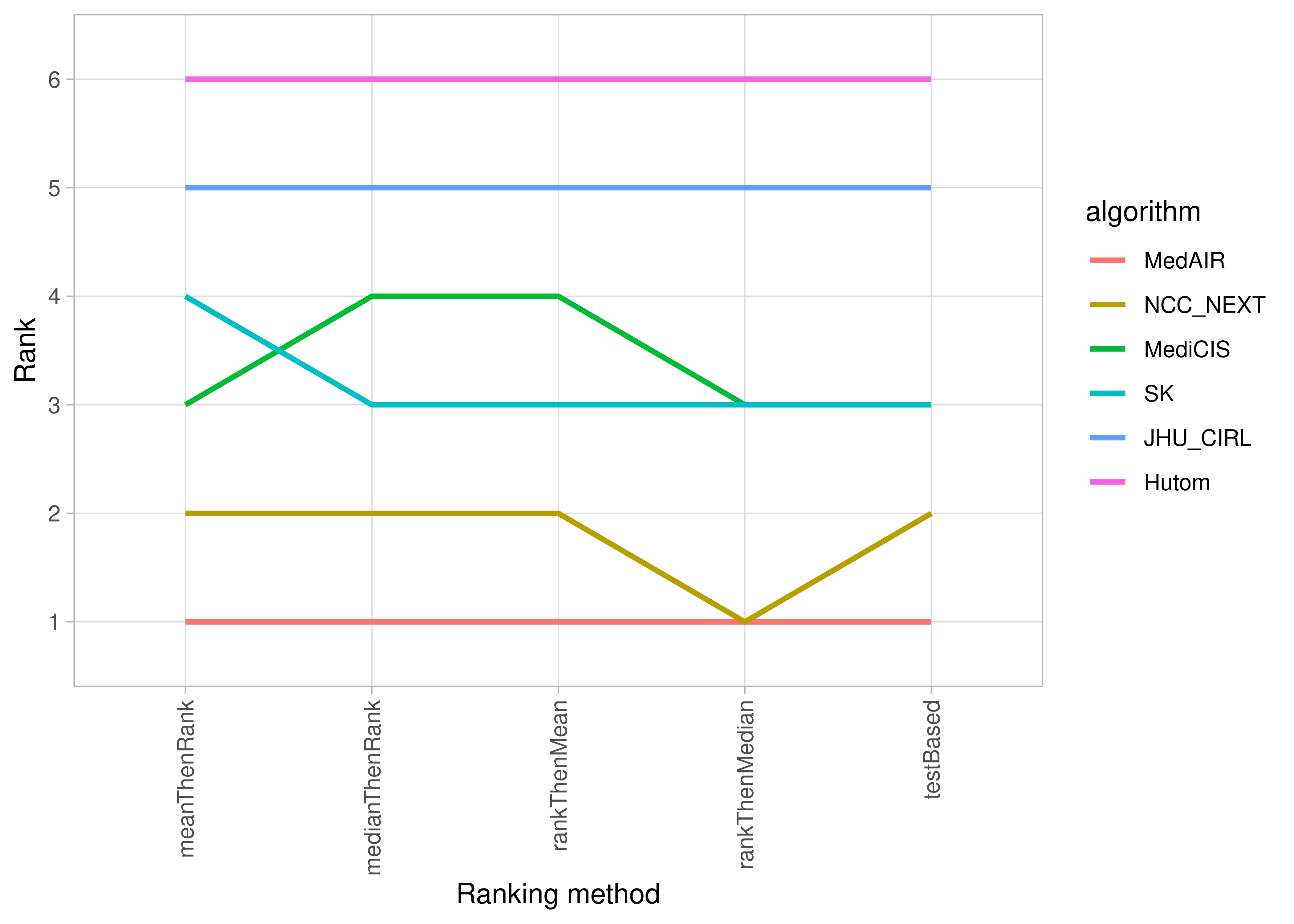}
    \caption{Task 2 recognition ranking stability using the indicated ranking methods.}
    \label{fig:T2_Team_rank}
\end{figure}

Ranking was not stable for team SK and team MediCIS (Figure \ref{fig:T2_Team_rank}). As MediCIS was a non-competing team, SK was ranked third for this task.

\subsubsection{Task 3: Segmentation-based workflow recognition}
Task 3 consisted of recognizing phases, steps, and hand verbs using semantic segmentation data only. First, the results of the segmentation models provided by the participants will be described, and then the workflow recognition models.

\textit{Segmentation models:}

Table \ref{tab:summary-task3-S} summarizes the methods used by the four participating teams to perform semantic segmentation.
\begin{table}[h]
    \centering
    \begin{tabular}{|l|c|c|c|c|}
    \hline
    Team & Hutom * & NCC Next * & SK & MediCIS \\ \hline
    Preprocessing & X & X & X & X \\ \hline
    Augmentation & X &  & X &  \\ \hline
    Model & DeepLabV3+ & DeepLabV3+ & U-Net & U-Net \\ \hline
    Optimizer & Adam & Radam & Adam & Adam \\ \hline
    Loss & \begin{tabular}[c]{@{}c@{}}Equalization v2 \\ \& Normsoftmax\end{tabular} & cross-entropy & cross-entropy & cross-entropy \\ \hline
    Learning Rate & 1e\textsuperscript{-3} & 1e\textsuperscript{-4} & 2.4e\textsuperscript{-5} & 1e\textsuperscript{-4} \\ \hline
    \end{tabular}
    \caption{Segmentation models used for task 3. Teams that resubmitted models are highlighted with an asterisk. An “X” means that the method performed preprocessing, data augmentation, or is causal.}
    \label{tab:summary-task3-S}
\end{table}

Comparison of the IoU values for each class independently and for all classes (Macro) (Table \ref{tab:task3_IoU_seq}) showed that, the IoU varied between 94.0\% and 91.1\% for Macro. Pegs were the least recognized structure (IoU between 83.9\% and 82.3\%). Specific sequences with lower performance were not identified.

\begin{figure}[h] 
    \centering
     \begin{subfigure}[t]{0.30\textwidth}
         \centering
         \includegraphics[width=\textwidth]{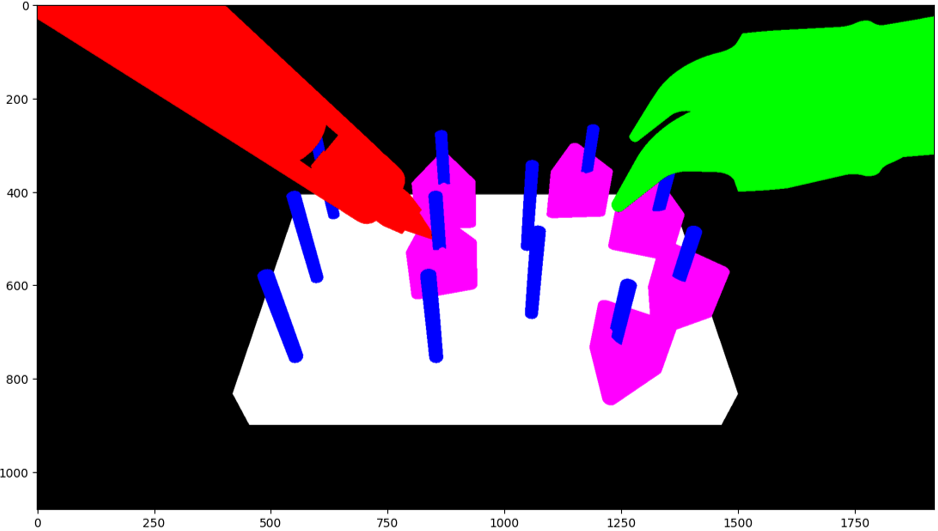}
         \caption{Ground Truth}
         \label{fig:seg-GT}
     \end{subfigure}
     \hfill
     \begin{subfigure}[t]{0.3\textwidth}
         \centering
         \includegraphics[width=\textwidth]{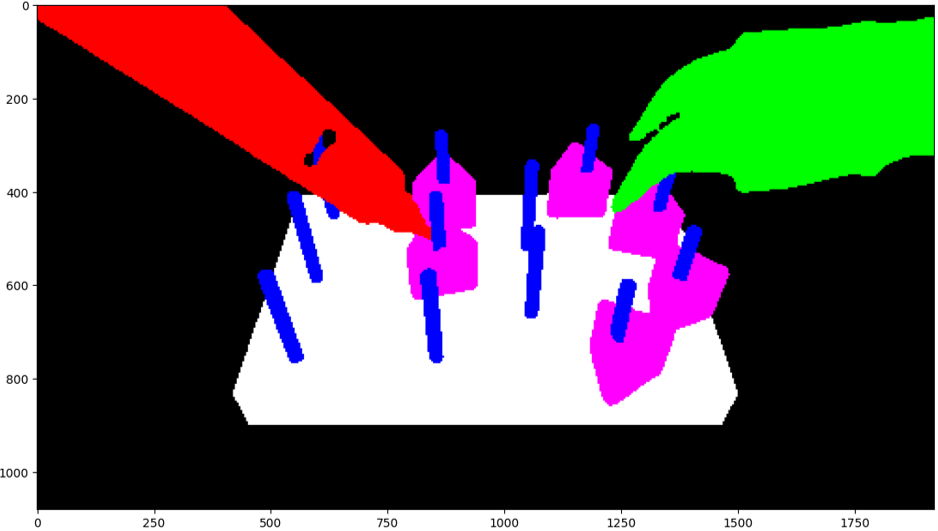}
         \caption{NCC Next}
         \label{fig:seg-NCC}
     \end{subfigure}
     \begin{subfigure}[t]{0.3\textwidth}
         \centering
         \includegraphics[width=\textwidth]{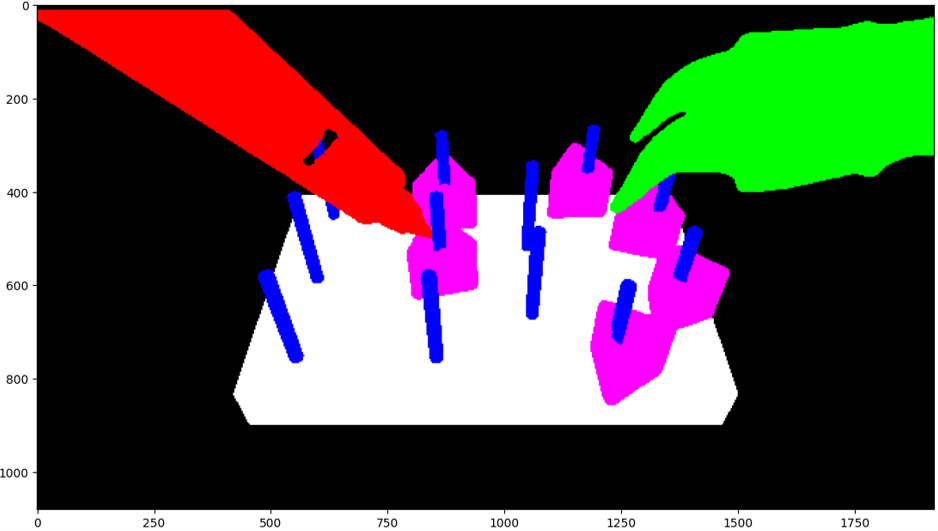}
         \caption{SK}
         \label{fig:seg-SK}
     \end{subfigure}
     \\
     \hfill
     \begin{subfigure}[t]{0.3\textwidth}
         \centering
         \includegraphics[width=\textwidth]{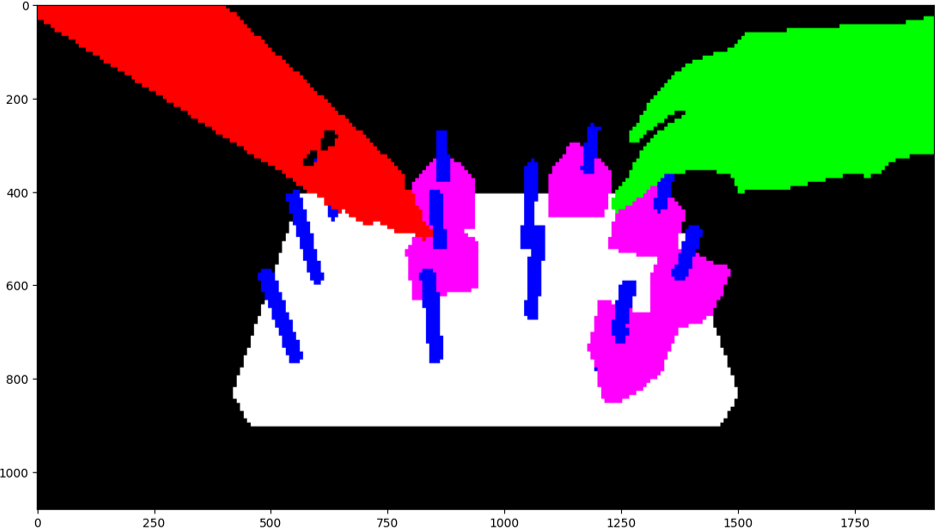}
         \caption{Medicis}
         \label{fig:seg-Medicis}
     \end{subfigure}
     \begin{subfigure}[t]{0.3\textwidth}
         \centering
         \includegraphics[width=\textwidth]{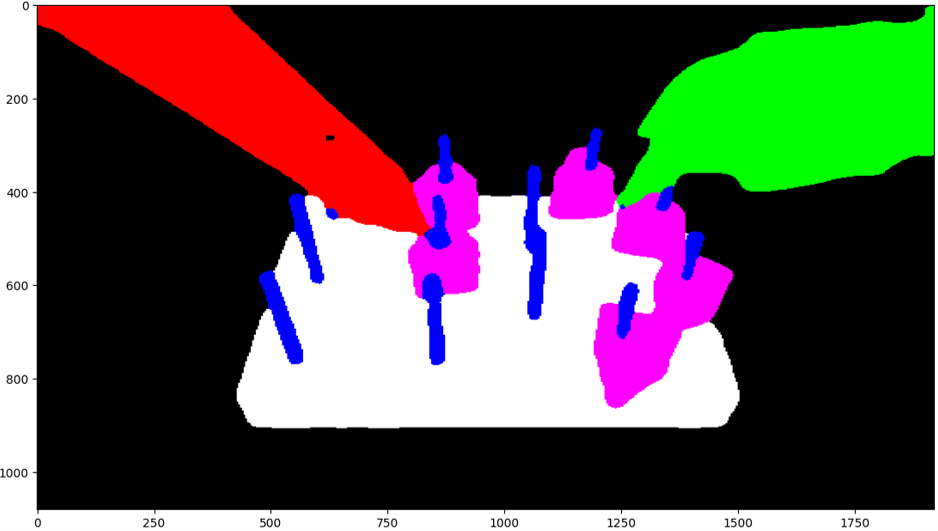}
         \caption{Hutom}
         \label{fig:seg-Hutom}
     \end{subfigure}
     \caption{Ground truth (\subref{fig:seg-GT}) and segmentation results for each team (\subref{fig:seg-NCC} to \subref{fig:seg-Hutom}) for one frame.}
     \label{fig:seg-Results}
\end{figure}

Comparison of the mean IoU values of each team for all classes (Macro) and for each class independently (Table \ref{tab:task3_IoU_team}) showed similar Macro results for the NCC Next, SK and MediCIS teams ( 96.9\%, 96.4\%, and 94.0\%, respectively). The Hutom team's Macro IoU was the lowest (85.0\%), mainly due to the IoU for pegs (63.3\%). Figure \ref{fig:seg-Results} presents the ground truth and the segmentation results of each team for one frame.

\begin{table}[h]
\centering
\begin{tabular}{|c|c|c|c|c|}
\hline
           & Mean & Median & Max  & Min \\ \hline
Background & 98.8 & 98.9   & 98.9 & 98.7 \\ \hline
Base       & 96.1 & 96.2   & 96.3 & 95.6 \\ \hline
Pegs       & 83.2 & 83.1   & 83.9 & 82.3 \\ \hline
Blocks     & 91.7 & 91.7   & 92.5 & 90.8 \\ \hline
Left tool  & 94.9 & 95.3   & 97.6 & 87.3 \\ \hline
Right tool & 94.0 & 94.5   & 96.9 & 88.9 \\ \hline
Macro      & 93.1 & 93.2   & 94.0 & 91.1 \\ \hline
\end{tabular}
\caption{Mean Intersection-Over-Union values for all classes of each sequence independently}
\label{tab:task3_IoU_seq}
\end{table}

\begin{table}[h]
\centering
\begin{tabular}{|c|c|c|c|c|}
\hline
           & Hutom * & NCC Next *    & SK            & MediCIS \\ \hline
Background & 97.7    & \textbf{99.5} & 99.2          & 98.9    \\ \hline
Base       & 91.4    & \textbf{98.4} & 98.4          & 96.1    \\ \hline
Pegs       & 63.3    & \textbf{92.1} & 92.0          & 85.3    \\ \hline
Blocks     & 82.8    & 96.0          & \textbf{96.0} & 92.2    \\ \hline
Left tool  & 89.3    & \textbf{98.1} & 96.1          & 96.0    \\ \hline
Right tool & 85.5    & \textbf{97.8} & 96.7          & 95.8    \\ \hline
Macro      & 85.0    & \textbf{96.9} & 96.4          & 94.0    \\ \hline
\end{tabular}
\caption{Mean Intersection-Over-Union values for all the classes of each team. Teams that resubmitted models are highlighted with an asterisk and best results are in bold.}
\label{tab:task3_IoU_team}
\end{table}

\textit{Workflow models}

Table \ref{tab:summary-task3-W} summarizes the methods used by the four participating teams to perform the workflow recognition.
\begin{table}[h]
    \centering
    \begin{tabular}{|l|c|c|c|c|}
    \hline
    Team & Hutom * & NCC Next * & SK & MediCIS \\ \hline
    Preprocessing & X & X & X & X \\ \hline
    Augmentation & X &  & X &  \\ \hline
    Model W & SlowFast50 & EfficientNetB7 & ResNet18 & \begin{tabular}[c]{@{}c@{}}ResNet50\\ \& MS-TCN++\end{tabular} \\ \hline
    Optimizer & Adam & Radam & Adam & Adam \\ \hline
    Loss & \begin{tabular}[c]{@{}c@{}}Equalization v2 \\ \& Normsoftmax\end{tabular} & cross-entropy & cross-entropy & cross-entropy \\ \hline
    Learning Rate & 1e\textsuperscript{-3} & 1e\textsuperscript{-4} & 1e\textsuperscript{-4} & 1e\textsuperscript{-4} \\ \hline
    Causal &  &  & & X\\\hline
    \end{tabular}
    \caption{Summary of the models used for task 3 (segmentation-based workflow recognition). Teams that resubmitted models are highlighted with an asterisk. An “X” means that the method performed preprocessing, data augmentation, or is causal.}
    \label{tab:summary-task3-W}
\end{table}

\begin{figure}[H]
    \centering
    \includegraphics[width=\linewidth]{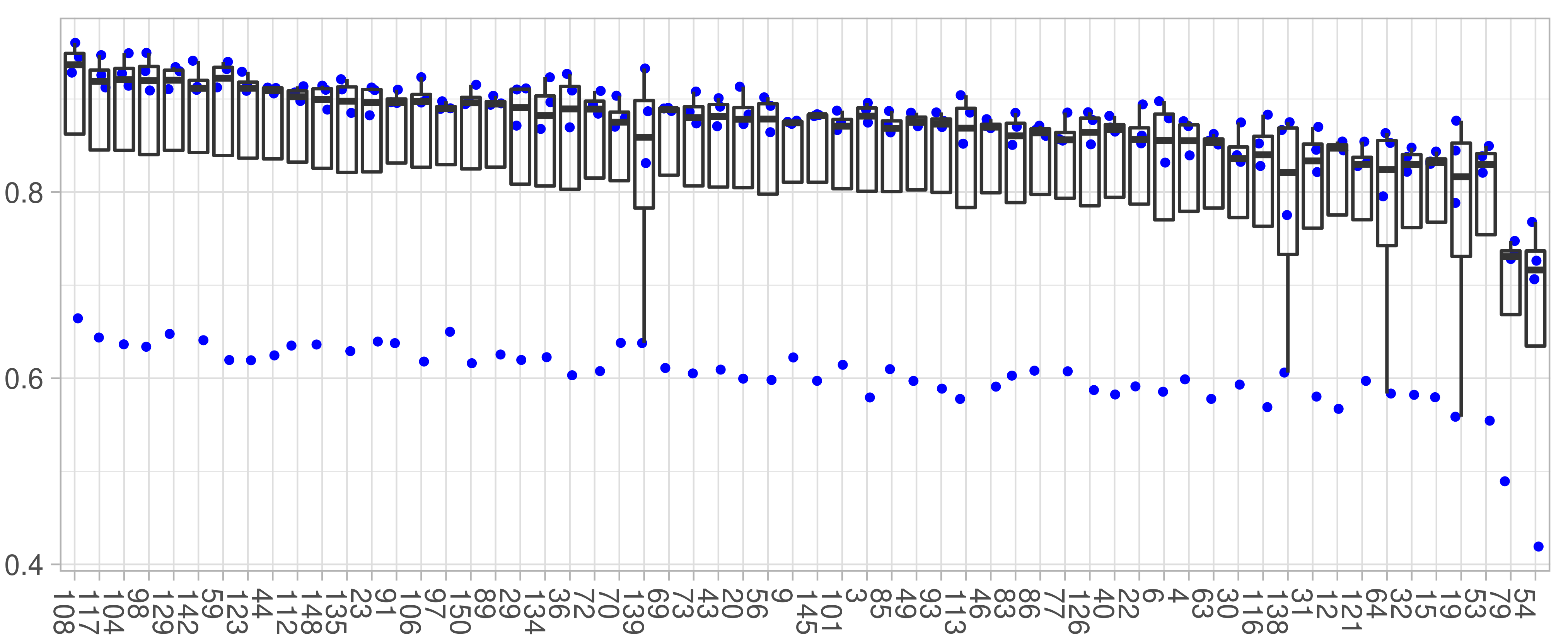}
    \caption{Task 3 recognition AD-Accuracy for each sequence. Each dot represents the AD-Accuracy of one model.}
    \label{fig:T3_seq_results_W}
\end{figure}

Comparison of the mean AD-Accuracy values for each test sequence (Figure \ref{fig:T3_seq_results_W}) showed that performance decreased from 87.5\% to 76.6\%. The same two sequences (79 and 54) displayed very low results (67.4\% and 65.5\%, respectively). Moreover, for all test cases, one model had results lower than 70\%.

\begin{figure}[H]
    \centering
    \includegraphics[width=\linewidth]{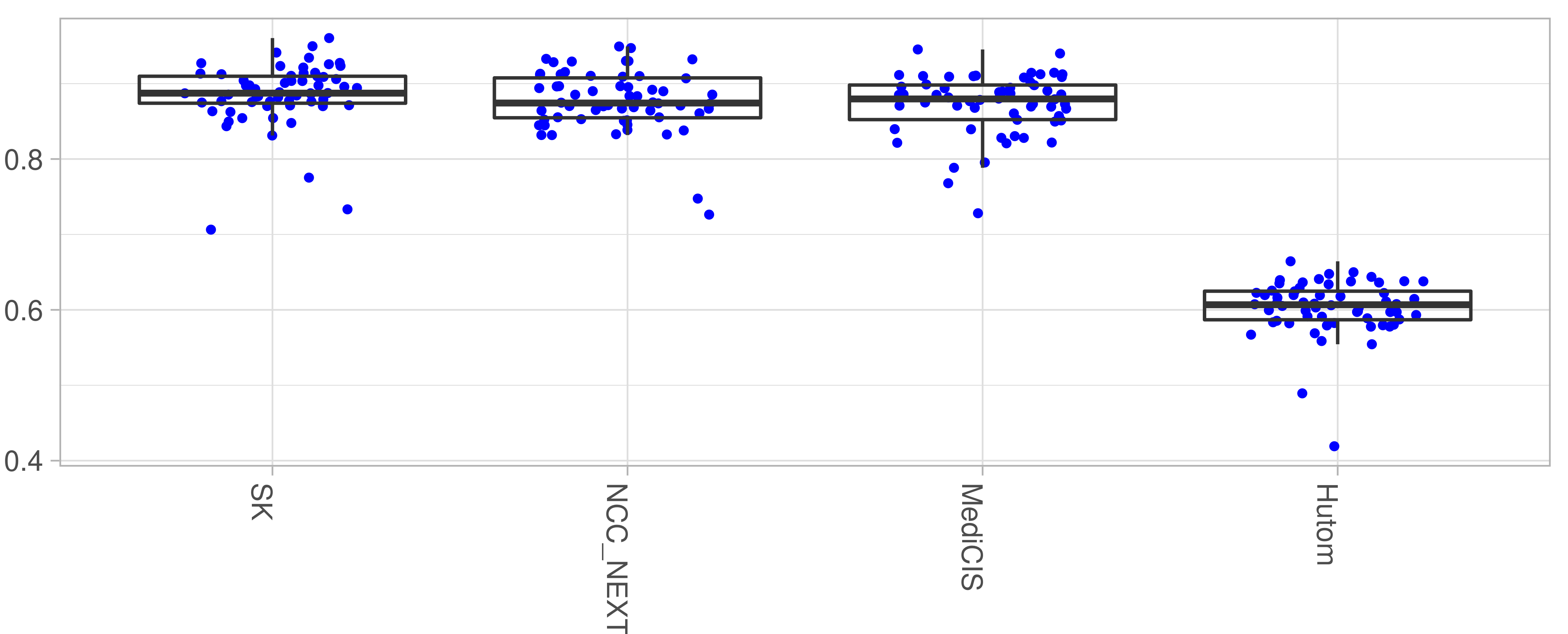}
    \caption{Mean recognition AD-Accuracy for each model for task 3. Each dot represents the AD-Accuracy for one sequence.}
    \label{fig:T3_Team_results_W}
\end{figure}

Comparison of the mean AD-Accuracy value for each model indicated that three teams obtained results between 88.5\% and 87.2\%, whereas the Hutom team had a mean AD-Accuracy value of 60.3\% (Figure \ref{fig:T3_Team_results_W}).

\begin{figure}[H]
    \centering
    \includegraphics[width=\linewidth]{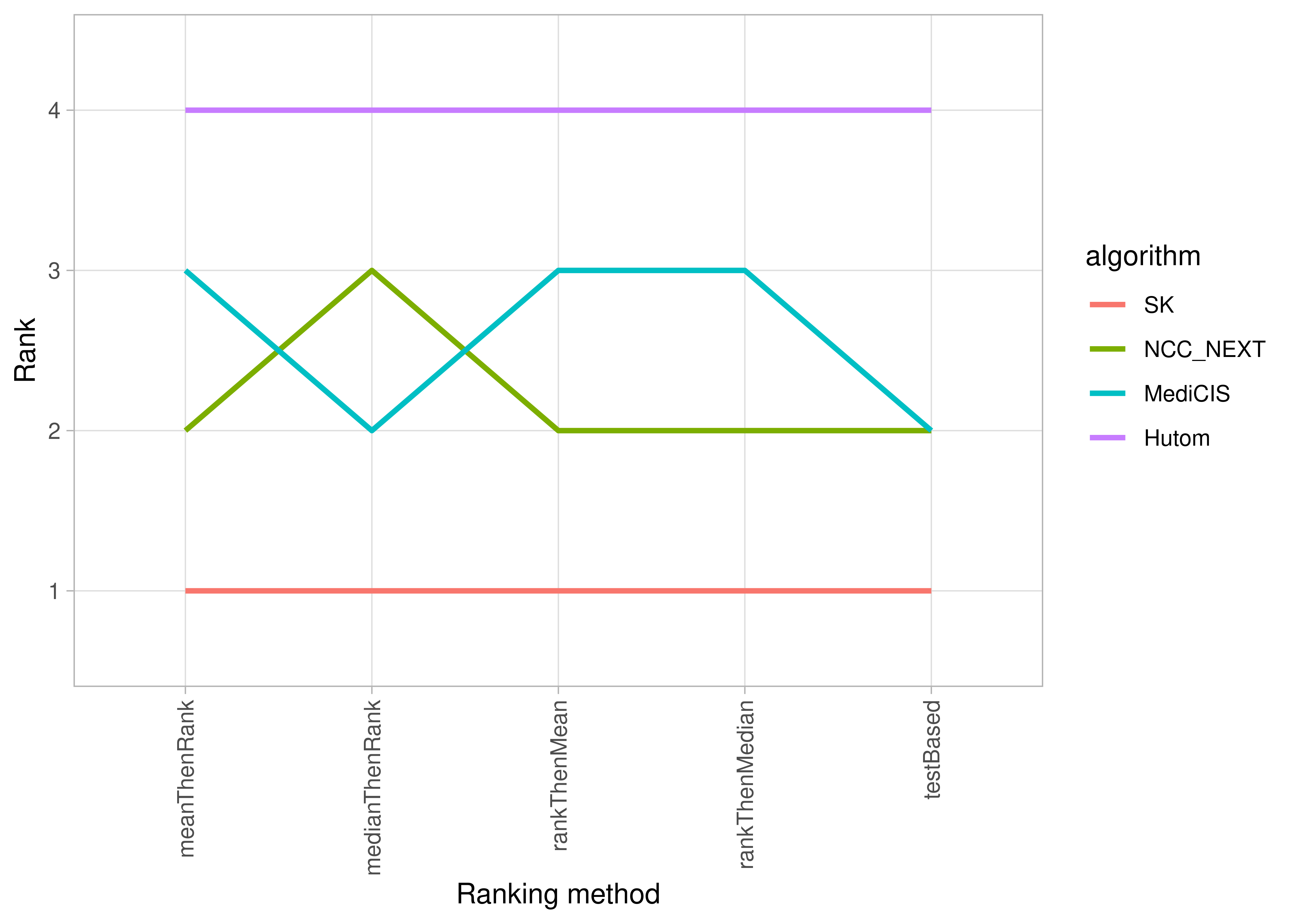}
    \caption{Task 3 recognition ranking stability using the indicated ranking methods.}
    \label{fig:T3_Team_rank_W}
\end{figure}

The choice of method did not influence the team ranking, except for the second (NCC NEXT) and the third (MediCIS) rank (Figure \ref{fig:T3_Team_rank_W}).

\subsubsection{Task 4: Video/kinematic-based workflow recognition}
Task 4 consisted of recognizing phases, steps, and hand verbs using video and kinematic data. Table \ref{tab:summary-task4} summarizes the methods used by the six participating teams.

\begin{table}[h]
    \centering
    \begin{tabular}{|l|c|c|c|c|c|c|}
    \hline
    Team & Hutom * & MedAIR & MMLAB & NCC NEXT * & SK & MediCIS \\ \hline
    Preprocessing & X & X & X & X & X & X \\ \hline
    Augmentation & X & X &  &  & X &  \\ \hline
    Model & \begin{tabular}[c]{@{}c@{}}3D ResNet \\ \& Bi-LSTM\end{tabular} & \begin{tabular}[c]{@{}c@{}}MRG-Net\\ \& CNN\end{tabular} & \begin{tabular}[c]{@{}c@{}}ResNet50\\ \& LSTM\end{tabular} & \begin{tabular}[c]{@{}c@{}}Xception \\ \& LightGBM\end{tabular} & \begin{tabular}[c]{@{}c@{}}ResNet18\\ \& Stacked\\-LSTM\end{tabular} & \begin{tabular}[c]{@{}c@{}}ResNet50\\ \& MS-TCN++\end{tabular} \\ \hline
    Optimizer & Adam & Adam & Adam & \begin{tabular}[c]{@{}c@{}}Radam\\ \& Gradient \\ Boosting\end{tabular} & Adam & Adam \\ \hline
    Loss & \begin{tabular}[c]{@{}c@{}}Equalization \\v2  \& \\Normsoftmax\end{tabular} & \begin{tabular}[c]{@{}c@{}}cross-\\entropy\end{tabular} & \begin{tabular}[c]{@{}c@{}}cross-\\entropy\end{tabular} & \begin{tabular}[c]{@{}c@{}}cross-\\entropy \\ \& MAE\end{tabular} & \begin{tabular}[c]{@{}c@{}}cross-\\entropy\end{tabular} & \begin{tabular}[c]{@{}c@{}}cross-\\entropy\end{tabular} \\ \hline
    Learning Rate & 1e\textsuperscript{-3} & 1e\textsuperscript{-4} & \begin{tabular}[c]{@{}c@{}}1e\textsuperscript{-3} \\ \& 1e\textsuperscript{-2}\end{tabular} & \begin{tabular}[c]{@{}c@{}}1e\textsuperscript{-1} \& 5e\textsuperscript{-2} \\\& 1e\textsuperscript{-3} \& 1e\textsuperscript{-4}\end{tabular} &  \begin{tabular}[c]{@{}c@{}}7.2e\textsuperscript{-4} \\\& 1.5e\textsuperscript{-3}\end{tabular} & 1e\textsuperscript{-4} \\ \hline
    Causal & & & X &  & &\\\hline
    \end{tabular}
    \caption{Summary of the models used for task 4. Teams that resubmitted models are highlighted with an asterisk. An “X” means that the method performed preprocessing, data augmentation, or is causal.}
    \label{tab:summary-task4}
\end{table}

\begin{figure}[H]
    \centering
    \includegraphics[width=\linewidth]{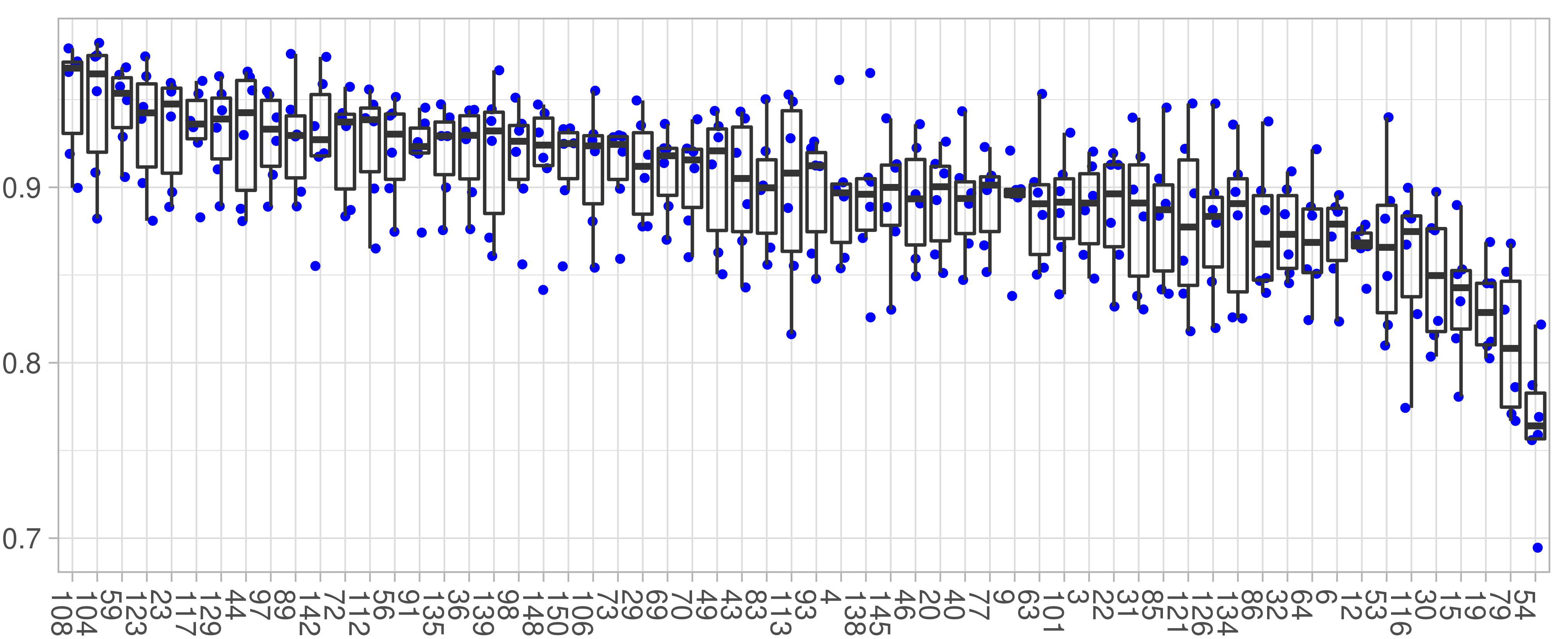}
    \caption{Task 4 recognition AD-Accuracy values for each sequence. Each dot represents the AD-Accuracy for one model.}
    \label{fig:T4_seq_results}
\end{figure}

AD-Accuracy values for each sequence were similar to those of the previous tasks (Figure \ref{fig:T4_seq_results}). Indeed, performance slightly decreased from 95.1\% to 83.1\% for most sequences, and was again low for sequences 79 and 54 (81.2\% and 76.5\%). For this task, the number of outliers was limited.

\begin{figure}[H]
    \centering
    \includegraphics[width=\linewidth]{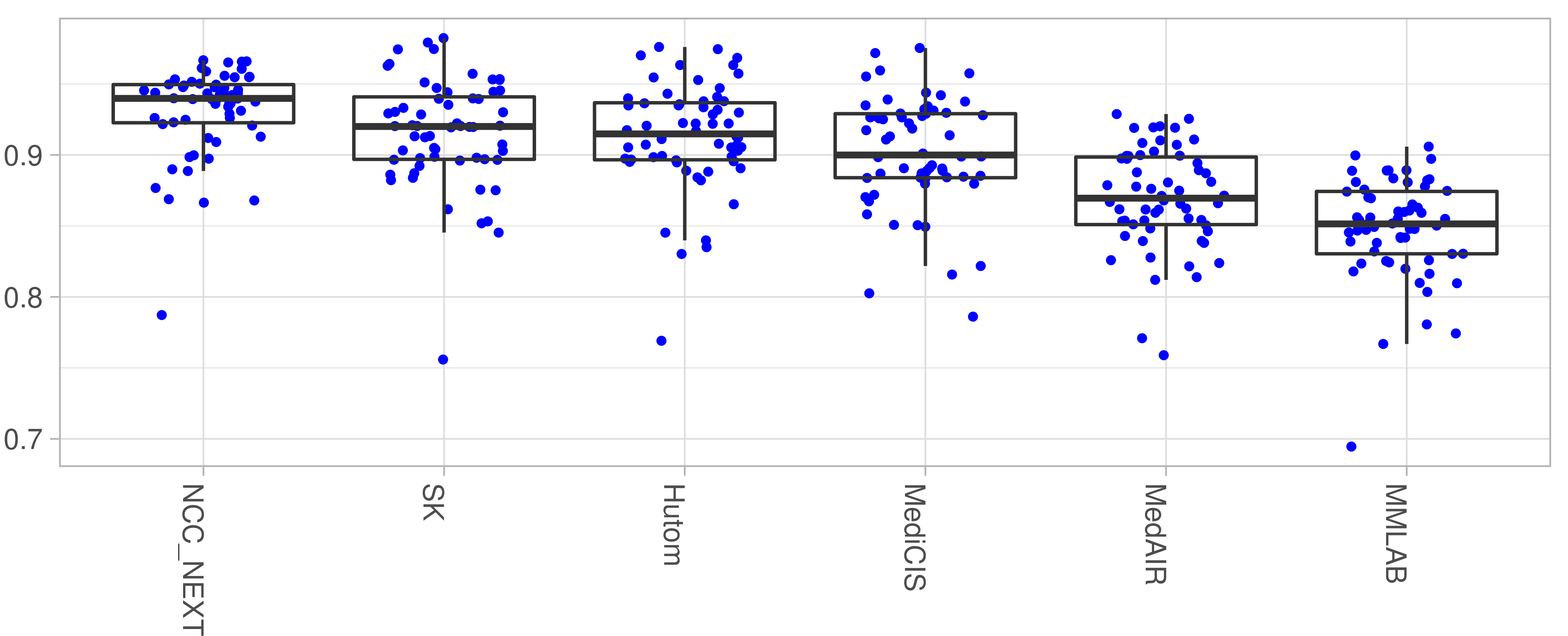}
    \caption{Mean task 4 recognition AD-Accuracy for each team. Each dot represents the AD-Accuracy for one sequence.}
    \label{fig:T4_Team_results}
\end{figure}

The NCC NEXT team obtained the best results (Figure \ref{fig:T4_Team_results}), with a mean AD-Accuracy of 93.1\%, followed by SK, Hutom, and MediCIS teams with results of between 91.6\% and 90.2\%. For the last two teams, the AD-Accuracy was above 84.5\%.

\begin{figure}[H]
    \centering
    \includegraphics[width=\linewidth]{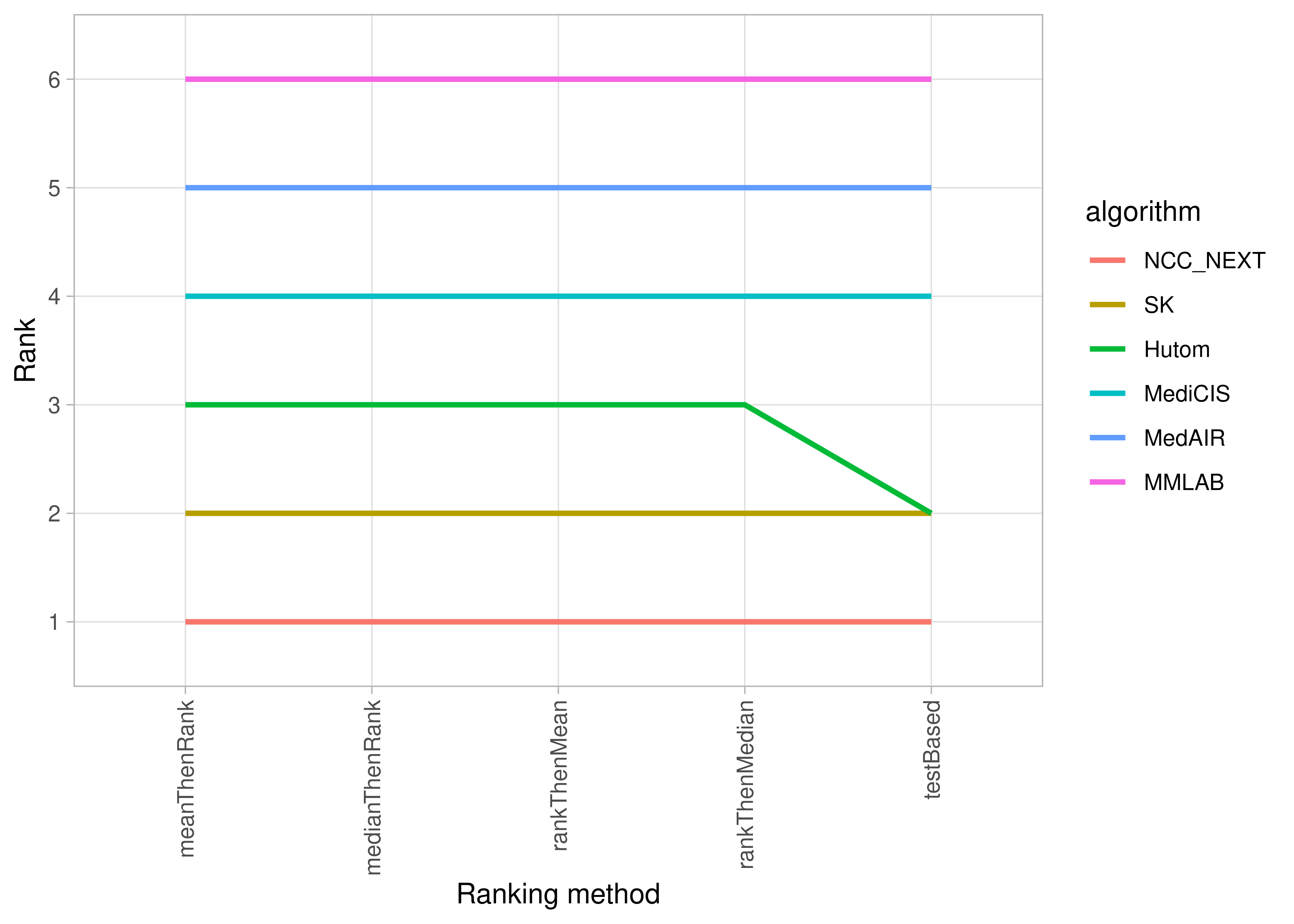}
    \caption{Task 4 recognition ranking stability using the indicated ranking methods.}
    \label{fig:T4_Team_rank}
\end{figure}

The ranking is stable according to the ranking method chosen (Figure \ref{fig:T4_Team_rank}). 

\subsubsection{Task 5: Video/kinematic/segmentation-based workflow recognition}
In task 5, teams recognized phases, steps, and hand verbs using video, kinematic and segmentation data. Table \ref{tab:summary-task5} summarizes the recognition methods used by the four participating teams. The models to create the segmentation were the same as those described in Table \ref{tab:summary-task3-S}.

\begin{table}[h]
    \centering
    \begin{tabular}{|l|c|c|c|c|}
    \hline
    Team & Hutom * & NCC NEXT * & SK & MediCIS \\ \hline
    Preprocessing & X & X & X & X \\ \hline
    Augmentation & X &  & X &  \\ \hline
    Model & \begin{tabular}[c]{@{}c@{}}3D ResNet\\ \& Bi-LSTM\end{tabular} & \begin{tabular}[c]{@{}c@{}}Xception, \\ EfficientNetB7\\\& LightGBM\end{tabular} & \begin{tabular}[c]{@{}c@{}}ResNet18\\ \& Staked-LSTM\end{tabular} & \begin{tabular}[c]{@{}c@{}}ResNet50\\ \& MS-TCN++\end{tabular} \\ \hline
    Optimizer & Adam &  \begin{tabular}[c]{@{}c@{}}Radam\\\& Gradient Boosting\end{tabular} & Adam & Adam \\ \hline
    Loss & \begin{tabular}[c]{@{}c@{}}Equalization v2 \\ \& Normsoftmax\end{tabular} & \begin{tabular}[c]{@{}c@{}}cross-entropy \\ \& MAE\end{tabular} & cross-entropy & cross-entropy \\ \hline
    Learning Rate & 1e\textsuperscript{-3} & \begin{tabular}[c]{@{}c@{}}1e\textsuperscript{-1}, 5e\textsuperscript{-2},\\ 1e\textsuperscript{-3} \& 1e\textsuperscript{-4}\end{tabular} & \begin{tabular}[c]{@{}c@{}} 7.2e\textsuperscript{-4},\\  1.5e\textsuperscript{-3} \& 1e\textsuperscript{-4}\end{tabular} & 1e\textsuperscript{-4} \\ \hline
    Causal& & & &\\\hline
    \end{tabular}
    \caption{Models used for task 5. Teams that resubmitted models are highlighted with an asterisk. An “X” means that the method performed preprocessing, data augmentation, or is causal.}
    \label{tab:summary-task5}
\end{table}

\begin{figure}[H]
    \centering
    \includegraphics[width=\linewidth]{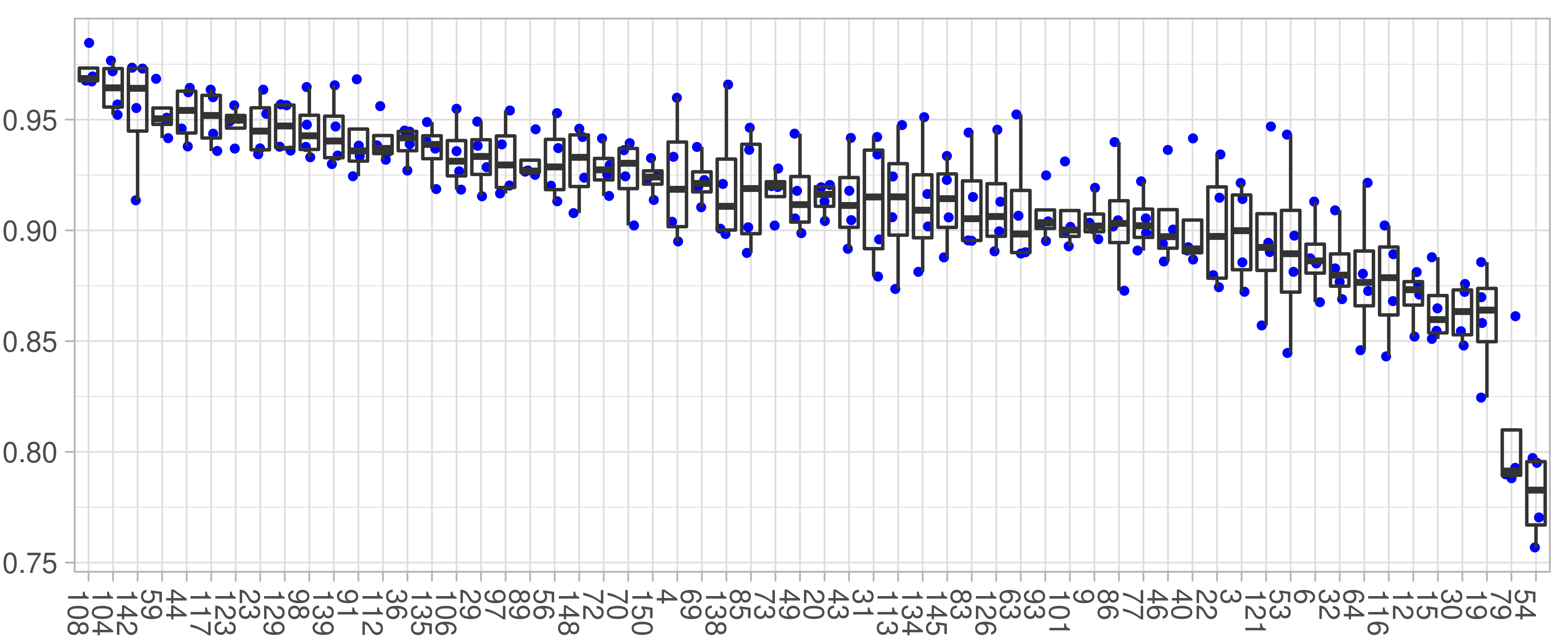}
    \caption{Task 5 AD-Accuracy for each sequence. Each dot represents the AD-Accuracy for one model.}
    \label{fig:T5_seq_results}
\end{figure}

As for the previous tasks, the mean AD-Accuracy values per sequence (Figure \ref{fig:T5_seq_results}) highlighted a slight performance decrease (from 97.2\% to 85.9\%). Sequences 79 and 54 again displayed the lowest performances (80.8\% and 78.0\%, respectively). 

\begin{figure}[H]
    \centering
    \includegraphics[width=\linewidth]{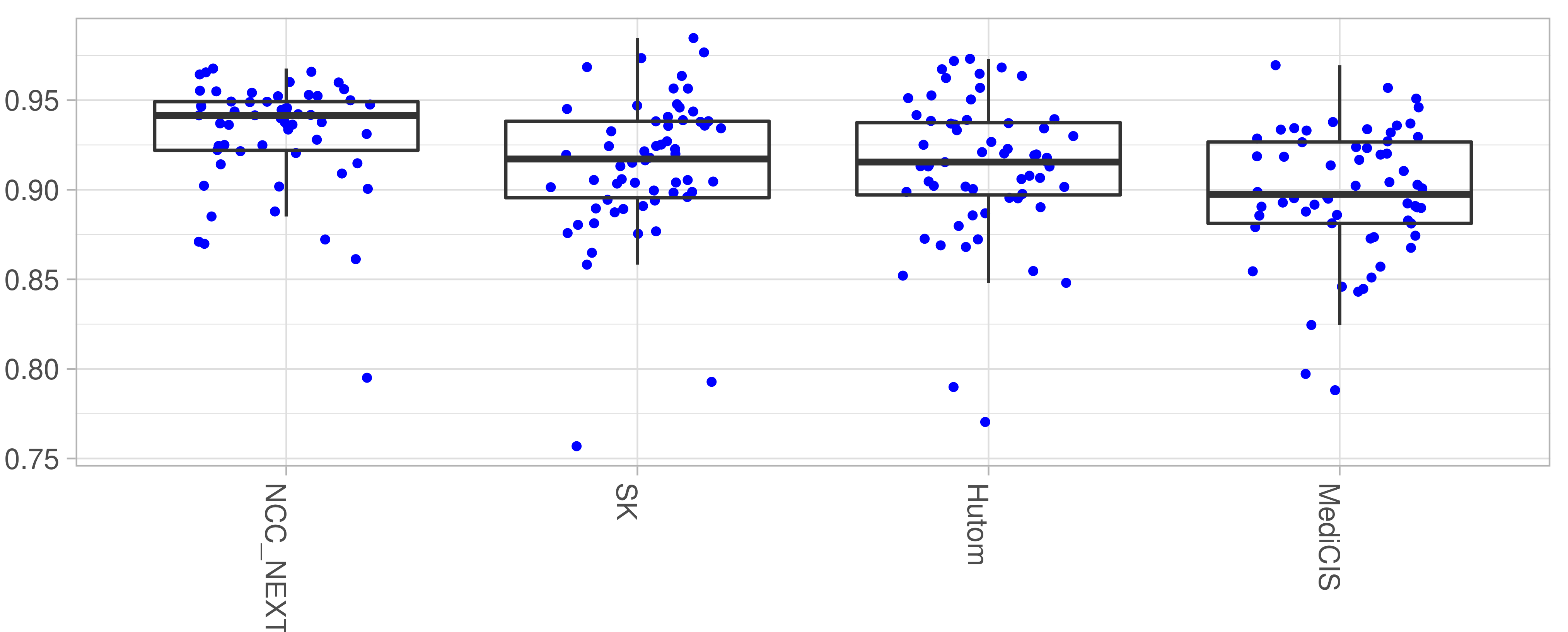}
    \caption{Average task 5 recognition AD-Accuracy for each team. Each dot represents the AD-Accuracy for one sequence.}
    \label{fig:T5_Team_results}
\end{figure}

\begin{figure}[H]
    \centering
    \includegraphics[width=\linewidth]{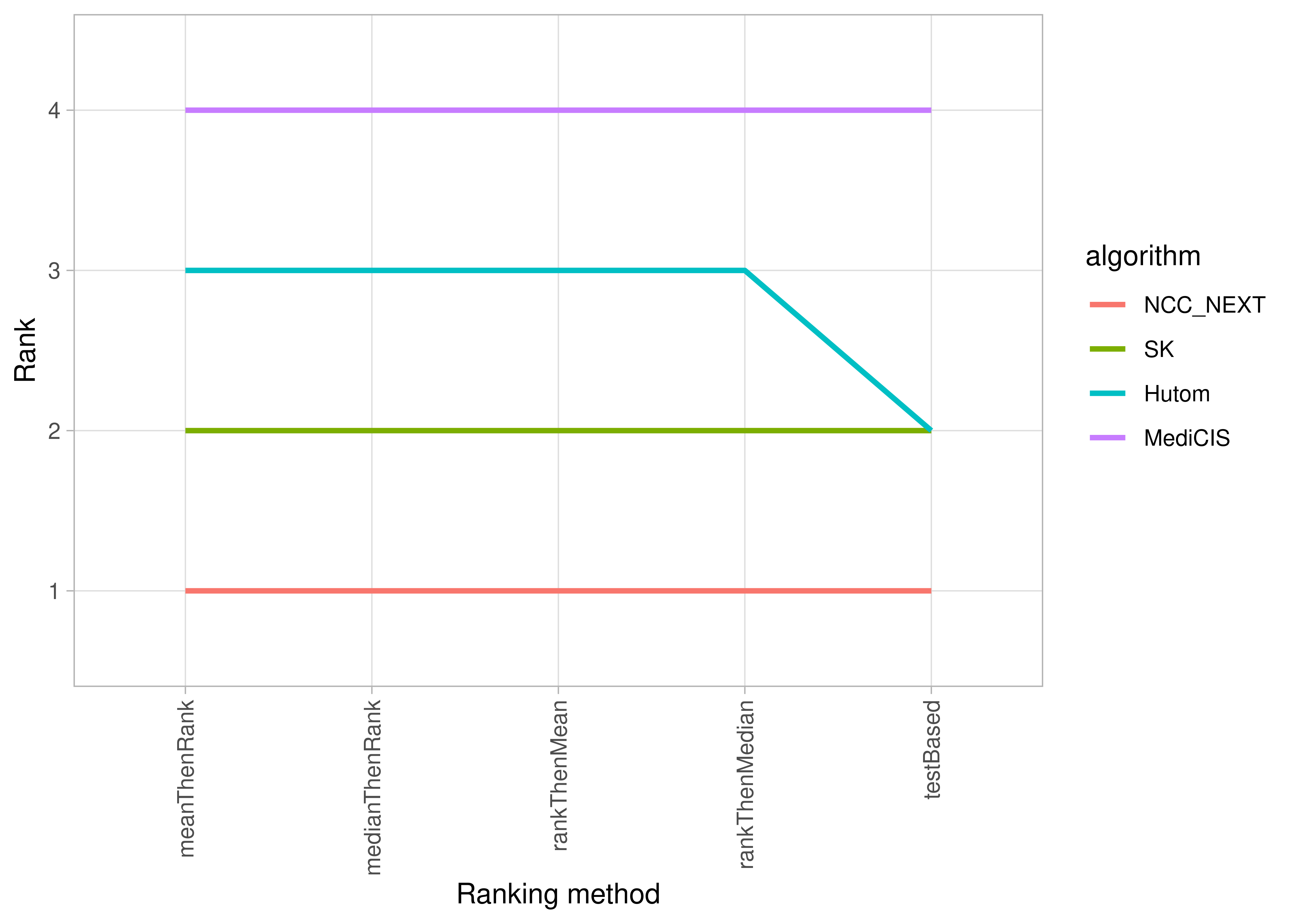}
    \caption{Task 5 ranking stability using the indicated ranking methods.}
    \label{fig:T5_Team_rank}
\end{figure}

The teams' mean AD-Accuracy values ranged between 93.1\% and 89.8\% (Figure \ref{fig:T5_Team_results}). The SK and Hutom teams displayed very similar results, with 91.4\% and 91.3\%, respectively. However, the chosen ranking method did not influence the final rank (Figure \ref{fig:T5_Team_rank}).

\subsubsection{Workflow recognition results summary}
Table \ref{tab:results_summary} summarizes the results of each team for the five tasks. All the best methods displayed mean AD-Accuracy superior to 90\%, except for task 3.

\begin{table}[H]
\centering
\begin{tabular}{|l|S[table-format=2.3]|c|S[table-format=2.3]|S[table-format=2.3]|S[table-format=2.3]|}
\hline
Team & \multicolumn{1}{|c|}{Task 1} & \multicolumn{1}{|c|}{Task 2} & \multicolumn{1}{|c|}{Task 3} & \multicolumn{1}{|c|}{Task 4} & \multicolumn{1}{|c|}{Task 5} \\ \hline
Hutom & 90.51* & 84.31 & 60.28* & 91.33* & 91.27* \\ \hline
JHU-CIRL &  & 86.45 &  &  &  \\ \hline
MedAIR & 84.31* & \textbf{90.72} &  & 86.98 &  \\ \hline
MMLAB &  &  &  & 84.80 &  \\ \hline
NCC NEXT & 87.77* & 90.32 & 87.71* & \textbf{93.09 *} & \textbf{93.09 *} \\ \hline
SK & \textbf{90.77} & 89.66 & \textbf{88.51} & 91.61 & 91.37 \\ \hline
MediCIS & 89.15 & 89.71 & 87.22 & 90.18 & 89.81 \\ \hline
\end{tabular}
\caption{Mean AD-Accuracy of each team for the five tasks. The best results are highlighted in bold for each task. Resubmitted models are highlighted with an asterisk.}
\label{tab:results_summary}
\end{table}

\subsection{Additional analyses}
\modif{The additional analyses concern four of the seven participating teams: Hutom, NCC Next, SK, and MediCIS. They were the only teams to participate with a combination of the same or similar models used for the unimodal tasks. Although MedAIR team participated in task 4 and the two corresponding unimodal tasks (1 and 2), the models used were too different to allow a model comparison.}

\subsubsection{Comparison between unimodal and multimodal models}
\modif{Table \ref{tab:uni_vs_multi} presents the results of the statistical analysis. For the four teams, the combination of video and kinematics (task 4) is statistically different than the use of only one modality (tasks 1 and 2). The same statistical differences exist between the combination of the three modalities (task 5) and each modality individually (tasks 1, 2, and 3), with the exception of task 2 and task 5 for the MediCIS team. However, the addition of the segmentation modality (task 5) to the video/kinematic-based (task 4) models was only significant for the MediCIS team.}

\begin{table}[h]
\centering
\begin{tabular}{|l|c|c|c|c|}
\hline
Team & Hutom & NCC NEXT & SK & MediCIS \\ \hline
T1 \textless{}\textgreater{} T4 & X & X & X & X \\ \hline
T2 \textless{}\textgreater{} T4 & X & X & X & X \\ \hline
T1 \textless{}\textgreater{} T5 & X & X & X & X \\ \hline
T2 \textless{}\textgreater{} T5 & X & X & X &  \\ \hline
T3 \textless{}\textgreater{} T5 & X & X & X & X \\ \hline
T4 \textless{}\textgreater{} T5 &  &  &  & X \\ \hline
\end{tabular}
\caption{Significant performance differences between unimodal and multimodal tasks. T1 \textless{}\textgreater{} T4: comparison of task 1 and task 4; X: significant performance variation (p-value \textless{} 0.05).}
\label{tab:uni_vs_multi}
\end{table}

\subsubsection{Execution time}
\modif{Table \ref{tab:execution_time} presents the execution time for the four teams and each task. For NCC Next team, the duration could not be determined because the predictions were locally written at the end of the Docker image execution. Execution time was highly variable among the teams, with the shortest (except task 2) achieved by the SK team. The shortest execution times overall were obtained for task 2 (3 min for SK and less than 1 minute for the Hutom and MediCIS teams).}

\begin{table}[h]
\centering
\begin{tabular}{|l|c|c|c|c|}
\hline
Team & Hutom & NCC NEXT & SK & MediCIS \\ \hline
Task 1 & 56 min & CBD & 50 min  & 202 min \\ \hline
Task 2 & \textless{} 1 min & CBD & 3 min & \textless{} 1 min  \\ \hline
Task 3 & 13 550 min & CBD & 145 min & 725 min \\ \hline
Task 4 & 57 min & CBD & 53 min & 203 min \\ \hline
Task 5 & 13 600 min & CBD & 175 min & 928 min \\ \hline
\end{tabular}
\caption{Execution times to compute the results of the 60 test sequences. CBD: Could not Be Determined}
\label{tab:execution_time}
\end{table}

\section{Discussion}
Accurate surgical workflow recognition is necessary for context-aware computer-assisted surgical systems. The proposed methods obtained good results but were not perfect and the PETRAW data set itself presented several limitations. Specifically, the peg transfer task is significantly easier than a real surgical intervention due to the simpler environment, clearly identifiable objects, static field of view, and constant lighting. In addition, each sequence was performed by the same operator resulting in lower data set variability.  

By analyzing the performance of the methods across individual sequences, we observed a gradual decrease in performance, except for two sequences (54 and 79) that displayed very low AD-Accuracy compared to the others regardless of modality. We analyzed these two sequences in detail to understand this poor performance. In sequence 54, the block was dropped twice during the transfer between hands, forcing the operator to catch the block for a second time. In addition, one block got stuck on the peg, forcing the operator to reposition it. Sequence 79 is one of the sequences identified as containing uncertainty (see Section \ref{seq:GT_uncertainties}). However, the overlapping steps (by 0.5 seconds) could not entirely explain the low performance, as the overlap was partially compensated by the delay of 0.25 seconds used to compute the AD-Accuracy. In addition, a block got stuck on a peg in this sequence and the order in which the blocks were caught did not correspond to the one used in most sequences. These deviations from the most common workflow might explain the low performance.

For task 1 (video-based recognition), ResNet-based models gave the best results, and the simplest model was ranked first. For task 2 (kinematic-based recognition), LSTM-based methods presented the worst results. For task 3, the two segmentation models used (DeepLabV3 and U-Net), displayed similar IoU values and the differences were probably due to differences in the training characteristics. For workflow recognition, the EfficientNetB7 and ResNet models obtained similar results. For Tasks 4 and 5, the NCC NEXT team's strategy (i.e., using the modality that gave the best results in the unimodal tasks for each workflow component) provided the best result.

For the segmentation-based recognition task (task 3), the segmentation quality seemed to influence workflow recognition up to a certain threshold. Indeed, the workflow recognition performances of the three teams with Macro IoU values superior to 94.0\% were similar (AD-Accuracy between 88.5\% and 87.2\%), but the ranking was inverted for the two first teams. Conversely, the workflow recognition performance with a Macro IoU value of 85\% dropped drastically (60.3\%). Additional research is required to fully quantify and understand the degree to which segmentation quality influences workflow recognition since, in this challenge, teams used different combinations of models for the segmentation and workflow recognition components.

For tasks 1 to 4, at least one team submitted a method that could be truly causal. It is important to note that several proposed methods were provably non-causal due to their preprocessing steps and not the core network such as with NCC NEXT (task 3), SK (task 1, 3, 4, 5), and MediCIS (task 2, 4 and 5). Causal methods generally have lower performance than non-causal models. With the exception of task 4, the causal methods displayed performances that were surprisingly close to that of the best method. For example, for task 2, the AD-Accuracy of the best method was 90.7\%, compared to 90.3\% and 89.7\% for the causal methods by NCC NEXT and SK, respectively. Obviously, it is not possible to conclude that causal methods give similar results to acausal models: i) because during the challenge we did not have the two versions of a similar method, ii) due to data simplicity. Nevertheless, the results of the causal methods are promising for developing applications, such as the implementation of automatic reports after training sessions on a virtual simulator.

\modif{Among the seven participating teams, four (Hutom, NCC Next, SK, and MediCIS) participated in the multimodal tasks (4 and 5) with a combination of the same or similar models used for the unimodal tasks. In all cases, recognition was improved when several modalities were used (Table \ref{tab:results_summary}); however, the addition of segmentation modality decreased the performance. The statistical analysis (Table \ref{tab:uni_vs_multi}) confirmed a significant performance improvement when using multimodal models, with the exception of tasks 2 and 5 for the MediCIS team. The performance decrease experienced with the addition of the segmentation modality to the video/kinematic-based models was only significant for the MediCIS team.}

\modif{Therefore, the combination of video and kinematic (task 4) data gives significantly better results compared with other modality combinations. The results obtained by the MedAIR team could contradict this point because they obtained better results for the kinematic-based recognition task than for the video/kinematic-based one. However, the models they used were very different: a Trans-SVNet and an MRG-Net combined with a CNN respectively. So, in this case, it is difficult to determine if the performance modifications were due to the model or to the modalities used. However, task 4 was more time-consuming than task 2 (53 vs. 3 minutes for SK, 57 vs. less than 1 for Hutom, and 203 vs. less than 1 for MediCIS). One may ask whether it is wise to spend 2,000\% to 20,000\% more computing time for less than a 3\% improvement. The training time should also be taken into account, as it is much more time-consuming \cite{PattersonCarbonTraining,Strubell2019EnergyNLP}, but we did not have access to this information. Data storage should also be considered. Video can require a lot of storage space, especially for long surgical interventions. Conversely, kinematic data are less voluminous.}

Future work should focus on overcoming the limitations of the current data set by including peg transfer sequences performed by several operators in different systems. Moreover, tests on more realistic data are necessary to validate the finding that kinematic data display the best performances in recognition rate and have less environmental impact thanks to the lowest computation time and storage cost.

\section*{Acknowledgements}
Authors thanks IRT b\textless \textgreater com for providing the ``Surgery Workflow Toolbox [annotate]'' software, used for this work.

\section*{Statements of ethical approval}
All procedures performed in studies involving human participants were in accordance with the ethical standards of the institutional and/or national research committee and with the 1964 Helsinki declaration and its later amendments or comparable ethical standards. This article does not contain patient data.

\section*{Conflict of interest statement}
The authors declare that they have no conflict of interest.

\bibliographystyle{unsrt}
\bibliography{PETRAW}

\section*{Supplementary material}

\subsection*{A. Authors' contributions}
A. Huaulmé was the challenge coordinator, the primary contact with the participant teams, and a member of the MediCIS team. He acquired the dataset, collected, computed, and analyzed the results, and wrote the paper. K. Harada and M. Mitsuishi were members of the challenge organizers and supervised the development of the peg transfer simulator. P. Jannin was the challenge supervisor. Q-M. Nguyen was a member of the MediCIS team. B. Park, S. Hong, and M-K Choi were members of the Hutom team. M. Peven was a member of the JHU-CIRL team. Y. Li, Y. Long, and Q. Dou were members of the MedAIR team. S. Kumar, S Lalithkumar, and R. Hongliang were members of the MMLAB Team. H. Matsizaki,Y. Ishikawa and Y. Harai were members of NCC Next team. S. Kondo was a member of the SK team. All co-authors participated in the proofreading of the parts concerning their work.

\subsection*{B. Data sets characteristics}
The following tables present the numerical value of the vocabulary term distribution in the training and test data set.

\begin{table}[h]
    \centering
    \begin{tabular}{c|c|c}
        Phase & Training& Test\\\hline
        Transfer Left to Right (L2R) & 48.54 \%&47.35\% \\
        Transfer Right to Left (R2L) & 47.87 \%&49.05\% \\
        Idle & 3.59  \%& 3.60 \%\\
    \end{tabular}
    \caption{Phase distribution in training and test data sets}
    \label{tab_sup:PhaseDistri}
\end{table}

\begin{table}[h]
    \centering
    \begin{tabular}{c|c|c}
        Step & Training& Test\\\hline
        Block 1 L2R&	8.29 \%&	8.96 \% \\
        Block 1 R2L&	8.50 \%&	8.51 \% \\
        Block 2 L2R&	8.03 \%&	8.12 \% \\
        Block 2 R2L&	7.58 \%&	8.41 \% \\
        Block 3 L2R&	8.42 \%&	7.72 \% \\
        Block 3 R2L&	8.14 \%&	8.04 \% \\
        Block 4 L2R&	7.82 \%&	7.40 \% \\
        Block 4 R2L&	7.49 \%&	7.34 \% \\
        Block 5 L2R&	7.91 \%&	7.65 \% \\
        Block 5 R2L&	7.48 \%&	7.83 \% \\
        Block 6 L2R&	8.07 \%&	7.51 \% \\
        Block 6 R2L&	8.69 \%&	8.92 \% \\
        Idle&	3.59 \%&	3.60 \% \\
    \end{tabular}
    \caption{Step distribution in training and test data sets}
    \label{tab_sup:StepDistri}
\end{table}

\begin{table}[h]
    \centering
    \begin{tabular}{c|c|c}
        Verb Left & Training& Test\\\hline
        Catch&	8.97 \%&	8.99 \% \\
        Drop&	2.15 \%&	1.91 \% \\
        Extract&	5.32 \%&	5.43 \% \\
        Hold&	26.33 \%&	25.90 \% \\
        Insert&	3.66 \%&	3.61 \% \\
        Touch&	0.59 \%&	0.60 \% \\
        Idle&	52.98 \%&	53.56 \% \\
    \end{tabular}
    \caption{Verb Left distribution in training and test data sets}
    \label{tab_sup:VerbLDistri}
\end{table}

\begin{table}[h]
    \centering
    \begin{tabular}{c|c|c}
        Verb Right & Training& Test\\\hline
        Catch&	7.65 \%&	7.64 \% \\
        Drop&	1.68 \%&	1.76 \% \\
        Extract&	4.75 \%&	5.01 \% \\
        Hold&	26.42 \%&	26.53 \% \\
        Insert&	3.75 \%&	3.66 \% \\
        Touch&	0.62 \%&	0.48 \% \\
        Idle&	55.14 \%&	54.91 \% \\

    \end{tabular}
    \caption{Verb Right distribution in training and test data sets}
    \label{tab_sup:VerbRDistri}
\end{table}

\subsection*{B. Detailed results for each team}
\label{supMat:detailed_results}
The results for each team by sequence and task are presented on the following pages. Eight balanced scores were computed, the frame-by-frame and application-dependent version of the accuracy, precision, recall, and F1. To limit the number of results, we only present here the ones for all granularities, as used during the challenge. The results for each granularity independently are available at \url{https://www.synapse.org/PETRAW} on the files section.

\subsubsection*{B.1 Hutom}
\begin{longtable}{ | c | c |  c | c | c || c  | c | c | c |}
    \caption{Results of team Hutom for task1 recognition.}
    \label{supMat:Hutom_Task1}\\
    \hline
    \multirow{2}{*}{Sequence}&\multicolumn{4}{c||}{Frame-by-Frame}&\multicolumn{4}{c|}{Application-Dependant}  \\\cline{2-9}
      &Accuracy&Precision&recall&F1&Accuracy&Precision&recall&F1\\\hline
    \endfirsthead
    \caption[]{Results of team Hutom for task1 recognition. (continued)}\\
    \hline
    \multirow{2}{*}{Sequence}&\multicolumn{4}{c||}{Frame-by-Frame}&\multicolumn{4}{c|}{Application-Dependant}  \\\cline{2-9}
      &Accuracy&Precision&recall&F1&Accuracy&Precision&recall&F1\\\hline\endhead
        003&85.80&95.45&94.88&94.95&88.51&96.53&95.95&96.03\\\hline
        004&86.28&96.39&96.40&96.30&89.69&97.38&97.35&97.29\\\hline
        006&86.23&96.45&96.34&96.34&90.35&97.34&97.28&97.25\\\hline
        009&89.03&96.08&95.93&95.91&90.85&96.95&96.81&96.78\\\hline
        012&83.62&94.13&93.92&93.87&86.19&94.86&94.71&94.64\\\hline
        015&83.17&93.48&92.64&92.84&86.05&94.18&93.52&93.64\\\hline
        019&83.01&93.36&93.09&93.00&85.20&94.25&94.06&93.90\\\hline
        020&87.17&95.38&95.25&95.18&90.05&96.39&96.27&96.20\\
\hline
\end{longtable}

\begin{longtable}{ | c | c |  c | c | c || c  | c | c | c |}
    \caption{Results of team Hutom for task2 recognition.}
    \label{supMat:Hutom_Task2}\\
    \hline
    \multirow{2}{*}{Sequence}&\multicolumn{4}{c||}{Frame-by-Frame}&\multicolumn{4}{c|}{Application-Dependant}  \\\cline{2-9}
      &Accuracy&Precision&recall&F1&Accuracy&Precision&recall&F1\\\hline
    \endfirsthead
    \caption[]{Results of team Hutom for task2 recognition. (continued)}\\
    \hline
    \multirow{2}{*}{Sequence}&\multicolumn{4}{c||}{Frame-by-Frame}&\multicolumn{4}{c|}{Application-Dependant}  \\\cline{2-9}
      &Accuracy&Precision&recall&F1&Accuracy&Precision&recall&F1\\\hline\endhead
        003&78.64&88.98&88.27&88.02&80.48&89.89&89.24&88.97\\\hline
        004&80.43&91.61&90.34&90.30&82.20&92.40&91.16&91.11\\\hline
        006&78.53&90.03&88.81&88.79&80.39&91.08&89.98&89.95\\\hline
        009&83.24&90.82&89.79&89.78&85.02&91.98&91.01&90.97\\\hline
        012&74.15&83.17&82.82&82.72&76.04&83.93&83.68&83.52\\\hline
        015&73.37&83.97&83.05&82.92&75.55&84.64&83.91&83.71\\\hline
        019&70.92&80.46&79.25&78.37&72.25&81.16&80.00&79.09\\\hline
        020&82.68&91.97&91.49&91.44&85.08&92.90&92.53&92.44\\
\hline
\end{longtable}

\begin{longtable}{ | c | c |  c | c | c || c  | c | c | c |}
    \caption{Results of team Hutom for task3 recognition.}
    \label{supMat:Hutom_Task3}\\
    \hline
    \multirow{2}{*}{Sequence}&\multicolumn{4}{c||}{Frame-by-Frame}&\multicolumn{4}{c|}{Application-Dependant}  \\\cline{2-9}
      &Accuracy&Precision&recall&F1&Accuracy&Precision&recall&F1\\\hline
    \endfirsthead
    \caption[]{Results of team Hutom for task3 recognition. (continued)}\\
    \hline
    \multirow{2}{*}{Sequence}&\multicolumn{4}{c||}{Frame-by-Frame}&\multicolumn{4}{c|}{Application-Dependant}  \\\cline{2-9}
      &Accuracy&Precision&recall&F1&Accuracy&Precision&recall&F1\\\hline\endhead
        003&55.95&84.67&84.49&81.66&57.93&85.10&84.96&82.20\\\hline
        004&57.98&87.05&86.56&83.61&59.88&87.49&87.04&84.14\\\hline
        006&56.44&86.43&86.45&84.18&58.55&86.97&87.06&84.97\\\hline
        009&60.72&83.95&85.74&83.04&62.23&84.61&86.35&83.79\\\hline
        012&54.60&85.06&84.68&82.23&56.71&85.49&85.10&82.67\\\hline
        015&54.90&85.01&83.94&81.81&57.96&85.67&84.42&82.38\\\hline
        019&54.71&83.61&83.15&81.09&55.86&84.00&83.53&81.56\\\hline
        020&58.28&83.46&85.31&82.05&59.95&83.98&85.81&82.65\\
\hline
\end{longtable}

\begin{longtable}{ | c | c |  c | c | c || c  | c | c | c |}
    \caption{Results of team Hutom for task4 recognition.}
    \label{supMat:Hutom_Task4}\\
    \hline
    \multirow{2}{*}{Sequence}&\multicolumn{4}{c||}{Frame-by-Frame}&\multicolumn{4}{c|}{Application-Dependant}  \\\cline{2-9}
      &Accuracy&Precision&recall&F1&Accuracy&Precision&recall&F1\\\hline
    \endfirsthead
    \caption[]{Results of team Hutom for task4 recognition. (continued)}\\
    \hline
    \multirow{2}{*}{Sequence}&\multicolumn{4}{c||}{Frame-by-Frame}&\multicolumn{4}{c|}{Application-Dependant}  \\\cline{2-9}
      &Accuracy&Precision&recall&F1&Accuracy&Precision&recall&F1\\\hline\endhead
        003&87.10&95.70&95.19&95.28&89.51&96.60&96.09&96.19\\\hline
        004&88.12&96.89&96.51&96.60&89.48&97.56&97.20&97.28\\\hline
        006&87.89&97.71&97.40&97.50&89.56&98.46&98.22&98.28\\\hline
        009&88.08&95.60&95.29&95.28&89.62&96.45&96.21&96.18\\\hline
        012&84.68&94.27&94.21&94.15&86.53&94.99&95.00&94.92\\\hline
        015&81.36&92.88&92.01&92.12&83.50&93.55&92.83&92.86\\\hline
        019&82.78&93.62&93.06&93.16&84.53&94.41&93.95&94.00\\\hline
        020&89.12&97.05&97.19&97.08&90.79&97.82&98.03&97.90\\
\hline
\end{longtable}

\begin{longtable}{ | c | c |  c | c | c || c  | c | c | c |}
    \caption{Results of team Hutom for task5 recognition.}
    \label{supMat:Hutom_Task5}\\
    \hline
    \multirow{2}{*}{Sequence}&\multicolumn{4}{c||}{Frame-by-Frame}&\multicolumn{4}{c|}{Application-Dependant}  \\\cline{2-9}
      &Accuracy&Precision&recall&F1&Accuracy&Precision&recall&F1\\\hline
    \endfirsthead
    \caption[]{Results of team Hutom for task5 recognition. (continued)}\\
    \hline
    \multirow{2}{*}{Sequence}&\multicolumn{4}{c||}{Frame-by-Frame}&\multicolumn{4}{c|}{Application-Dependant}  \\\cline{2-9}
      &Accuracy&Precision&recall&F1&Accuracy&Precision&recall&F1\\\hline\endhead
        003&85.19&95.08&94.27&94.41&87.23&96.01&95.29&95.41\\\hline
        004&90.82&96.14&95.72&95.81&93.32&97.04&96.66&96.72\\\hline
        006&88.27&96.48&96.11&96.21&91.30&97.32&97.07&97.12\\\hline
        009&90.57&95.97&95.57&95.62&91.92&96.83&96.54&96.54\\\hline
        012&83.17&91.88&90.25&90.53&85.21&92.61&91.17&91.37\\\hline
        015&83.57&92.19&90.19&90.77&85.46&92.71&91.05&91.50\\\hline
        019&85.55&92.64&91.49&91.65&88.57&93.50&92.60&92.64\\\hline
        020&88.56&95.81&95.41&95.47&91.31&96.77&96.47&96.50\\
\hline
\end{longtable}

\subsubsection*{B.2 JHU CIRL}
\begin{longtable}{ | c | c |  c | c | c || c  | c | c | c |}
    \caption{Results of team JHU\_CIRL for task2 recognition.}
    \label{supMat:JHU_CIRL_Task2}\\
    \hline
    \multirow{2}{*}{Sequence}&\multicolumn{4}{c||}{Frame-by-Frame}&\multicolumn{4}{c|}{Application-Dependant}  \\\cline{2-9}
      &Accuracy&Precision&recall&F1&Accuracy&Precision&recall&F1\\\hline
    \endfirsthead
    \caption[]{Results of team JHU\_CIRL for task2 recognition. (continued)}\\
    \hline
    \multirow{2}{*}{Sequence}&\multicolumn{4}{c||}{Frame-by-Frame}&\multicolumn{4}{c|}{Application-Dependant}  \\\cline{2-9}
      &Accuracy&Precision&recall&F1&Accuracy&Precision&recall&F1\\\hline\endhead
        003&83.56&95.46&95.82&95.60&85.70&96.35&96.67&96.47\\\hline
        004&81.71&94.33&94.42&94.31&83.80&95.18&95.19&95.09\\\hline
        006&82.33&96.04&96.05&95.97&85.14&97.16&97.11&97.05\\\hline
        009&85.46&95.33&95.41&95.18&87.85&96.25&96.28&96.09\\\hline
        012&80.88&93.19&93.25&93.08&83.10&93.88&93.96&93.77\\\hline
        015&74.66&88.49&85.84&85.31&76.58&89.03&86.44&85.91\\\hline
        019&79.44&93.38&93.97&93.41&81.41&94.28&94.72&94.22\\\hline
        020&83.88&95.84&96.35&96.03&87.13&96.90&97.36&97.06\\
\hline
\end{longtable}

\subsubsection*{B.3 MedAIR}
\begin{longtable}{ | c | c |  c | c | c || c  | c | c | c |}
    \caption{Results of team MedAIR for task1 recognition.}
    \label{supMat:MedAIR_Task1}\\
    \hline
    \multirow{2}{*}{Sequence}&\multicolumn{4}{c||}{Frame-by-Frame}&\multicolumn{4}{c|}{Application-Dependant}  \\\cline{2-9}
      &Accuracy&Precision&recall&F1&Accuracy&Precision&recall&F1\\\hline
    \endfirsthead
    \caption[]{Results of team MedAIR for task1 recognition. (continued)}\\
    \hline
    \multirow{2}{*}{Sequence}&\multicolumn{4}{c||}{Frame-by-Frame}&\multicolumn{4}{c|}{Application-Dependant}  \\\cline{2-9}
      &Accuracy&Precision&recall&F1&Accuracy&Precision&recall&F1\\\hline\endhead
        003&79.26&93.65&93.38&92.86&81.80&94.45&94.15&93.69\\\hline
        004&82.85&96.31&96.23&95.90&84.89&96.98&96.93&96.64\\\hline
        006&79.67&95.28&95.30&94.84&81.61&96.08&96.02&95.62\\\hline
        009&82.09&94.36&94.07&93.53&84.19&95.14&94.87&94.39\\\hline
        012&78.92&94.45&94.03&93.70&81.21&95.17&94.67&94.39\\\hline
        015&77.64&92.95&92.97&92.54&80.36&93.47&93.52&93.09\\\hline
        019&76.80&92.73&92.69&92.00&77.98&93.25&93.21&92.56\\\hline
        020&79.74&93.80&93.81&93.12&82.43&94.54&94.51&93.90\\
\hline
\end{longtable}

\begin{longtable}{ | c | c |  c | c | c || c  | c | c | c |}
    \caption{Results of team MedAIR for task2 recognition.}
    \label{supMat:MedAIR_Task2}\\
    \hline
    \multirow{2}{*}{Sequence}&\multicolumn{4}{c||}{Frame-by-Frame}&\multicolumn{4}{c|}{Application-Dependant}  \\\cline{2-9}
      &Accuracy&Precision&recall&F1&Accuracy&Precision&recall&F1\\\hline
    \endfirsthead
    \caption[]{Results of team MedAIR for task2 recognition. (continued)}\\
    \hline
    \multirow{2}{*}{Sequence}&\multicolumn{4}{c||}{Frame-by-Frame}&\multicolumn{4}{c|}{Application-Dependant}  \\\cline{2-9}
      &Accuracy&Precision&recall&F1&Accuracy&Precision&recall&F1\\\hline\endhead
        003&88.30&96.98&96.96&96.93&89.47&97.67&97.75&97.68\\\hline
        004&88.73&97.35&97.08&97.14&89.56&97.92&97.79&97.80\\\hline
        006&90.80&97.50&97.42&97.41&92.12&98.25&98.26&98.22\\\hline
        009&89.91&96.56&96.20&96.22&90.97&97.24&97.00&96.99\\\hline
        012&81.56&91.89&90.56&90.43&83.34&92.41&91.13&90.98\\\hline
        015&85.33&94.77&93.27&93.57&87.00&95.38&94.01&94.23\\\hline
        019&79.11&89.97&87.65&87.57&80.20&90.58&88.34&88.20\\\hline
        020&92.39&97.30&96.80&96.96&93.96&98.00&97.66&97.77\\
\hline
\end{longtable}

\begin{longtable}{ | c | c |  c | c | c || c  | c | c | c |}
    \caption{Results of team MedAIR for task4 recognition.}
    \label{supMat:MedAIR_Task4}\\
    \hline
    \multirow{2}{*}{Sequence}&\multicolumn{4}{c||}{Frame-by-Frame}&\multicolumn{4}{c|}{Application-Dependant}  \\\cline{2-9}
      &Accuracy&Precision&recall&F1&Accuracy&Precision&recall&F1\\\hline
    \endfirsthead
    \caption[]{Results of team MedAIR for task4 recognition. (continued)}\\
    \hline
    \multirow{2}{*}{Sequence}&\multicolumn{4}{c||}{Frame-by-Frame}&\multicolumn{4}{c|}{Application-Dependant}  \\\cline{2-9}
      &Accuracy&Precision&recall&F1&Accuracy&Precision&recall&F1\\\hline\endhead
        003&83.87&95.44&95.16&95.16&86.15&96.47&96.17&96.19\\\hline
        004&83.56&95.43&95.31&95.28&85.39&96.24&96.10&96.09\\\hline
        006&83.25&96.11&95.80&95.88&85.37&97.08&96.76&96.86\\\hline
        009&87.12&95.37&95.33&95.22&89.42&96.33&96.29&96.21\\\hline
        012&83.75&94.31&93.27&93.59&87.87&94.91&94.03&94.29\\\hline
        015&79.91&93.64&91.74&92.22&81.39&94.14&92.45&92.86\\\hline
        019&79.96&92.62&92.21&92.09&81.20&93.41&92.99&92.84\\\hline
        020&84.58&95.17&95.14&95.02&86.17&96.18&96.23&96.09\\
\hline
\end{longtable}

\subsubsection*{B.4 MMLAB}
\begin{longtable}{ | c | c |  c | c | c || c  | c | c | c |}
    \caption{Results of team MMLAB for task4 recognition.}
    \label{supMat:MMLAB_Task4}\\
    \hline
    \multirow{2}{*}{Sequence}&\multicolumn{4}{c||}{Frame-by-Frame}&\multicolumn{4}{c|}{Application-Dependant}  \\\cline{2-9}
      &Accuracy&Precision&recall&F1&Accuracy&Precision&recall&F1\\\hline
    \endfirsthead
    \caption[]{Results of team MMLAB for task4 recognition. (continued)}\\
    \hline
    \multirow{2}{*}{Sequence}&\multicolumn{4}{c||}{Frame-by-Frame}&\multicolumn{4}{c|}{Application-Dependant}  \\\cline{2-9}
      &Accuracy&Precision&recall&F1&Accuracy&Precision&recall&F1\\\hline\endhead
        003&82.34&93.42&93.12&92.95&84.80&94.31&93.94&93.84\\\hline
        004&83.66&94.86&94.85&94.73&85.98&95.77&95.72&95.66\\\hline
        006&80.34&94.78&94.66&94.52&82.35&95.62&95.36&95.27\\\hline
        009&81.20&92.87&92.84&92.37&83.80&93.87&93.66&93.29\\\hline
        012&79.78&91.22&90.75&90.63&84.21&91.86&91.41&91.28\\\hline
        015&76.61&90.11&89.88&89.62&78.06&90.74&90.51&90.23\\\hline
        019&78.75&92.23&92.49&92.00&80.96&93.12&93.18&92.77\\\hline
        020&82.75&94.86&95.12&94.87&85.11&95.69&95.87&95.66\\
\hline
\end{longtable}

\subsubsection*{B.5 NCC NEXT}
\begin{longtable}{ | c | c |  c | c | c || c  | c | c | c |}
    \caption{Results of team NCC\_NEXT for task1 recognition.}
    \label{supMat:NCC_NEXT_Task1}\\
    \hline
    \multirow{2}{*}{Sequence}&\multicolumn{4}{c||}{Frame-by-Frame}&\multicolumn{4}{c|}{Application-Dependant}  \\\cline{2-9}
      &Accuracy&Precision&recall&F1&Accuracy&Precision&recall&F1\\\hline
    \endfirsthead
    \caption[]{Results of team NCC\_NEXT for task1 recognition. (continued)}\\
    \hline
    \multirow{2}{*}{Sequence}&\multicolumn{4}{c||}{Frame-by-Frame}&\multicolumn{4}{c|}{Application-Dependant}  \\\cline{2-9}
      &Accuracy&Precision&recall&F1&Accuracy&Precision&recall&F1\\\hline\endhead
        003&83.09&94.96&95.05&94.79&84.71&95.80&95.78&95.58\\\hline
        004&83.48&95.80&95.58&95.49&85.70&96.63&96.37&96.29\\\hline
        006&82.50&96.00&95.53&95.66&84.73&97.05&96.63&96.75\\\hline
        009&84.32&94.77&94.72&94.44&86.70&95.88&95.81&95.58\\\hline
        012&82.26&94.48&94.92&94.58&84.58&95.34&95.70&95.42\\\hline
        015&81.77&94.83&94.00&94.16&83.55&95.50&94.68&94.84\\\hline
        019&81.64&93.95&94.19&93.94&83.58&94.75&95.00&94.77\\\hline
        020&85.01&95.89&96.01&95.83&88.84&97.06&97.10&96.98\\
\hline
\end{longtable}

\begin{longtable}{ | c | c |  c | c | c || c  | c | c | c |}
    \caption{Results of team NCC\_NEXT for task2 recognition.}
    \label{supMat:NCC_NEXT_Task2}\\
    \hline
    \multirow{2}{*}{Sequence}&\multicolumn{4}{c||}{Frame-by-Frame}&\multicolumn{4}{c|}{Application-Dependant}  \\\cline{2-9}
      &Accuracy&Precision&recall&F1&Accuracy&Precision&recall&F1\\\hline
    \endfirsthead
    \caption[]{Results of team NCC\_NEXT for task2 recognition. (continued)}\\
    \hline
    \multirow{2}{*}{Sequence}&\multicolumn{4}{c||}{Frame-by-Frame}&\multicolumn{4}{c|}{Application-Dependant}  \\\cline{2-9}
      &Accuracy&Precision&recall&F1&Accuracy&Precision&recall&F1\\\hline\endhead
        003&83.79&90.86&87.72&88.17&84.97&91.66&88.77&89.13\\\hline
        004&78.99&84.75&79.57&80.29&81.30&85.28&80.34&80.95\\\hline
        006&82.27&93.73&90.26&90.79&83.88&94.56&91.32&91.77\\\hline
        009&86.10&93.56&91.70&92.20&87.50&94.44&92.98&93.32\\\hline
        012&80.25&88.75&84.36&85.28&81.48&89.17&84.94&85.76\\\hline
        015&83.64&90.83&87.27&88.46&85.60&91.33&88.09&89.15\\\hline
        019&76.84&86.51&81.03&81.56&78.02&87.02&81.74&82.19\\\hline
        020&87.49&92.70&90.22&90.99&89.14&93.42&91.26&91.90\\
\hline
\end{longtable}

\begin{longtable}{ | c | c |  c | c | c || c  | c | c | c |}
    \caption{Results of team NCC\_NEXT for task3 recognition.}
    \label{supMat:NCC_NEXT_Task3}\\
    \hline
    \multirow{2}{*}{Sequence}&\multicolumn{4}{c||}{Frame-by-Frame}&\multicolumn{4}{c|}{Application-Dependant}  \\\cline{2-9}
      &Accuracy&Precision&recall&F1&Accuracy&Precision&recall&F1\\\hline
    \endfirsthead
    \caption[]{Results of team NCC\_NEXT for task3 recognition. (continued)}\\
    \hline
    \multirow{2}{*}{Sequence}&\multicolumn{4}{c||}{Frame-by-Frame}&\multicolumn{4}{c|}{Application-Dependant}  \\\cline{2-9}
      &Accuracy&Precision&recall&F1&Accuracy&Precision&recall&F1\\\hline\endhead
        003&84.17&94.95&94.81&94.67&87.47&96.18&96.04&95.91\\\hline
        004&84.80&95.67&95.68&95.62&87.09&96.78&96.78&96.72\\\hline
        006&81.08&94.99&94.60&94.65&83.17&96.13&95.77&95.81\\\hline
        009&83.70&94.50&94.18&94.08&87.33&96.11&95.80&95.69\\\hline
        012&81.11&94.94&94.52&94.66&84.48&95.98&95.51&95.64\\\hline
        015&80.69&93.93&92.84&93.05&83.28&94.66&93.71&93.86\\\hline
        019&82.46&93.87&93.95&93.73&84.47&94.72&94.77&94.56\\\hline
        020&84.73&95.11&94.75&94.65&88.33&96.52&96.18&96.10\\
\hline
\end{longtable}

\begin{longtable}{ | c | c |  c | c | c || c  | c | c | c |}
    \caption{Results of team NCC\_NEXT for task4 recognition.}
    \label{supMat:NCC_NEXT_Task4}\\
    \hline
    \multirow{2}{*}{Sequence}&\multicolumn{4}{c||}{Frame-by-Frame}&\multicolumn{4}{c|}{Application-Dependant}  \\\cline{2-9}
      &Accuracy&Precision&recall&F1&Accuracy&Precision&recall&F1\\\hline
    \endfirsthead
    \caption[]{Results of team NCC\_NEXT for task4 recognition. (continued)}\\
    \hline
    \multirow{2}{*}{Sequence}&\multicolumn{4}{c||}{Frame-by-Frame}&\multicolumn{4}{c|}{Application-Dependant}  \\\cline{2-9}
      &Accuracy&Precision&recall&F1&Accuracy&Precision&recall&F1\\\hline\endhead
        003&89.92&95.84&94.36&94.86&91.19&96.69&95.47&95.88\\\hline
        004&93.79&96.96&95.58&96.08&96.12&97.58&96.47&96.86\\\hline
        006&87.17&96.74&95.00&95.64&88.87&97.54&96.07&96.61\\\hline
        009&88.29&95.38&93.95&94.46&89.85&96.32&95.28&95.64\\\hline
        012&85.37&93.67&90.29&91.41&86.64&94.06&90.90&91.92\\\hline
        015&87.20&94.41&90.75&92.02&88.99&94.92&91.58&92.71\\\hline
        019&85.74&94.85&92.22&93.17&86.88&95.36&92.94&93.81\\\hline
        020&90.97&95.78&93.66&94.38&92.60&96.51&94.69&95.28\\
\hline
\end{longtable}

\begin{longtable}{ | c | c |  c | c | c || c  | c | c | c |}
    \caption{Results of team NCC\_NEXT for task5 recognition.}
    \label{supMat:NCC_NEXT_Task5}\\
    \hline
    \multirow{2}{*}{Sequence}&\multicolumn{4}{c||}{Frame-by-Frame}&\multicolumn{4}{c|}{Application-Dependant}  \\\cline{2-9}
      &Accuracy&Precision&recall&F1&Accuracy&Precision&recall&F1\\\hline
    \endfirsthead
    \caption[]{Results of team NCC\_NEXT for task5 recognition. (continued)}\\
    \hline
    \multirow{2}{*}{Sequence}&\multicolumn{4}{c||}{Frame-by-Frame}&\multicolumn{4}{c|}{Application-Dependant}  \\\cline{2-9}
      &Accuracy&Precision&recall&F1&Accuracy&Precision&recall&F1\\\hline\endhead
        003&90.16&95.93&94.58&95.05&91.42&96.78&95.70&96.07\\\hline
        004&93.63&96.83&95.46&95.95&95.99&97.48&96.39&96.77\\\hline
        006&86.85&96.39&94.62&95.26&88.51&97.16&95.66&96.20\\\hline
        009&88.50&95.60&94.16&94.67&90.05&96.51&95.47&95.83\\\hline
        012&85.87&94.09&90.75&91.88&87.10&94.44&91.33&92.35\\\hline
        015&87.02&94.10&90.45&91.71&88.79&94.60&91.27&92.39\\\hline
        019&85.79&95.00&92.40&93.35&86.98&95.58&93.18&94.05\\\hline
        020&90.43&95.29&93.04&93.78&92.05&95.99&94.06&94.66\\
\hline
\end{longtable}

\subsubsection*{B.6 SK}
\begin{longtable}{ | c | c |  c | c | c || c  | c | c | c |}
    \caption{Results of team SK for task1 recognition.}
    \label{supMat:SK_Task1}\\
    \hline
    \multirow{2}{*}{Sequence}&\multicolumn{4}{c||}{Frame-by-Frame}&\multicolumn{4}{c|}{Application-Dependant}  \\\cline{2-9}
      &Accuracy&Precision&recall&F1&Accuracy&Precision&recall&F1\\\hline
    \endfirsthead
    \caption[]{Results of team SK for task1 recognition. (continued)}\\
    \hline
    \multirow{2}{*}{Sequence}&\multicolumn{4}{c||}{Frame-by-Frame}&\multicolumn{4}{c|}{Application-Dependant}  \\\cline{2-9}
      &Accuracy&Precision&recall&F1&Accuracy&Precision&recall&F1\\\hline\endhead
        003&88.23&96.50&96.05&96.15&91.08&97.48&97.03&97.14\\\hline
        004&87.26&96.84&96.50&96.59&88.82&97.63&97.32&97.39\\\hline
        006&86.00&96.50&95.93&96.12&89.57&97.39&96.97&97.10\\\hline
        009&88.79&95.59&94.81&94.98&90.59&96.51&95.81&95.94\\\hline
        012&85.68&95.48&95.12&95.23&88.50&96.40&96.07&96.16\\\hline
        015&83.33&94.83&93.33&93.75&86.11&95.51&94.12&94.49\\\hline
        019&83.06&94.07&93.14&93.23&84.37&94.80&93.97&94.01\\\hline
        020&87.88&96.31&96.08&96.10&90.35&97.05&96.86&96.85\\
\hline
\end{longtable}

\begin{longtable}{ | c | c |  c | c | c || c  | c | c | c |}
    \caption{Results of team SK for task2 recognition.}
    \label{supMat:SK_Task2}\\
    \hline
    \multirow{2}{*}{Sequence}&\multicolumn{4}{c||}{Frame-by-Frame}&\multicolumn{4}{c|}{Application-Dependant}  \\\cline{2-9}
      &Accuracy&Precision&recall&F1&Accuracy&Precision&recall&F1\\\hline
    \endfirsthead
    \caption[]{Results of team SK for task2 recognition. (continued)}\\
    \hline
    \multirow{2}{*}{Sequence}&\multicolumn{4}{c||}{Frame-by-Frame}&\multicolumn{4}{c|}{Application-Dependant}  \\\cline{2-9}
      &Accuracy&Precision&recall&F1&Accuracy&Precision&recall&F1\\\hline\endhead
        003&87.85&96.24&96.04&96.04&89.37&97.07&97.02&96.95\\\hline
        004&87.69&96.73&96.18&96.25&89.02&97.41&96.97&97.00\\\hline
        006&87.35&97.42&97.06&97.17&89.03&98.31&98.08&98.15\\\hline
        009&89.21&96.38&96.42&96.28&91.05&97.22&97.41&97.21\\\hline
        012&78.89&90.14&89.49&89.27&80.61&90.74&90.22&89.95\\\hline
        015&81.24&92.63&89.98&88.80&82.72&93.05&90.62&89.36\\\hline
        019&79.12&90.98&89.31&89.35&80.50&91.55&90.00&89.97\\\hline
        020&88.56&96.30&96.31&96.24&90.45&97.00&97.18&97.04\\
\hline
\end{longtable}

\begin{longtable}{ | c | c |  c | c | c || c  | c | c | c |}
    \caption{Results of team SK for task3 recognition.}
    \label{supMat:SK_Task3}\\
    \hline
    \multirow{2}{*}{Sequence}&\multicolumn{4}{c||}{Frame-by-Frame}&\multicolumn{4}{c|}{Application-Dependant}  \\\cline{2-9}
      &Accuracy&Precision&recall&F1&Accuracy&Precision&recall&F1\\\hline
    \endfirsthead
    \caption[]{Results of team SK for task3 recognition. (continued)}\\
    \hline
    \multirow{2}{*}{Sequence}&\multicolumn{4}{c||}{Frame-by-Frame}&\multicolumn{4}{c|}{Application-Dependant}  \\\cline{2-9}
      &Accuracy&Precision&recall&F1&Accuracy&Precision&recall&F1\\\hline\endhead
        003&86.19&95.29&94.73&94.92&88.85&96.33&95.95&96.06\\\hline
        004&85.50&95.64&95.20&95.34&87.61&96.52&96.17&96.28\\\hline
        006&87.45&95.99&94.92&95.24&89.77&96.96&96.05&96.32\\\hline
        009&85.56&94.96&94.14&94.29&87.55&95.97&95.25&95.37\\\hline
        012&82.77&94.46&93.00&93.52&85.43&95.23&93.88&94.35\\\hline
        015&81.81&93.98&91.42&92.29&84.35&94.63&92.27&93.04\\\hline
        019&85.39&94.23&92.65&92.95&87.67&95.06&93.58&93.84\\\hline
        020&88.13&95.26&94.77&94.90&91.33&96.46&96.16&96.20\\
\hline
\end{longtable}

\begin{longtable}{ | c | c |  c | c | c || c  | c | c | c |}
    \caption{Results of team SK for task4 recognition.}
    \label{supMat:SK_Task4}\\
    \hline
    \multirow{2}{*}{Sequence}&\multicolumn{4}{c||}{Frame-by-Frame}&\multicolumn{4}{c|}{Application-Dependant}  \\\cline{2-9}
      &Accuracy&Precision&recall&F1&Accuracy&Precision&recall&F1\\\hline
    \endfirsthead
    \caption[]{Results of team SK for task4 recognition. (continued)}\\
    \hline
    \multirow{2}{*}{Sequence}&\multicolumn{4}{c||}{Frame-by-Frame}&\multicolumn{4}{c|}{Application-Dependant}  \\\cline{2-9}
      &Accuracy&Precision&recall&F1&Accuracy&Precision&recall&F1\\\hline\endhead
        003&89.94&97.50&97.44&97.42&92.04&98.32&98.28&98.24\\\hline
        004&88.41&97.88&97.93&97.88&90.29&98.67&98.75&98.70\\\hline
        006&86.98&97.58&97.52&97.52&88.61&98.39&98.38&98.35\\\hline
        009&90.25&96.76&96.64&96.56&92.09&97.72&97.56&97.49\\\hline
        012&86.17&95.75&95.83&95.72&87.51&96.34&96.46&96.34\\\hline
        015&83.86&94.87&94.19&94.14&85.32&95.40&94.91&94.79\\\hline
        019&83.42&95.05&94.18&94.22&84.53&95.78&94.95&94.95\\\hline
        020&88.26&96.73&96.74&96.66&91.33&97.59&97.64&97.55\\
\hline
\end{longtable}

\begin{longtable}{ | c | c |  c | c | c || c  | c | c | c |}
    \caption{Results of team SK for task5 recognition.}
    \label{supMat:SK_Task5}\\
    \hline
    \multirow{2}{*}{Sequence}&\multicolumn{4}{c||}{Frame-by-Frame}&\multicolumn{4}{c|}{Application-Dependant}  \\\cline{2-9}
      &Accuracy&Precision&recall&F1&Accuracy&Precision&recall&F1\\\hline
    \endfirsthead
    \caption[]{Results of team SK for task5 recognition. (continued)}\\
    \hline
    \multirow{2}{*}{Sequence}&\multicolumn{4}{c||}{Frame-by-Frame}&\multicolumn{4}{c|}{Application-Dependant}  \\\cline{2-9}
      &Accuracy&Precision&recall&F1&Accuracy&Precision&recall&F1\\\hline\endhead
        003&89.78&97.61&97.61&97.56&92.14&98.62&98.67&98.61\\\hline
        004&88.65&97.68&97.65&97.63&90.39&98.51&98.52&98.49\\\hline
        006&86.69&97.16&96.90&96.94&88.74&98.16&98.00&98.01\\\hline
        009&88.84&96.50&96.32&96.25&90.34&97.44&97.30&97.19\\\hline
        012&85.75&96.08&96.14&96.04&87.54&96.78&96.91&96.78\\\hline
        015&84.69&95.61&94.81&94.95&86.48&96.12&95.47&95.56\\\hline
        019&84.41&95.39&94.40&94.45&85.82&96.18&95.23&95.25\\\hline
        020&89.76&96.97&96.94&96.89&91.95&97.74&97.81&97.72\\
\hline
\end{longtable}

\subsubsection*{B.6 MediCIS}
\begin{longtable}{ | c | c |  c | c | c || c  | c | c | c |}
    \caption{Results of team MediCIS for task1 recognition.}
    \label{supMat:MediCIS_Task1}\\
    \hline
    \multirow{2}{*}{Sequence}&\multicolumn{4}{c||}{Frame-by-Frame}&\multicolumn{4}{c|}{Application-Dependant}  \\\cline{2-9}
      &Accuracy&Precision&recall&F1&Accuracy&Precision&recall&F1\\\hline
    \endfirsthead
    \caption[]{Results of team MediCIS for task1 recognition. (continued)}\\
    \hline
    \multirow{2}{*}{Sequence}&\multicolumn{4}{c||}{Frame-by-Frame}&\multicolumn{4}{c|}{Application-Dependant}  \\\cline{2-9}
      &Accuracy&Precision&recall&F1&Accuracy&Precision&recall&F1\\\hline\endhead
        003&85.61&96.36&96.54&96.41&88.70&97.51&97.70&97.57\\\hline
        004&86.52&96.93&96.93&96.84&88.88&97.81&97.79&97.72\\\hline
        006&84.20&96.57&96.47&96.47&86.25&97.51&97.39&97.40\\\hline
        009&86.84&95.98&96.09&95.92&89.00&96.98&97.13&96.94\\\hline
        012&84.08&95.23&95.33&95.21&86.69&95.95&96.06&95.94\\\hline
        015&83.74&95.62&95.12&95.14&86.20&96.24&95.86&95.83\\\hline
        019&78.48&88.99&87.36&87.19&80.06&89.77&88.16&87.98\\\hline
        020&87.04&97.06&96.98&96.95&90.21&98.05&97.97&97.93\\
\hline
\end{longtable}

\begin{longtable}{ | c | c |  c | c | c || c  | c | c | c |}
    \caption{Results of team MediCIS for task2 recognition.}
    \label{supMat:MediCIS_Task2}\\
    \hline
    \multirow{2}{*}{Sequence}&\multicolumn{4}{c||}{Frame-by-Frame}&\multicolumn{4}{c|}{Application-Dependant}  \\\cline{2-9}
      &Accuracy&Precision&recall&F1&Accuracy&Precision&recall&F1\\\hline
    \endfirsthead
    \caption[]{Results of team MediCIS for task2 recognition. (continued)}\\
    \hline
    \multirow{2}{*}{Sequence}&\multicolumn{4}{c||}{Frame-by-Frame}&\multicolumn{4}{c|}{Application-Dependant}  \\\cline{2-9}
      &Accuracy&Precision&recall&F1&Accuracy&Precision&recall&F1\\\hline\endhead
        003&85.54&96.05&96.12&96.04&87.60&96.93&96.96&96.90\\\hline
        004&88.48&96.34&95.86&95.94&91.73&97.00&96.45&96.54\\\hline
        006&88.24&96.80&96.44&96.56&93.43&97.89&97.54&97.65\\\hline
        009&86.65&95.47&95.41&95.32&88.52&96.29&96.30&96.17\\\hline
        012&82.99&94.40&93.57&93.83&84.84&95.06&94.32&94.53\\\hline
        015&78.54&89.23&87.77&87.52&81.42&89.87&88.49&88.20\\\hline
        019&77.73&87.63&86.82&86.92&79.43&88.34&87.62&87.67\\\hline
        020&87.50&96.64&96.44&96.51&89.97&97.51&97.33&97.39\\
\hline
\end{longtable}

\begin{longtable}{ | c | c |  c | c | c || c  | c | c | c |}
    \caption{Results of team MediCIS for task3 recognition.}
    \label{supMat:MediCIS_Task3}\\
    \hline
    \multirow{2}{*}{Sequence}&\multicolumn{4}{c||}{Frame-by-Frame}&\multicolumn{4}{c|}{Application-Dependant}  \\\cline{2-9}
      &Accuracy&Precision&recall&F1&Accuracy&Precision&recall&F1\\\hline
    \endfirsthead
    \caption[]{Results of team MediCIS for task3 recognition. (continued)}\\
    \hline
    \multirow{2}{*}{Sequence}&\multicolumn{4}{c||}{Frame-by-Frame}&\multicolumn{4}{c|}{Application-Dependant}  \\\cline{2-9}
      &Accuracy&Precision&recall&F1&Accuracy&Precision&recall&F1\\\hline\endhead
        003&86.78&95.85&95.75&95.75&89.59&96.89&96.75&96.75\\\hline
        004&82.49&95.00&94.84&94.84&83.94&95.63&95.49&95.48\\\hline
        006&84.63&95.81&95.52&95.60&87.92&96.84&96.59&96.64\\\hline
        009&85.60&95.13&95.13&94.99&87.67&96.02&96.06&95.92\\\hline
        012&82.65&95.00&94.46&94.49&84.97&95.73&95.26&95.27\\\hline
        015&81.43&93.90&93.08&93.01&83.03&94.48&93.75&93.63\\\hline
        019&77.07&89.05&87.37&87.43&78.84&89.78&88.16&88.19\\\hline
        020&84.68&95.95&96.01&95.91&87.31&97.04&97.11&97.01\\
\hline
\end{longtable}

\begin{longtable}{ | c | c |  c | c | c || c  | c | c | c |}
    \caption{Results of team MediCIS for task4 recognition.}
    \label{supMat:MediCIS_Task4}\\
    \hline
    \multirow{2}{*}{Sequence}&\multicolumn{4}{c||}{Frame-by-Frame}&\multicolumn{4}{c|}{Application-Dependant}  \\\cline{2-9}
      &Accuracy&Precision&recall&F1&Accuracy&Precision&recall&F1\\\hline
    \endfirsthead
    \caption[]{Results of team MediCIS for task4 recognition. (continued)}\\
    \hline
    \multirow{2}{*}{Sequence}&\multicolumn{4}{c||}{Frame-by-Frame}&\multicolumn{4}{c|}{Application-Dependant}  \\\cline{2-9}
      &Accuracy&Precision&recall&F1&Accuracy&Precision&recall&F1\\\hline\endhead
        003&86.70&96.77&96.76&96.66&88.69&97.57&97.58&97.50\\\hline
        004&87.96&97.33&97.29&97.24&89.89&98.06&98.04&97.99\\\hline
        006&85.25&96.88&96.78&96.78&87.19&97.91&97.79&97.81\\\hline
        009&88.30&96.48&96.61&96.45&89.89&97.30&97.49&97.30\\\hline
        012&84.69&94.33&94.17&94.07&87.03&95.41&94.96&94.88\\\hline
        015&83.18&95.03&93.03&92.81&85.05&95.53&93.58&93.32\\\hline
        019&78.62&88.43&87.80&87.39&80.25&89.07&88.50&88.08\\\hline
        020&87.00&96.80&96.90&96.80&89.27&97.69&97.78&97.67\\
\hline
\end{longtable}

\begin{longtable}{ | c | c |  c | c | c || c  | c | c | c |}
    \caption{Results of team MediCIS for task5 recognition.}
    \label{supMat:MediCIS_Task5}\\
    \hline
    \multirow{2}{*}{Sequence}&\multicolumn{4}{c||}{Frame-by-Frame}&\multicolumn{4}{c|}{Application-Dependant}  \\\cline{2-9}
      &Accuracy&Precision&recall&F1&Accuracy&Precision&recall&F1\\\hline
    \endfirsthead
    \caption[]{Results of team MediCIS for task5 recognition. (continued)}\\
    \hline
    \multirow{2}{*}{Sequence}&\multicolumn{4}{c||}{Frame-by-Frame}&\multicolumn{4}{c|}{Application-Dependant}  \\\cline{2-9}
      &Accuracy&Precision&recall&F1&Accuracy&Precision&recall&F1\\\hline\endhead
        003&86.27&96.85&96.70&96.63&88.56&97.72&97.55&97.50\\\hline
        004&87.75&97.30&97.20&97.13&89.50&97.97&97.88&97.83\\\hline
        006&85.08&97.10&97.12&97.09&86.76&97.88&97.93&97.89\\\hline
        009&87.84&96.41&96.55&96.36&89.60&97.23&97.44&97.23\\\hline
        012&85.49&95.99&96.23&96.06&88.12&96.95&97.20&97.04\\\hline
        015&82.71&94.94&93.43&93.38&85.10&95.62&94.10&94.03\\\hline
        019&80.90&91.08&89.96&89.71&82.45&91.79&90.68&90.42\\\hline
        020&87.93&97.50&97.52&97.40&90.42&98.33&98.42&98.28\\
\hline
\end{longtable}

\end{document}